\documentclass{egpubl}
\usepackage{eg2023}
\usepackage{xcolor}
 
\STAR                   %

\usepackage[T1]{fontenc}
\usepackage{dfadobe}  

\usepackage{cite}  %
\BibtexOrBiblatex
\electronicVersion
\PrintedOrElectronic
\ifpdf \usepackage[pdftex]{graphicx} \pdfcompresslevel=9
\else \usepackage[dvips]{graphicx} \fi

\usepackage{egweblnk} 

\usepackage{amsfonts}

\usepackage{csquotes}

\usepackage{pifont} %
\usepackage{booktabs}
\usepackage{multirow}
\usepackage{adjustbox}
\usepackage{fontawesome} %
\usepackage{amsmath} %
\usepackage{float}
\usepackage{caption}
\usepackage{subcaption}
\usepackage{cleveref}
\usepackage{adjustbox}

\usepackage{tabularx}
\usepackage{lscape}
\DeclareMathOperator*{\argmin}{arg\,min}

\usepackage{wrapfig}

\title[State of the Art in Dense Monocular Non-Rigid 3D Reconstruction\extended{~\textcolor{red}{(Extended Version)}}]{State of the Art in Dense Monocular Non-Rigid 3D Reconstruction\extended{\newline\Large \textcolor{red}{---Extended Version---}}}

\EGpagenumber

\newcommand{\etal       }     {\textit{et~al.}}
\newcommand{\apriori    }     {\textit{a~priori}}

\newcommand{\eg         }     {\textit{e.g.~}}

\newcommand{\ie         }     {\textit{i.e.~}}
\newcommand{\naive      }     {{na\"{\i}ve}}

\newcommand{\extended}[1]{}

\definecolor{cmarkcolor}{rgb}{0.49,0.74,0.49}
\definecolor{xmarkcolor}{rgb}{0.86,0.34,0.34}

\newcommand{\xmark}{\textcolor{xmarkcolor}{\ding{55}}}

\definecolor{opA}{rgb}{0.9,0.6,0.0}
\definecolor{opB}{rgb}{0.35,0.70,0.90}
\definecolor{opC}{rgb}{0.8,0.40,0.0}
\definecolor{opD}{rgb}{0.0,0.60,0.50} %
\definecolor{opE}{rgb}{0.8,0.6,0.7}
\definecolor{opF}{rgb}{0.,0.45,0.70}

\def\code#1{
    \ifx&#1&
        \xmark{}
    \else
        {\href{#1}{\faExternalLink}}
    \fi
}

\newcommand{\revision}[1]{#1}

\begin{document}
\author[Edith Tretschk \& Navami Kairanda et al.]
{
\parbox{\textwidth}
{\centering \vspace{-0.5cm}
Edith Tretschk$^{1\star}$~~~Navami Kairanda$^{1\star}$~~~Mallikarjun B R$^{1}$~~~Rishabh Dabral$^{1}$~~~Adam Kortylewski$^{1,2}$\\Bernhard Egger$^{3}$~~~Marc Habermann$^{1}$~~~Pascal Fua$^{4}$~~~Christian Theobalt$^{1}$~~~Vladislav Golyanik$^{1}$
}
\\
\parbox{\textwidth}
{\centering
$^{1}$Max Planck Institute for Informatics, Saarland Informatics Campus~~~
$^{2}$University of Freiburg\\
$^{3}$Friedrich-Alexander-Universität Erlangen-Nürnberg (FAU)~~~
$^{4}$EPFL~~~$^{\star}$denotes equal contribution
}
}

\teaser{
  \vspace{-0.8cm}
  \includegraphics[width=\linewidth]{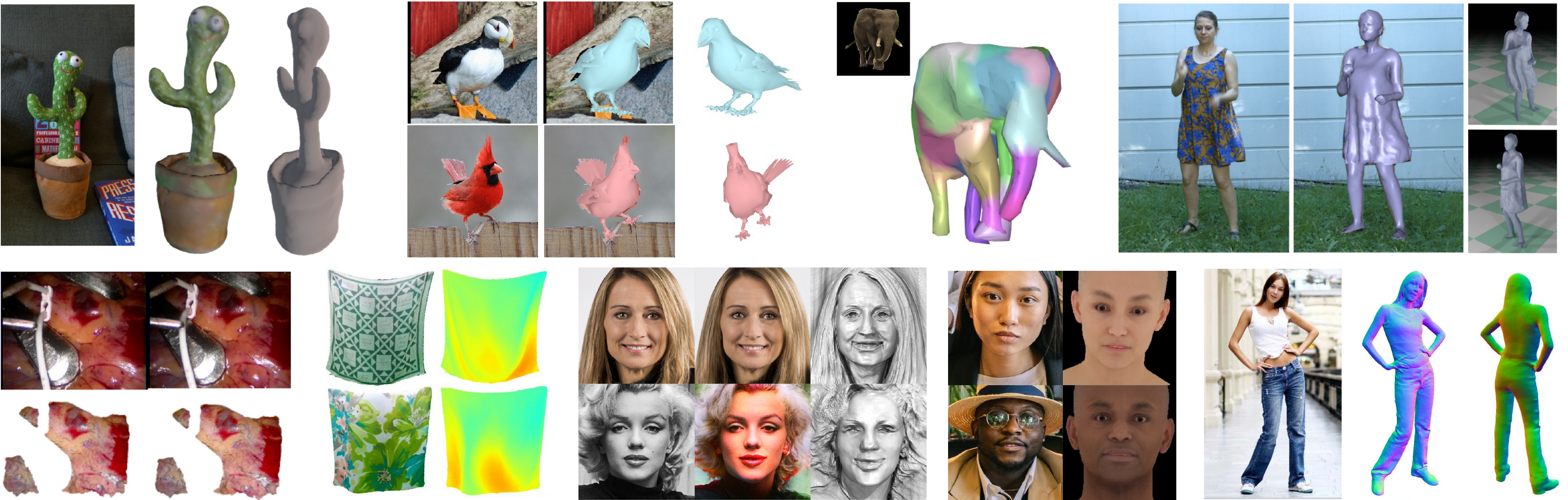}
  \centering
  \caption{We review state-of-the-art methods for the dense 3D reconstruction of deformable objects from monocular images and videos, such as general deformable surfaces, soft body tissues in medical scenarios, animals, and human bodies and body parts. 
  Images adapted from \cite{johnson2022ub4d, wang21aves,  yang2021viser, DeepCap, Sidhu2020, kair2022sft, Chan2021,  Feng:TRUST:ECCV2022, saito2020pifuhd}. 
  }
  \vspace{0.3cm}
  \label{fig:teaser}
}

\maketitle

\begin{abstract}
3D reconstruction of deformable (or \emph{non-rigid}) scenes from a set of monocular 2D image observations is a long-standing and actively researched area of computer vision and graphics. 
It is an ill-posed inverse problem, since---without additional prior assumptions---it permits infinitely many solutions leading to accurate projection to the input 2D images. 
Non-rigid reconstruction is a foundational building block for downstream applications like robotics, AR/VR, or visual content creation. 
The key advantage of using monocular cameras is their omnipresence and availability to the end users as well as their ease of use compared to more sophisticated camera set-ups such as stereo or multi-view systems. 
This survey focuses on state-of-the-art methods for dense non-rigid 3D reconstruction of various deformable objects and composite scenes from monocular videos or sets of monocular views. 
It reviews the fundamentals of 3D reconstruction and deformation modeling from 2D image observations. 
We then start from general methods---that handle arbitrary scenes and make only a few prior assumptions---and proceed towards techniques making stronger assumptions about the observed objects and types of deformations (\eg human faces, bodies, hands, and animals). 
A significant part of this STAR is also devoted to classification and a high-level comparison of the methods, as well as an overview of the datasets for training and evaluation of the discussed techniques. 
We conclude by discussing open challenges in the field and the social aspects associated with the usage of the reviewed methods. 


\end{abstract}

\section{Introduction}\label{sec:introduction} 

Humans can close one eye, look around, and get a fair sense of their surroundings in terms of their 3D geometry, appearance, and even deformations \cite{Schiller1986, WadeJones1997}. 
Nevertheless, designing computational methods that densely reconstruct a dynamic scene in 3D using a single monocular camera remains a challenging task that is far from solved, as this STAR shows. 

Monocular 3D reconstruction is a challenging domain of computer vision and graphics motivated by fundamental questions and practical applications. 
The rigid case has been studied for decades and mature methods are available nowadays \cite{LonguetHiggins1987,agarwal2011building, MurArtal2015}; 
the rigidity assumption, \textit{i.e.,} that the transformation can be entirely described by a single translation and a single rotation (6DoF), significantly simplifies the formulation compared to the non-rigid case ($>$6DoF, often $\gg$6DoF) and provides strong prior knowledge about the expected 3D structure. 
At the same time, while some objects preserve their states longer than others, all eventually deform over time \cite{Laraudogoitia2022} while being exposed to physical forces. 
We thus live in a constantly changing world, irrespective of the scale: The scale of our galaxy, the solar system, Earth, ecosystems on our planet, individual living species, humans, human body parts (\eg face and hands) and organs (\eg heart), cells or atoms. 
Many spectacular dynamic effects are inherently non-rigid.

There is much interest in monocular approaches both in the computer vision and graphics communities, as evidenced by the many published works \revision{in a wide range of domains, \eg monocular depth estimation \cite{Garg2016}, image segmentation \cite{Li2022}, or image synthesis \cite{Karras2019}.} 
Since RGB cameras are ubiquitous and single-camera setups are much easier to deploy than multi-camera ones, monocular methods are relevant not just out of sheer curiosity about the limits of reconstruction 
under the most challenging conditions 
but also because they enable a multitude of applications. 
\revision{Applications for monocular 3D reconstruction} range from geometry acquisition  \cite{agarwal2011building}, novel view synthesis  \cite{tretschk2021non} and elastic parameter estimation  \cite{kair2022sft} to scene or video editing \cite{Garrido2016} and 
scene recognition and understanding \cite{Cordts2016}. 
All these applications are highly relevant for such fields of science and engineering as VR/AR,
movie and game production, content creation, computer-assisted design, cultural heritage, 
robotics, space exploration, experimental physics, medicine, zoology and many others. 
In other words, reliable solutions to monocular 3D reconstruction have the potential to impact society in significant ways.

This is the first STAR devoted to non-rigid 3D reconstruction from single monocular cameras (we discuss related surveys in Sec.~\ref{sec:related_surveys}). %
In recent years, monocular 3D reconstruction has been reinvigorated by several breakthroughs, including widely applicable parametric models~\cite{SMPL-X:2019, Qian2020, li2021coarsetofine}; neural  parametrizations~\cite{mildenhall2020nerf}; machine learning  techniques~\cite{cmrKanazawa18}; high-quality, large-scale datasets \cite{WahCUB_200_2011, Liu2015, Moon2020}; and powerful computational resources, to name a few. 
Thus, monocular reconstruction methods nowadays produce 3D outputs of impressive visual quality that are suitable for many applications discussed above, including computer graphics; see Fig.~\ref{fig:teaser} for representative 3D reconstructions by state-of-the-art methods. 
Even a few years back, this was not generally true. 
Still, despite great progress, there remain a lot of unsolved problems in monocular non-rigid 3D reconstruction. 
Since the field has recently undergone massive change, we seek to document its current state and the challenges researchers will face during the upcoming years.

\subsection{Scope of this  STAR}\label{sec:scope} 

This STAR focuses on methods from recent years for non-rigid 3D reconstruction  that take  one or several consecutive views 
from a single camera %
as input and that output dense 3D reconstructions of the scene in each view or point in time spanning the  observations. 
\extended{
We cover both state-of-the-art methods that are learning-based (neural) or model/optimization-based (classical). }
We put a special emphasis on the emerging fields of neural scene representations and neural rendering, %
 physics-based reconstruction, and 
 reconstruction from event cameras. 
We next explain the meaning of each core word of this STAR's title in more detail.

\noindent\textbf{Dense.} 
We focus on dense 3D reconstructions and leave the  sparse case out of scope for several reasons: 
1) Dense reconstructions provide a more complete scene description; %
2) Nowadays, sufficient computational resources and the availability  %
of reliable dense preprocessing methods %
allow many downstream applications to assume dense deformable 3D scenes; 
3) Many principles are shared between the dense and sparse cases %
and, hence, most fundamentals we discuss in Sec.~\ref{sec:fundamentals} cover both; and %
4) Considering the literature volume in the field, even a survey of a format like this one cannot cover both cases with satisfactory depth.

\noindent\textbf{Monocular.} 
We only consider methods where no more than a single view observes each 3D scene state (no multi-view). %
We mainly focus on sensors that register incoming light in the visible spectrum (wavelengths in the range $320{-}1100$ nm). 
Thus, event cameras %
are in scope of this STAR (see Sec.~\ref{sec:rendering-into-2d}) but active sensor  systems with active emitters such as RGB-D cameras are not. 
However, we only apply these criteria at test time, and hence any supervision  (including 3D) at training time is inside the scope. 

\noindent\textbf{Non-Rigid.} %
We only consider 
objects that can deform. %
We cover methods for static (3D; single timestep) and dynamic (4D; multiple timesteps) reconstruction. 
Many approaches parametrize non-rigid deformations by statistical 3D 
models. %
Widely-used 
parametric human body models %
neglect different clothing styles, facial expressions and hairstyles, and only provide shape proxies, \ie approximate shapes that do not allow recognizing a person from the reconstructed geometry. 
We thus believe it is not enough to %
instantiate a parametric human body model to claim that an approach reconstructs dense 3D human shapes. 
Hence, methods that do not perform geometric refinement and 
add identity-specific characteristics on top of shape proxies are out of our scope. 
The situation is different with human hands, human faces, and animals. 
Hands are mostly observed naked, are easier to capture, and vary less across people. 
Similar statements apply to faces. 
Next, there is little work on dense 3D animal reconstruction from monocular views. 
Hence, we include methods that use %
parametric  hand, face, and animal models. %

\noindent\textbf{3D.} 
We focus on true 3D representations and ignore image-based (\eg 2.5D/depth) or intermediate representations (\eg light fields). %

\noindent\textbf{Reconstruction.} 
We seek a model that ideally represents the scene as it was observed. %
We do not require it to be generative or editable. %

\subsection{Related Surveys}\label{sec:related_surveys}

Several method surveys and STARs were published over the last twelve years; 
some of them are outdated as of 2022. 
Salzmann and Fua \cite{salzmann2010deformable} %
review methods for deformable 3D surface reconstruction, only 
a few of which
addressed the dense case in 2010. %
Jensen \etal~\cite{Jensen2021} review Non-Rigid Structure-from-Motion (NRSfM) and introduce the sparse 
\emph{NRSfM 2017 challenge dataset}. %
They focus on sparse techniques from before 2020; dense NRSfM techniques are not systematically discussed unless evaluated on the proposed dataset. 
A recent short survey %
\cite{khan2022} reviews generalizable deep-learning methods for dense 3D reconstruction of rigid and non-rigid objects from a single image, with weak prior knowledge about the object class. %
Our report is much more exhaustive, including category-specific methods and volume-rendering techniques. 
A STAR by Zollhoefer \etal~\cite{Zollhoefer2018_3D} %
focuses on rigid and non-rigid 3D reconstruction %
from RGB-D cameras, which are out of our scope. 
STARs on neural rendering \cite{sotaNeuralRendering, neuralFields, tewari2021advances} focus on novel view synthesis of rigid and non-rigid scenes. 
They cover a small subset of  techniques that we cover. %
Other surveys are devoted to only face  \cite{zollhofer2018state, egger20203d} or bird reconstruction \cite{MarvastiZadeh2022}. 
Tian \etal~\cite{Tian2022} cover monocular 3D human mesh recovery %
using parametric %
models. %
In contrast, we focus on monocular methods that can regress human shapes beyond shape proxies and naked humans. 
An unpublished survey \cite{xia2023aware} discusses 3D-aware image synthesis methods but covers only a few of the non-rigid methods covered by our survey. 
A recent survey  \cite{Gallego2022} on event-based vision covers \emph{static} simultaneous localization and mapping (SLAM) from event cameras, while we discuss \emph{non-rigid} event-based methods. %

\textit{All in all, the STAR at hand is the first one that systematically reviews all types of monocular dense non-rigid 3D reconstruction techniques for various scenes and objects (together with the fundamentals for introducing the field or catching up with the field), whereas previous surveys address only small parts of this report. %
}

\subsection{Paper Selection Criteria} 
We predominantly discuss works from international computer vision and graphics conferences and journals that are in scope of this STAR (\textit{cf.}~Sec.~\ref{sec:scope}). 
We also include a few recent technical reports on arXiv.org. 
However, considering how fast the field is developing, we cannot claim completeness in either case.

\subsection{Structure of this STAR} 
We first motivate this STAR %
in this introductory Sec.~\ref{sec:introduction}. 
We next describe the basics of non-rigid 3D reconstruction  in Sec.~\ref{sec:fundamentals}, which covers many aspects that are useful in order to understand any particular work on non-rigid 3D reconstruction. 
At the core of our report is the discussion of the current state of the art in Sec.~\ref{sec:works}, which is ordered by the object category that is to be reconstructed. 
We discuss cross-sectional aspects and open challenges in Sec.~\ref{sec:discussion}. 
Finally, we provide an overview of the impact that this field has on society in Sec.~\ref{sec:social}, and draw conclusions in Sec.~\ref{sec:conclusion}. 
\section{Fundamentals}\label{sec:fundamentals} 

This section describes the main building blocks of the design pipeline of non-rigid 3D reconstruction methods. 
We aim to \revision{provide a guide to} the reader of all the pieces involved, making a critical reading of recent works possible. 
However, we do not claim full coverage of all aspects. 
For example, we focus on the basics of computer graphics and computer vision required to understand this STAR, and we assume pre-requisite knowledge in machine learning on the part of the reader. %
We first take a functional look at the components of 3D reconstruction in Sec.~\ref{sec:background}, which we then use in Sec.~\ref{sec:problem} to formulate the reconstruction problem that we are concerned with. 
In Sec.~\ref{sec:parametrization}, we describe how to parametrize the functions discussed earlier. 
Sec.~\ref{sec:dataterms} describes data terms that are commonly used to obtain consistency between the input and the model parametrization. 
Sec.~\ref{sec:difficulties} discusses multiple challenges that we face when trying to obtain a solution. 
Then, in Sec.~\ref{sec:priors}, we specifically look at the underconstrained nature of the problem and provide a high-level description of several standard priors to tackle it. %
Finally, Sec.~\ref{sec:optimization} shortly describes the optimization of the solution parameters with respect to the resulting loss function.

\subsection{Background: A Functional Look}\label{sec:background}

In this section, we introduce basic concepts from computer graphics\revision{\cite{foley1996computer}}: geometry, deformations, appearance, and rendering. 
\revision{For didactic purposes, we explicitly split the commonly used term \emph{representation} into its two constitutive concepts of \emph{function} and \emph{parametrization}.} 
While this section takes a theoretical, functional perspective, we present practical parametrizations in a later section. 

\extended{
We take a geometry-centered perspective here. 
For a more theoretical, light-centered perspective, we refer to Bergen \etal~\cite{bergen1991plenoptic} for a thorough seminal treatment of the plenoptic function, a full description of the observations made by cameras.  
}

\noindent\textbf{Notation.} 
We use $\mathbf{x} \in \mathbb{R}^n$ for a vector, $\mathbf{A}\in\mathbb{R}^{m\times n}$ for a matrix, $S$ for a set, $f: S\to T$ for a function from $S$ to $T$, $\times$ for the cross product. 
We use $d$ for deformations; subscripts, \eg $d_t$, to denote time $t$; vertex index $i$ in triangulations. 
\revision{Time} derivatives have dots on top: $\dot{\mathbf{x}},\ddot{\mathbf{x}}$. 

\subsubsection{Geometry Functions}
We first require a representation of the 3D geometry of the object or scene. %
The most common way of representing an object's geometry is via its 2D surface $S \subset \mathbb{R}^3$. 
A 2D surface can be described \emph{implicitly}, \ie defined on a volumetric/3D domain, via an \revision{\emph{indicator function}} $s:\mathbb{R}^3\to \{0,1\}$ that is 1 on the surface and 0 otherwise: 
\begin{align}\label{eq:volume_occupancy}
    s(\mathbf{x}) = 
    \begin{cases}
    1 & \text{if } \mathbf{x} \in S\\
    0 & \text{else}
    \end{cases}.
\end{align}
A level-set function provides additional information about where the surface can be found:
\begin{align}
     s(\mathbf{x}) = \min_{\mathbf{y}\in S } \lVert \mathbf{x} - \mathbf{y} \rVert_2.
\end{align}
This \emph{unsigned distance function (UDF)} %
specifies, for each point in 3D space, how far away it is from the closest surface. 
The surface lies at the 0-level set: $\{\mathbf{x} | s(\mathbf{x}) = 0\}$. 
This can be turned into the \emph{signed distance function (SDF)} by giving points inside the object negative distance, \ie for $\mathbf{x}$ inside the object, $\mathit{SDF}(\mathbf{x}) = -\mathit{UDF}(\mathbf{x})$. 
\revision{Thus, an SDF is only defined for closed surfaces.} 

In contrast to these implicit surface representations, we can make use of the mathematical definition of a 2D manifold embedded in 3D space. 
A manifold admits an \emph{explicit} surface description: a map from a subset $U$ of 2D space to 3D space \revision{(by embedding all charts of the surface atlas into the same $\mathbb{R}^2$ space)}. 
The resulting surface of such a UV map, $\mathit{UV}:U \subset \mathbb{R}^2 \to \mathbb{R}^3$, is called a \emph{parametric surface}:
\begin{align}
    s(\mathbf{u}) = \mathit{UV}(\mathbf{u}).
\end{align}

We sometimes want to model objects without forcing a clearly defined surface on them (\eg smoke) or without putting the surface at the center of the parametrization (\eg fluids). %
In such cases, volumetric representations encode geometry in a soft manner. 
Density fields $v:\mathbb{R}^3 \to [0,+\infty)$ are the most common volumetric function, where a density of $0$ denotes empty space:
\begin{align}\label{eq:volume_density}
    v(\mathbf{x}) = \operatorname{density}(\mathbf{x}).
\end{align}
Furthermore, applying volumetric representations to surfaces allows for more slack in the optimization~\cite{mildenhall2020nerf}.

\subsubsection{Deformation Functions} \label{subsec:deformations}

\begin{figure}
    \centering
    \includegraphics[width=\linewidth]{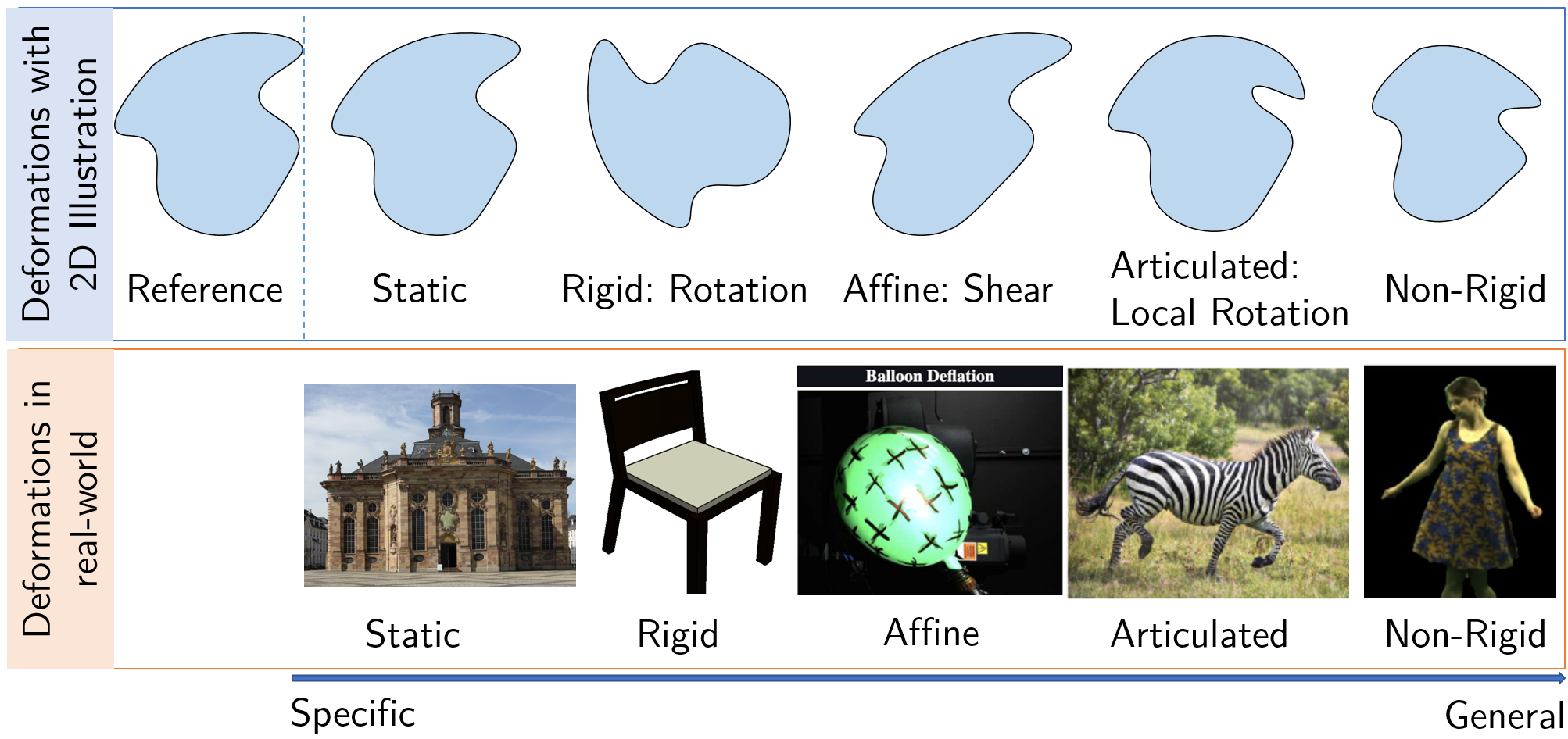}
    \caption{\small Different types of deformations%
    , in order of expressiveness. %
    Top: Illustration of the deformation map from the reference to the deformed geometry%
    . Bottom: Example objects %
    commonly studied in computer vision. 
    Images adapted from \cite{rudnev2022nerfosr,chang2015shapenet,Jensen2021,yao2022lassie,DeepCap}.
    }
    \label{fig:deformable_scenes_examples}
\end{figure}

We next specify different types of deformations that can be applied to a geometry, \revision{with each one generalizing the previous type}, see Fig.~\ref{fig:deformable_scenes_examples}. 
\extended{We discuss static and dynamic deformations. }
We define the deformed geometry $g_d$ with respect to an undeformed template or reference geometry $g$, where \eg $g=s$ for a surface. 

\noindent\textbf{Static} objects do not move locally or globally%
, with the deformed geometry $g_d$ trivially defined as $g_d(\mathbf{x}) = g(\mathbf{x})$. 

\noindent\textbf{Rigid} objects may move around globally, without changing their shape and size. 
\extended{
Specifically, the object can change its orientation and position, but cannot scale or shear. }
Mathematically, the deformed geometry $g_d$ can rotate and translate: $g_d(\mathbf{x}) = g(\mathbf{R} \mathbf{x} + \mathbf{t})$, for a 3D rotation matrix $\mathbf{R}\in\mathit{SO}(3)$ and a 3D translation $\mathbf{t}\in\mathbb{R}^3$. 
This is an idealization where the deformation is so small that it can be neglected. 
\extended{If $\det{R}=-1$ is also allowed, it is a \emph{reflection}. 
If $\det{R}=\pm 1$ and a scaling $s\in\mathbb{R}$ is allowed, it is a \emph{similarity} transform: $g_d(\mathbf{x}) = g(s\mathbf{R} \mathbf{x} + \mathbf{t})$.
}

\noindent\textbf{Affine} deformations are also global like rigid deformations, except that now shearing and scaling are allowed: $g_d(\mathbf{x}) = g(\mathbf{A} \mathbf{x} + \mathbf{t})$, where $\mathbf{A}\in\mathbb{R}^{3\times 3}$ is \revision{an invertible} linear map.

\noindent\textbf{Articulated} deformations generalize the previous classes to (piece-wise) ensembles of $N$ local rigid/affine deformations $\{ d_i \}^N_{i=1}$, where $d_i(\mathbf{x})=\mathbf{R}_i \mathbf{x} + \mathbf{t}_i$ in the rigid case:
\begin{align}
    g_d(\mathbf{x}) = g( d_i(\mathbf{x})) \quad \text{if}~d_i(\mathbf{x}) \in U_i,
\end{align}
where $\{U_i \subset \mathbb{R}^3\}^N_{i=1}$ is a partition of $\mathbb{R}^3$. 
$U_i$ is a \emph{local part} and deforms according to its own associated deformation $d_i$. 
\extended{
Thus, the deformed geometry is obtained by deforming each local part $U_i$ of the template geometry according to its own rigid/affine transform $d_i$. 
} 
Humans or animals are sometimes modeled with articulated deformations. %
\extended{
There, each large bone (\eg an upper arm) would be a local part. %
}

\noindent\textbf{Non-Rigid}
objects undergo elastic deformations as they respond naturally to applied forces, constraints and contacts with self or obstacles. 
This most generic formulation describes the behavior of most real, physical objects and deformation $d$ is any \revision{(in general} non-linear\revision{)} map that displaces an undeformed point to a deformed one: %
\begin{align}
    g_d(\mathbf{x}) = g( d(\mathbf{x})), \text{ where } \mathit{d}:\mathbb{R}^3 \to \mathbb{R}^3.
\end{align}
\extended{
For instance, cloths deform freely and often form fine local surface deformations such as folds and wrinkles. 
}

\extended{
The above definitions are for equilibrium deformations; dynamically deforming states of an object vary as a function of time: 
\begin{align}
    g_d(\mathbf{x}, t) = g( d(\mathbf{x}, t)) \text{ where } \mathit{d}:\mathbb{R}^3 \times \mathbb{R}^{+} \to \mathbb{R}^3.
\end{align}
We use the short-hand notation $\{g_t\}_t$ for deformed states at each timestep $t$.
}

\noindent\textbf{Deformation Measures} 
can be used to quantify the \revision{geometric} amount of deformation and they often derive %
from differential geometry of solids, surfaces, and curves\revision{~\cite{do2016differential}}. %
As they measure deformations, they need to be invariant under rigid transformations. 
We later derive deformation constraints from them that act as priors, see Sec.~\ref{sec:priors}. %
\par
Let $\mathbf{m}$ be the \textit{intrinsic} or \textit{material coordinates} of a point in an undeformed object. 
For volumes, surfaces, and curves, we have $ \mathbf{m}\in\mathbb{R}^3$, $\mathbf{m}\in\mathbb{R}^2$, and $\mathbf{m}\in\mathbb{R}$, respectively. 
We can map the point from its material position to the world-space position $\mathbf{x} = \mathbf{x}(\mathbf{m}) \in \mathbb{R}^3$. 
The shape of any such object is determined by the Euclidean distances between nearby \revision{world-space} points. 
Non-rigid deformations \revision{may} change these distances, while rigid deformations \revision{(and reflections)} do not. 
If $\mathbf{m}$ and $\mathbf{m} + d\mathbf{m}$ are the material coordinates of two nearby points, the differential length $\mathit{dl}$ between them in the deformed object is: 
\begin{align}
    \mathit{dl} = \sum_{i, j} G_{ij} dm_{i} dm_{j},
\end{align}
where the symmetric and positive definite matrix $ G \in \mathbb{R}^{3\times3}$, \ie 
\begin{align}
G_{ij}(\mathbf{x}(\mathbf{m})) = \frac{\partial \mathbf{x}}{\partial m_i} \cdot \frac{\partial \mathbf{x}}{\partial m_j}
\end{align}
is known as the \textit{metric tensor} or the \textit{first fundamental form}.
Two volumetric objects have the same shape (up to a rigid motion) \revision{and considered \textit{isometric}} if their metric tensors are identical functions of $\mathbf{m}$ %
everywhere. 
\revision{Isometry} is not a sufficient condition for rigidity \revision{of surfaces and curves which are} volumes infinitesimally thin in one or two dimensions. %
Surfaces can change shape by \revision{bending}, even while preserving \revision{geodesic} distances between nearby points. 
\revision{While the metric tensor describes the in-plane stretching and shearing of surfaces,}
the \textit{curvature tensor} $ B \in \mathbb{R}^{2\times2}$, or \textit{second fundamental form}, \revision{quantifies bending and} %
is defined as: %
\begin{align}
B_{ij}(\mathbf{x}(\mathbf{m})) = \mathbf{n} \cdot \frac{\partial^2 \mathbf{x}}{\partial m_i \partial m_j}, 
\end{align}
where $\mathbf{n}$ is the unit surface normal. 
Together, %
they form the \textit{shape operator} $G^{-1}B$. 
Two surfaces have the same shape if their first and second fundamental forms are identical. 
\extended{
In the case of curves in space, the metric and curvature tensors are called \textit{arc length} $s(\mathbf{x}(m))$ and \textit{curvature} $\kappa(\mathbf{x}(m))$. 
Curves have a further degree of freedom: the twisting around themselves. %
Thus, the deformation of curves can be locally determined by the change in arc length, curvature, and \textit{torsion} $\tau(\mathbf{x})$. 
}

\label{para:dynamics}
\noindent\textbf{Deformation Dynamics} of an object occur under applied forces. 
The \revision{dynamics} are determined by \revision{the object's} initial shape and its material configuration. 
Continuum mechanics\revision{~\cite{sifakis2012fem}} and elasticity theory~\cite{slaughter2012linearized} %
formulate quantitative descriptions for the deformation of a continuous object.
One can then arrive at partial differential equations (PDEs) that model the dynamic behaviors of objects. %
\par
Let the time-varying \textit{deformation map} be given as $ \mathbf{x} = \mathbf{x}(\mathbf{m}, t)$, with $\mathbf{X} \equiv \mathbf{x}(\mathbf{m}, 0)$ as the undeformed \emph{reference configuration}. %
An important physical quantity derived from $\mathbf{x}$ is the \textit{deformation gradient} $\mathbf{F} \revision{\equiv \nabla_\mathbf{m} \mathbf{x}}\in \mathbb{R}^{3 \times 3}$, the spatial Jacobian of the deformation map.
The metric tensor $ G \equiv \mathbf{F}^T\mathbf{F}$ provides a measure for local distortion of lengths and angles relative to the reference shape \revision{(and connects the physical response described here with the geometric deformations described in the previous subsection)}. 
We can derive strain measures from the deformation gradient to quantify the geometric severity of the deformation.
Assuming an identity metric tensor for the reference shape, the \textit{Green strain tensor} $\epsilon \in \mathbb{R}^{3 \times 3}$, a commonly used strain measure, is given as: 
\begin{align}
\epsilon = \frac{1}{2} (\mathbf{F}^T\mathbf{F} - \mathbf{I}).
\end{align}
This strain omits information unrelated to shape change from $\mathbf{F}$ but retains information about the local deformation magnitude. 
\extended{
For instance, in the case of static objects, $\epsilon = \mathbf{0}$ as $\mathbf{F} = \mathbf{I}$. 
In the case of rigid objects, where $\mathbf{x} = \mathbf{R} \mathbf{X}+ \mathbf{t}$, %
we have $\mathbf{F} = \mathbf{R} $ 
and thus $\epsilon = \mathbf{0}$, since rotation matrices are orthogonal: $\mathbf{R}^{T}\mathbf{R} = \mathbf{I}$.
}

As a result of elastic deformation, the objects accumulate  potential energy and the resulting internal elastic forces are often described by the \textit{Cauchy stress tensor} $\sigma \in \mathbb{R}^{3 \times 3}$. %
Constitutive models relate the (geometric) strain to the (physical) material response it triggers, such as force, stress, or strain energy. 
In its most general form, the constitutive equation is formulated as \textit{Hooke's law}:
\begin{align}
    \label{eq:hookes_law}
    \sigma_{ij}= \mathbf{C}_{ijkl} \epsilon_{kl},
\end{align} 
where $\mathbf{C}$ is a rank-4, possibly non-linear elastic tensor. 
\extended{
Depending on the physical properties of the object, 
there exist linear/non-linear, isotropic/an-isotropic, or volume-preserving formulations of the material models, including linear elasticity, corotated linear elasticity, St. Venant-Kirchhoff, and Neo-Hookean models~\cite{sifakis2012fem}. 
While strain is the dimensionless deformation, stress has units of force per length for surfaces or per area for volumes. 
}

\begin{figure}
    \centering
    \includegraphics[width=\linewidth]{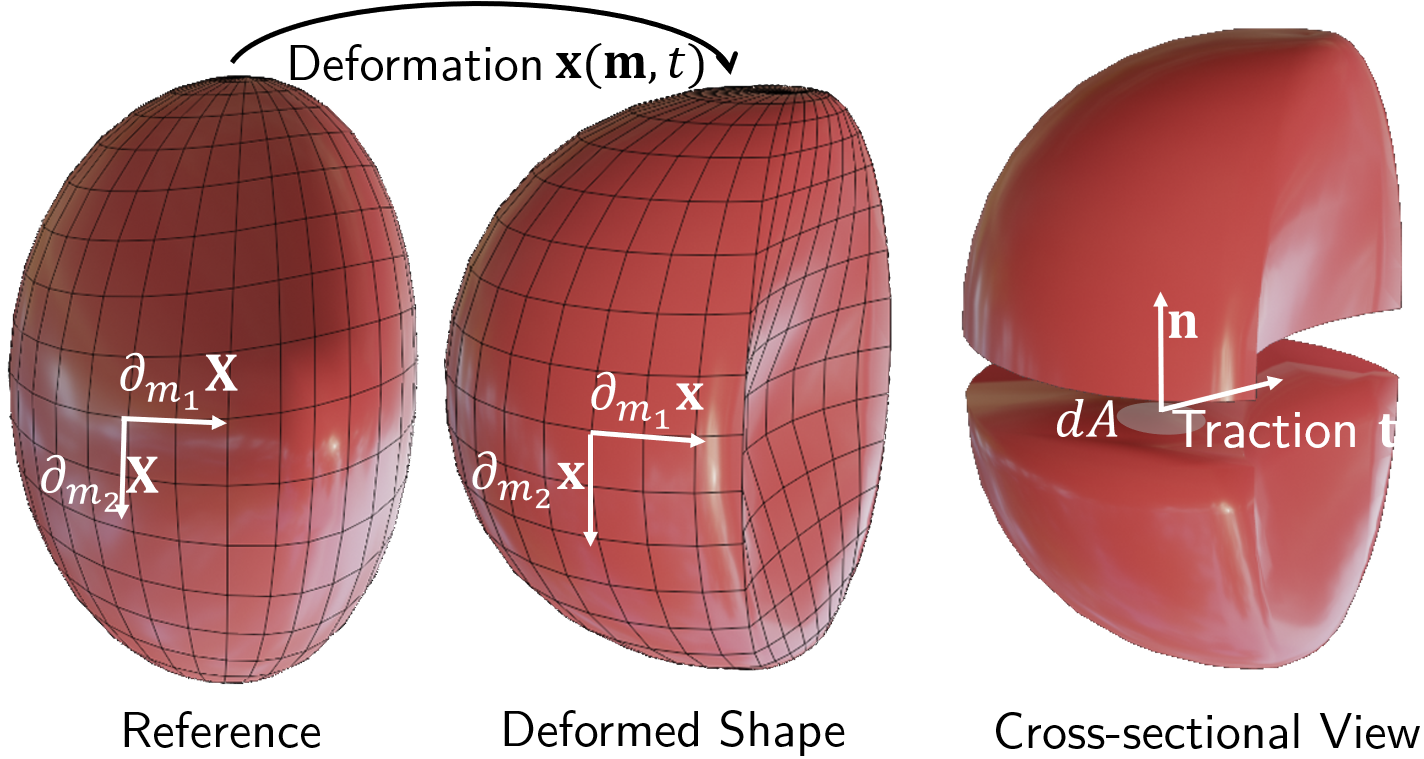}
    \caption{\small 
    Local tangents on the volume can be used to measure the deformation between the reference $\mathbf{X}$ and the deformed state $\mathbf{x}$. 
    The traction $\mathbf{t}$ is the density of the internal forces on the area $dA$ with normal $\mathbf{n}$. %
    }
    \label{fig:deformable_dynamics}
\end{figure}

The distribution of internal elastic forces that result from a deformation can be described as follows: 
Consider a slice of the deformable object with differential surface area $dA$ and normal $\mathbf{n}$; see Fig. \ref{fig:deformable_dynamics}. 
Then the traction $\mathbf{t}$ along the cut is the (surface) force density function that measures the force per unit undeformed area:
 \begin{align} \mathbf{t} = \lim_{dA\to 0} \frac{d\mathbf{f}}{dA} = \sigma \mathbf{n},
 \end{align} 
where the Cauchy stress tensor $\sigma$ serves as a fundamental force descriptor that generalizes traction for every normal direction $\mathbf{n}$ and relates the internal forces to the deformations using \eqref{eq:hookes_law}. 
\par
Consider now a volume element $V$ of a deformable object with boundary surface $\partial{V}$.
Let $\mathbf{f}$  be the external forces such as gravity, wind per unit volume acting on $V$. 
Balance of linear momentum postulates that the resultant of the external forces acting on the object is equal to the rate of change of its total linear momentum: %
\begin{align}
      \label{eq:weak_form}
     \int_{\partial{V}} \mathbf{t}~da + \int_{V} \mathbf{f}~dv = \int_{V} \rho \mathbf{\ddot{x}} ~dv, 
\end{align}
where $\rho$ is the mass density and $\mathbf{\ddot{x}}$ is the material acceleration, which together constitute the inertial forces. 
The divergence theorem lets us change the surface integral in \eqref{eq:weak_form} into a volume integral: 
\begin{align}
     \int_{\partial{V}} \mathbf{t}~da = \int_{\partial{V}} \sigma \mathbf{n}~da  = \int_{V} \text{div}~\sigma~dv.
\end{align}
As \eqref{eq:weak_form} 
must hold for any enclosed volume, the point-wise equation of motion, the so-called strong form, follows as: 
\begin{align}
   \label{eq:motion}
    \text {div }\sigma (\mathbf{x},\mathbf{\dot{x}} ) +  \mathbf{f} =\rho \mathbf{\ddot{x}},
\end{align}
after  accounting for velocity-dependent damping forces. 
\par
\revision{This 3D case gives rise to specialized theories when one of the dimensions becomes very small; for example, the continuum mechanics of 2D surfaces are given by the classical \emph{Kirchhoff-Love shell theory}~\cite{wempner2003mechanics}, while two very small dimensions lead to beam theory, for example \emph{Euler-Bernoulli beam theory}~\cite{zienkiewicz2000finite} for 1D curves}. %

\subsubsection{Appearance Functions}
Apart from geometries along with their deformations, a 3D scene description in computer graphics requires specifying the lights and the material models. 
While geometry captures the macro-structure of an object or scene, material is determined by the object's micro-structure.
Then, the interaction of the geometry with the lights determines its (surface) appearance towards a camera. 
Consequently, physically-based simulation of light transport forms the basis for rendering. 
Based on the material composition and surface roughness, same geometry can reflect light differently and thus have different surface appearance~\cite{lensch2005realistic}.  
The commonly used \textit{Lambertian} or \textit{diffuse} material refers to rough surfaces, where light is reflected multiple times within the material that it loses directionality. 
It thus does not vary with the viewing angle. %
A smoother surface generates \textit{glossy} appearance while perfectly smooth one leads to a \textit{specular (mirror)} reflection.
More generally, \textit{Bidirectional Reflectance Function (BRDF)} describes surface reflection where the appearance depends on illumination direction and viewing direction. %
In inverse rendering, appearance is commonly factorized as view-independent diffuse albedo, view-dependent specular BRDF, normals and light visibility for all incoming directions~\cite{zhang2021nerfactor}. 

\subsubsection{Camera} 
\revision{We also need to model the sensor that collects the input data from which we seek a reconstruction.} 
Physically, a camera sensor collects incoming light rays (photons) and translates them into digital signals along $c$ channels on an image grid $\textbf{I}\in \mathbb{R}^{h \times w \times c}$ of height $h$ $\times$ width $w$ many pixels. 

\noindent\textbf{Camera Models.}
Camera models are described by their intrinsics. 
The pinhole camera model is the most widely used model; %
it parametrizes %
a perspective projection by a focal length $f\in\mathbb{R}$ and the location of the camera center $c_x,c_y \in \mathbb{R}$. 
Then 3D point $\mathbf{x} = (x,y,z) \in \mathbb{R}^3$ projects to the 2D point $(u,v) \in \mathbb{R}^2$ in the image plane as $u = f  \frac{x - c_x}{z} $ and $v = f  \frac{y - c_y}{z}  $.
Other models like weak perspective or \revision{orthographic} camera models are also sometimes used. 
Due to lens distortions, real-world cameras do not follow a simple camera model exactly, which needs to be accounted for by the model if very high fidelity is desired. 
The extrinsic placement of the camera in the world can be described by its position, or \emph{translation} $\mathbf{t}\in\mathbb{R}^3$, and its orientation, or \emph{rotation} $\mathbf{R}\in \mathit{SO}(3)$. %

\noindent\textbf{Camera Types.}
Conventional cameras are RGB cameras that record the red, green, and blue (RGB) colors. 
Formally, a pixel $(u,v)$ of an RGB image $\mathbf{I}_t$ contains the RGB color $\mathbf{c}\in[0,1]^3$ that was captured at time $t$. 
\extended{
Note that usually all pixels are considered to be captured at the same time $t$. 
When high-fidelity reconstructions are sought, rolling shutter effects and similar might be taken into account, where the exact timestamp may vary slightly across pixels of an image. 
}
\emph{Event cameras} are a new camera type that is, as of this writing, rarely but increasingly used \cite{Indiveri2011, Gallego2022}. 
They follow the same camera models as conventional RGB cameras, but they record asynchronous per-pixel brightness changes instead of synchronous 2D images. 
Specifically, an event camera outputs a discrete stream of asynchronous events. 
An event $(u,v,t,p)$ signifies that the brightness at pixel $(u,v)$ has changed at time $t$ by more than some threshold since the last event at that pixel. 
If the brightness has increased, the polarity $p$ is ${+}1$, and ${-}1$ if it has decreased. 
The practical advantages over standard RGB cameras are a high dynamic range and microsecond temporal resolution of events.  
Hence, they allow to capture very fast movements with virtually no motion blur. %

\noindent\textbf{Monocular vs.~Multi-View.} 
A \emph{multi-view} recording is a set of images where each dynamic deformation state is captured by more than one camera. 
A \emph{monocular} recording is a recording of a dynamic scene that is not multi-view. 
There are two important cases: (1) a temporal sequence is a recording of a single scene across time, \eg a typical video; and (2) an image collection is a set of images where each image captures a different scene (not just a different deformation state), \eg Internet image search results for ``people''. 
\extended{
When a static camera observes an object from a single view point, significant occlusions may occur. 
Image sequences recorded with a moving camera resemble multi-view images with deformations in each view \cite{tretschk2021non,park2021nerfies}. 
(This is exploited in NRSfM, where initializing the geometry with a rigid SfM-reconstruction estimate~\cite{TomasiKanade1992} of the non-rigid scene is common.) 
}

\subsection{Problem Setting: What We Aim to Achieve}\label{sec:problem}

Our problem setting takes as input monocular images from a standard RGB camera or an event camera. 
We then seek a 3D reconstruction $g_t$ for each point $t$ in time. 
Optionally, the appearance $a_t$ may also be reconstructed. 
In the case of a temporal sequence, the reconstruction is called a \emph{4D reconstruction}. 
When the geometries $\{g_t\}_t$ are deformed states of a single template $g$, we say that the reconstructions are \emph{in correspondence}. 
Instead of a single template, it is also possible to split a longer sequence into shorter pieces with their own templates (called \emph{keyframes}). 
\extended{
Interestingly, restricting the solution space to solutions that have correspondences can both stabilize the optimization (by propagating information across different deformation states) and lower the consistency with the input data (by effectively acting as a regularizer). 
}

\extended{
The most common and straightforward application of the resulting reconstruction is novel-view synthesis. 
To that end, we can render the obtained reconstruction into a novel camera. %
Due to the 3D nature of the reconstruction, such renderings are guaranteed to be 3D-consistent with each other (also called multi-view consistency). 
}

\subsection{Parametrization: Representing the Solution}\label{sec:parametrization}

In this section, we discuss how to parametrize the functions discussed in Sec.~\ref{sec:background} such that they can be implemented and become tractable to compute. 
For example, while restricted deformation types like rigid, affine, or articulated deformations directly imply a trivial parametrization via their small set of parameters, non-rigid deformations require, \apriori, an infinite number of parameters (one offset vector per point on the geometry). 
We therefore need to design approximate parametrizations that are finitely parametrized \revision{via parameters $\theta$} while still being sufficiently expressive. 
Parametrizations act as hard priors, \ie they impose their assumptions as hard constraints. 
In the later Sec.~\ref{sec:priors} on soft priors, we will also discuss priors that can be encouraged as soft constraints.

\subsubsection{Geometry Parametrizations}\label{sec:geometry_paramtrization}

In the past, geometry used to be parametrized primarily by point clouds and meshes (which provide samples of the surface indicator function), and by voxel grids (which are the natural discretization of volumetric geometry functions). 
Since such samples of the surface indicator function (which is an implicit function) allow direct access to the surface similar to a UV map, such \emph{parametrizations} are called \emph{explicit}. 
In recent years, several concurrent works~\cite{Park_2019_CVPR,Occupancy_Networks,chen2018implicit_decoder,michalkiewicz2019deep} introduced global, coordinate-based \revision{multi-layer perceptron (MLP)s} as geometry parametrization, both for surfaces~\cite{Park_2019_CVPR} and volumes~\cite{Occupancy_Networks}. 
Such neural parametrizations are common in \revision{Neural Radiance Field} (NeRF)-like works~\cite{mildenhall2020nerf,wang2021neus}. 
Since neural parametrizations typically do not provide direct access to the surface, they are called \emph{implicit parametrizations} in both cases. 
(Nonetheless, neural parametrizations can also be used for explicit surface geometry via UV mapping \cite{groueix2018}.) 
Extracting surface meshes from these implicit parametrizations is possible with Marching Cubes \cite{lorensen1987marching,Park_2019_CVPR,mildenhall2020nerf}. 
Several differentiable variants of Marching Cubes exist, which allow to define losses on the extracted geometry and then backpropagate gradients into the implicit parametrization \cite{liao2018deep,remelli2020meshsdf,shen2021deep}.

\subsubsection{Appearance Parametrizations}

When parametrizing a reconstruction, geometry is usually primary and appearance is afterwards attached to it in a suitable manner, \eg by defining the local appearance on each mesh vertex. %
We note that arbitrary properties can be similarly coupled to the geometry. 
In works where challenging appearance is not the focus, a time-invariant Lambertian model is the first choice due to its simplicity. %
It can be coupled with the geometry via a UV map. 
If the UV map is parametrized by an image grid with appearance parameters in each pixel, it is a classical texture map. 
For view-dependent effects, spherical harmonics, Fourier-like basis functions on the sphere, are popular. %
NeRF \cite{mildenhall2020nerf} was the key work that took neural parametrizations from geometry to appearance. 
\extended{
Interestingly, it entangles the geometry and appearance parametrization in a single MLP. 
}
It uses a generic position-and-view-conditioned MLP head to regress view-dependent color volumetrically, \ie anywhere in 3D space.

\extended{
Similarly, learned feature vectors can be attached and rendered, and then passed through a 2D convolutional neural network (CNN) to generate an RGB rendering \cite{sitzmann2019deepvoxels}. 
If they are coupled to the geometry via a UV map of features, they are called neural textures \cite{thies2019neural}. 
Such 2D neural renderings allow for more details but are not multi-view consistent by design. 
}

\subsubsection{Deformation Parametrizations}

This section discusses explicit deformation parametrizations. 
Like appearance, deformations are usually attached to the geometry. 
\revision{In practice, applying deformation models depends on their \emph{direction}, which greatly influences the method design.} 
The definitions in Sec.~\ref{subsec:deformations} are \emph{backward} deformation models where, in order to determine the deformed geometry $g_d$, we first pick a point $\mathbf{x}$ in the deformed state, apply the deformation model $d$ to get to the reference state, and then query $g$: $g_d(\mathbf{x}) = g(d(\mathbf{x}))$. 
This warps the deformed state into the template state, which is common for volumetric representations (\eg in ray tracing).  %
If we instead turn the composition around and first query $g$ at reference point $\mathbf{x}$ and then deform the resulting point to the deformed state, we obtain a \emph{forward} model: $g_d(d^{-1}(\mathbf{x})) = g(\mathbf{x})$. 
Such models are useful for surface representations (rasterization), where one first selects a point $\mathbf{x}$ on the reference surface $g$ and then deforms it to $d^{-1}(\mathbf{x})$.
\revision{We first discuss forward models in the following, and end with backwards models for volumes.} %

\noindent\textbf{Physics Simulation.}
The most accurate way to model deformations is by imposing the true physical laws that govern an object's behavior. 
Most methods that use physics as a hard constraint~\cite{kair2022sft} are based on the Finite Element Method (FEM) from mechanical engineering.
In FEM, a surface or volume is represented as a set of triangular or tetrahedral elements connected at nodes. 
For each element, the mass $\mathbf{M}$, stiffness $\mathbf{K}$, and the damping $\mathbf{D}$ matrices are separately built \extended{and grouped as block diagonal matrices} to capture their physical properties, like Young's modulus, Poisson's ratio and shear modulus. 
\extended{For large deformations, the matrices are non-linear functions of the displacement $\mathbf{u}$ but are often pre-computed assuming linear behavior.}
One can spatially discretize the PDE  \eqref{eq:motion} to obtain an ODE in time, allowing for numerical simulation. 
Then, the full dynamical behavior of the object that describes the unknown vertex displacement $\mathbf{u}$ is given by:  %
\begin{align}
    \label{eq:fem}
    \mathbf{M} \mathbf{\ddot{u}} +  \mathbf{D} \mathbf{\dot{u}} + \mathbf{K} \mathbf{u} = \mathbf{f}.
\end{align}
Despite being ideal in principle, physics simulation is difficult to model completely and to implement, and is computationally expensive. %
\revision{Thus, the vast majority of 3D reconstruction works use non-physical approximations, which we discuss next.}

\noindent\textbf{Template Offsets.} 
A simple deformation model consists of per-vertex offsets of a template, which is particularly popular due to recent methods trained on general image-collections \cite{cmrKanazawa18}. 
As per-vertex offsets are severely underconstrained, they are often combined with soft deformation priors; see Sec.~\ref{sec:priors}. 

\noindent\textbf{Skinning.} 
As deformations tend to be spatially smooth, it is common to \emph{skin} a detailed template geometry to a coarser \revision{graph embedded in 3D} (whose parameters  are thus the deformation parameters) by specifying skinning weights. %
When deforming the coarse graph, its deformations are transferred to the detailed template by interpolating according to these weights. 
Embedded graphs \cite{sumner2007embedded} are common for general objects and skeleton skinning (itself based on a kinematic chain) is common for category-specific models~\cite{SMPL:2015,MANO:SIGGRAPHASIA:2017,ZuffiSmal2017}. %

\noindent\textbf{Linear Subspace Models.} 
Instead of deforming a single template, linear subspace models linearly combine a limited number of \revision{basis deformations%
} to obtain the deformed geometry. %
The coefficients of this combination are often globally constant across space. %
\extended{
To model the non-linear behavior of real objects better, spatially varying, local coefficients are sometimes introduced. %
}
This kind of parametrization \extended{where the shape is restricted to lie on a low-rank space spanned by deformation modes} 
is common in NRSfM \cite{Bregler2000}. 
Similar to shape space, the low-rank assumption can equivalently be made on the trajectory or force space. 
Linear subspaces are sometimes used for the underlying skeletons in skinned models~\cite{SMPL:2015,MANO:SIGGRAPHASIA:2017,ZuffiSmal2017}. 
Models that parametrize a low-dimensional space, especially by linear combinations of some (usually fixed) basis, are called \emph{parametric models}. %

\noindent\textbf{3D Morphable Models (3DMMs).}  
More sophisticated versions of linear models where the basis is built from collections of 3D scans via statistical methods (\eg PCA) are called 3DMMs. 
They are, thus, category-specific and most popular for faces. %
They factorize deformations into independent identity and expression parameters. 
In addition, they often include appearance. 
Most morphable models work on a mesh geometry and are created via PCA, which is linear in the instance-specific coefficients.  
Recently, there are attempts to learn volumetric neural morphable models that are non-linear in the  instance-specific latent code; see  %
Sec.~\ref{sec:faces_implicit_morphable_models}. 

\noindent\textbf{Volumetric Deformations.}  
Implicit geometry parametrizations tend to use volumetric backward deformation models to avoid the need for directly accessing the surface. 
The earlier work Neural Volumes~\cite{Lombardi2019neuralvol} uses an explicit mixture of affine warps, while recent neural-rendering methods use an MLP parametrization. 
In the fully non-rigid case, such an MLP can output an offset per point in 3D  space~\cite{tretschk2021non}, while more articulated deformations benefit from an $\mathit{SE}(3)$ output~\cite{park2021nerfies}. 
Since these are fully unregularized (up to the smoothness of the MLP and soft deformation priors), there are also first attempts to extend skinning to the volumetric case~\cite{chen2021snarf,yang2022banmo}. %

\subsubsection{Camera Parametrizations} 
The definitions of camera intrinsics and extrinsics imply direct, natural parametrizations. 
\revision{For example, a 3D translation can be trivially parametrized by a 3D vector.} 
\revision{Solely camera rotations, which are elements of the 3D rotation group $\mathit{SO}(3)$, inherently cannot have a natural (smooth, unique, without boundary) parametrization (this is ultimately due to the universal cover of $\mathit{SO}(3)$ being a \emph{double} cover by $\mathit{SU}(2)$).} 
Common parametrizations are Euler angles, axis-angle, quaternions, and rotation matrices. 
We refer to Zhou \etal~\cite{zhou2019continuity} for details. %
The distorted ray directions caused by lens distortions can be parametrized well by correctives following, for example, the Brown-Conrady model~\cite{brown1966lenses}. %
\extended{
These models typically reproject the distorted image into an undistorted pinhole camera. 
Inverting them to instead obtain corrected ray directions for the distorted image is non-trivial~\cite{park2021nerfies}. 
}

\subsubsection{Large-Scale Image Collections}\label{sec:fundamentals-data-driven}
While it is tractable for temporal sequences and small-scale image collections~\cite{yao2022lassie} to directly optimize for the reconstruction parameters of the scene, this becomes impractical for large-scale image collections with thousands of images. 
Instead, the scene-specific parameters $\theta_s$ are output by a meta-reconstruction function $\theta_s = f_\theta(s)$ (called \emph{data-driven prior}) that accumulates generalizable reconstruction knowledge about the image collection it is fit on. 
In practice, this data-driven prior is typically a neural network (specifically, a CNN in the case of input images) that regresses the scene-specific parameters of its input scene. 
\extended{
Instead of scene-specific parameters $\theta_s$, we optimize for the parameters $\theta$ of the data-driven prior, \eg the weights of the CNN. 
Thus, it becomes feasible to optimize over large image collections where each image might be only seen in a handful of training iterations. %
}

\subsection{Rendering: Connecting 3D and 2D }\label{sec:rendering-into-2d}
\revision{In the reconstruction setting, we are provided with 2D input data which we need to relate to the 3D model. 
To that end, rendering is crucial as it allows us to extract 2D information from the 3D model. 
Given the scene decomposition consisting of lights, material, and (deformed) geometry, a virtual camera generates a 2D observation of the 3D world in the rendering process. } 
Rendering is the computational model of the physical light-collecting process of a camera.

\noindent\textbf{Rendering.}
\revision{Works on reconstruction employ a small set of standard rendering techniques.} 
Explicit geometry parametrizations like meshes or point clouds (see Sec.~\ref{sec:geometry_paramtrization}) are typically rendered using \emph{rasterization}, which projects each geometric primitive (\eg triangle or point) using a virtual camera. 
If instead an implicit or volumetric geometry parametrization like an MLP or a voxel grid is used, \emph{ray tracing} is \revision{typically} applied. 
For each pixel of the virtual camera, it traces a ray into the scene, trying to hit geometry. 
For surface geometry functions, surface rendering \revision{can be} used~\cite{DVR,sitzmann2019srns}, which picks the first surface along the ray as the point to be rendered, while volumetric geometry functions \revision{can} use volume rendering, which accumulates geometry and appearance along the ray \cite{mildenhall2020nerf,Lombardi2019neuralvol}.

\noindent\textbf{Inverse Rendering}
is the inverse operation of rendering, \ie recovering the intrinsic components (geometry, material, illumination, and deformations) of a 3D or 4D scene from images.
To that end, we can exploit (forward) rendering for \emph{analysis by synthesis}, where we obtain the 3D or 4D scene reconstruction (analysis) \emph{by ensuring that it can render (synthesize) the 2D input}. 
\extended{
This is often done iteratively: we synthesize data from our current analysis, compare it with the input, and then update the analysis to better match the input. 
}
(This STAR also covers methods that primarily use 3D supervision at training time and hence do not follow the analysis-by-synthesis paradigm.) 
\extended{
Analysis by synthesis is an increasingly popular approach for monocular non-rigid reconstruction due to the success of differentiable rasterizers~\cite{liu2019softras} and volumetric rendering~\cite{mildenhall2020nerf}. 
}

\noindent\textbf{Differentiable Rendering.} 
Rendering is naively not differentiable and hence prevents gradients from propagating from the image loss to the 3D model. 
In the simple case of point-based rendering, bilinear interpolation of the input image provides gradients to each 3D point \cite{tewari17MoFA}. 
Several works introduce methods that make mesh rasterization differentiable \cite{loper2014opendr,kato2018renderer,liu2019softras}. %
For ray tracing, differentiable surface rendering is challenging because determining the surface intersection is not naturally differentiable~\cite{DVR,sitzmann2019srns}. 
However, differentiable rendering of a volumetric scene is rather straightforward with volumetric rendering because no surface needs to be determined~\cite{Lombardi2019neuralvol,mildenhall2020nerf}, which also provides a workaround for differentiable surface rendering~\cite{wang2021neus}. 

\subsection{Data Terms: Ensuring Consistency With the Input}\label{sec:dataterms}
Now, we turn to the inverse-problem aspect of the reconstruction problem. 
\extended{
Unlike previous sections, this section concerns itself with real-world observations and hence computer vision. 
}
We require data terms that fit the model to the input data by encouraging consistency between the reconstruction and the input. 
When provided as input, consistency is usually easy to obtain with camera extrinsics and intrinsics, a template geometry, boundary points, timestamps, or a texture: we simply set the parameters of our geometry parametrization, for example, to the template geometry. 
The consistency is therefore ``hard'' in these cases. 
Other inputs, which lack such a nice correspondence to the parametrization, are more difficult and typically consistency is merely ``soft'', \ie encouraged (but not enforced) via losses. 
\subsubsection{\revision{Common Data Terms}}
\revision{Since supervision most often happens via 2D input data, we need to render our model into 2D and then compare to the input data.} 
In the case of RGB image input, typical \revision{photometric} losses are $\ell_1$ or $\ell_2$ losses, and, in recent years, perceptual losses like LPIPS~\cite{zhang2018perceptual}. 
Similar to these appearance-focused losses, 2D object segmentations are usually easy to obtain from the model geometry and can then be compared to input segmentations masks, which tends to help with coarser mismatches. 
Correspondences across time can also be extracted from the deformation parametrization and then be fitted to \revision{dense} 2D input correspondences (optical flow) or 3D input correspondences (scene flow~\revision{\cite{zhai2021optical}}). 
Sparse 2D correspondences from feature point tracking (\eg via SIFT~\cite{lowe2004distinctive}) are sometimes used too, as they help with reconstructing large deformations. 
When correspondences across images---and not images themselves---are the input to the method, the latter falls into the category  Non-Rigid Structure-from-Motion (NRSfM). 
In analytical Shape-from-Template (SfT) methods, matches between the template and input image are provided instead of an RGB image.
In the case of image collections, we might be given certain model parameters as input (\eg morphable model parameters). 
The estimated deformation parameters and the input parameters can then be compared in a similarity loss. 
\subsubsection{\revision{Other Data Terms}}
\extended{
\newline
\noindent\textbf{Learned Features.} 
}
Beyond appearance-based matching, some recent methods exploit \emph{2D-3D consistency of learned features}. 
Thus, ViSER~\cite{yang2021viser} learns them from scratch, BANMo~\cite{yang2022banmo} uses Continuous Surface Embeddings~\cite{neverova2020cse}, and LASSIE~\cite{yao2022lassie} uses DINO~\cite{caron2021emerging} features. 
\extended{
These features are computed for each pixel and then lifted to the 3D geometry. 
}
A rendering loss encourages consistency between the features attached to the geometry and the image features. 
Unlike appearance, these features can more readily incorporate local and global context, and hence provide more information about larger-scale mismatches. 
\extended{
They are therefore also more robust to challenging texture (like small repetitive patterns or textureless surfaces). 
}
\extended{
\newline
\noindent\textbf{Adversarial Losses.} \label{sub_sub_sec:other_data_terms}
}
Typically used in a generative setting, \emph{3D-aware GANs} use a 3D representation in the generator to render 2D images. 
A 2D discriminator then encourages those images to resemble the distribution of some given set of input images. 
Since this imposes consistency with input data in a looser fashion, it leaves the generator more freedom to hallucinate finer details that look plausible, instead of having to reconstruct the input exactly. 
To apply such GANs to reconstruction, they are first trained in a generative manner and then need to be inverted at test time, \ie the right latent code for some input image needs to be determined. 
See \emph{2D Supervision} in Sec.~\ref{sec:faces_implicit_morphable_models} for more details. 
\extended{
\newline
\noindent\textbf{Estimated Depth.} 
While monocular reconstruction does not directly allow for depth input, per-frame estimates of the visible geometry offer major guidance to the reconstruction. 
Recent progress in monocular depth estimation \cite{LuoVideoDepth2020} has enabled monocular methods to use decent depth estimates in place of true input depth \cite{xian2021space}. 
}

\subsection{Challenges: What Makes the Problem Difficult}\label{sec:difficulties}

Unfortunately, there are challenges on multiple levels when trying to find a 3D reconstruction. 
In this section, we discuss a variety of them and mention some potential solutions. 
The next section focuses on the main challenge: the underconstrained nature of the 3D reconstruction problem and priors to tackle it. 

\subsubsection{Inherent Challenges} 

Several issues are inherent to the problem formulation and cannot be solved by any amount of data.

\noindent\textbf{Challenge: Occlusions.}
\revision{Especially} in the monocular setting, an object may self-occlude\revision{, \eg due to the movement of human body parts and folding of cloths, or become occluded due to an external object.} %
We thus have no input information about its current state and the data terms are unavailable, \apriori~preventing reconstruction. %
This is a root challenge in monocular reconstruction.
\begin{wrapfigure}{r}{0.2\textwidth}
    \includegraphics[width =
    \linewidth] 
    {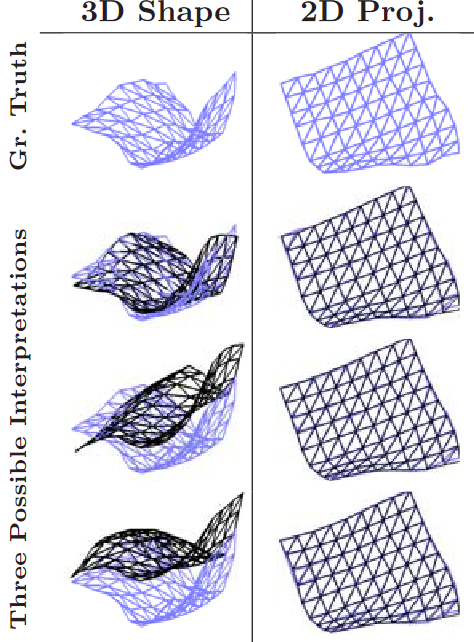} 
    \caption{2D-3D depth ambiguity. Image adapted from \cite{moreno2010exploring}. 
    }
    \vspace{-0.7cm}
    \label{fig:depth_ambiguity}
\end{wrapfigure}
\newline
\noindent\textit{Solution: Regularization.} 
Soft priors (Sec.~\ref{sec:priors}) are used to fill-in missing information. 
We note that many methods do not explicitly consider occlusions and instead apply the same prior to occluded and visible regions. %

\noindent\extended{\textbf{Non-Rigid Loop Closure.} 
A special case of the occlusion challenge is recognizing whether a hidden part that becomes visible has been seen before or not. %
Due to possibly \emph{non-rigid} deformations of the part, this is an even harder problem than loop closure in the static case, which is caused by an incorrect estimation of \emph{rigid} deformations. 
Most methods fail to recognize sameness, which results in ghosting. 
}

\noindent\textbf{Depth Ambiguity.} 
Another root challenge is the lack of depth when using monocular visual measurements. 
All points on the (optical) ray are projected to the same image point, leading to depth ambiguities: 
Different 3D geometries can lead to the same 2D projection in image space, as Fig.~\ref{fig:depth_ambiguity} shows. Several special cases arise from this, which we discuss next.

\noindent\extended{\textbf{Depth-Scale Ambiguity.} 
This is a special case of depth ambiguity. 
When using a pinhole camera model, there is no principled way of determining the absolute distance (depth) of an object from the camera and therefore its size (scale). 
For example, moving an object to twice its depth while doubling its scale will result in the same, indistinguishable rendering. 
}

\noindent\textbf{Geometry-Appearance Ambiguity.} %
Correctly attributing fine-scale image details to geometry versus appearance (\eg in the case of fine wrinkles and textures) is \apriori~ambiguous. %

\noindent\textbf{View Dependence.} 
Arbitrary view-dependent appearance makes it possible to attribute the image formation to almost any geometry with view-dependent appearance. 
\newline
\noindent\textit{Few Degrees of Freedom (DoFs) for View Dependence.} 
The DoFs are typically limited by using only spherical harmonics up to degree three or using a very small MLP for NeRF-style view dependence.

\noindent\textbf{Focal Length.} In image collections, the actual focal length typically varies for each image. 
However, a single image is insufficient to estimate the focal length. %
\newline
\noindent\textit{Fix to Arbitrary Value.} 
In practice, it is common to approximate the focal length by a fixed value across all images. %

\noindent\extended{\textbf{Force-Material Ambiguity.}
The third root challenge occurs when using physical deformation models. 
There is an ambiguity between external forces applied to an object and the object's material stiffness.
Is the force weak and the material very compliant or is the force strong and the material more resistant?
}

\noindent\extended{\textbf{Lack of 3D Correspondences.} 
Establishing geometry correspondences across time or different instances is the last root challenge. 
\newline
\noindent\textit{Brightness Constancy.} 
Many methods implicitly assume temporally constant or smooth appearance to establish correspondences, relying on brightness constancy for well-textured surfaces. 
In practice, however, shadows, including self-shadows in the case of human bodies or hands, can severely affect the appearance of a point across time. 
}

\subsubsection{Parametrization Challenges}

\noindent\textbf{Topology Change.} Topology changes occur when a surface starts \revision{merging or splitting apart}. %
They are difficult to handle because they need to be detected and then accounted for in the parametrization. 
\newline
\noindent\textit{Discarding Correspondences.} 
Current methods that handle topology changes do so by not having a single, consistent geometry parametrization across time, thereby discarding correspondences. 

\noindent\textbf{Discretizing Losses.} %
Loss functions, consisting of data terms or priors, are in most cases easier and more intuitive to formulate in a continuous manner. 
\extended{
For example, a smoothness loss may be described by a small gradient. 
}
Discretizing such continuous formulations onto discrete parametrizations (like meshes) is non-trivial, and one continuous formulation may give rise to different discretizations with different optimization behavior and formal guarantees. 
\extended{
A \emph{consistent} discretization is preferred, such that the discrete solution converges to the continuous one when the resolution is increased. 
}

\noindent\extended{\textbf{Initialization.} 
The parameters of the reconstruction need to be initialized to sensible values, as the optimization can otherwise easily get stuck in local minima. 
This also includes the weights of neural networks. 
}

\noindent\textbf{Camera-Rigid-Motion Ambiguity.} 
Without static background, the rigid motions of the camera and object are ambiguous. %
\newline
\noindent\textit{Assume Static Camera or Arbitrarily Factorize.} 
Small camera motion can be modeled as a rigid transform of the object under a static camera, especially for SfT. %
Some methods, in particular in NRSfM, separately account for the camera movement, and output the camera rotation and translation. %

\noindent\textbf{Identity-Deformation Ambiguity.} %
When reconstructing faces or humans, it is often desired to factorize the deformations of the template geometry into identity-specific (invariant per person) and pose-specific (varying over time) components. 
However, image collections often contain only one image per person, making such a factorization \apriori~ambiguous.

\subsubsection{\revision{Data Acquisition} Challenges}

\noindent\textbf{Background.} Static background is visible when recording. %
\newline
\noindent\textit{Ignoring, Partial Modeling, Full Modeling.} 
Most methods remove it at the input level via background subtraction (for static cameras) or image segmentation. %
Especially in the case of image collections, the background is sometimes kept in the input but then the reconstruction either ignores it or reconstructs it only badly~\cite{wu2020unsup3d}. 
Recently, a handful of NeRF-based methods properly reconstruct the static background as well, see Sec.~\ref{sec:general-nerf}. %

\noindent\textbf{Motion Blur.} When recording a fast-moving scene or moving a camera quickly, a pixel may collect color from different points of the scene within the short time frame when the sensor is active for the current frame. 
This leads to so-called \emph{motion blur}. 
\newline
\noindent\textit{Filtering.} Motion blur is difficult to account for in a model and is hence seldom modeled. 
Instead, blurry images tend to be discarded or heuristically de-blurred during pre-processing. 
\extended{Alternatively, event cameras can be used in place of RGB cameras as they offer very high temporal resolution.}

\noindent\textbf{Lens Distortions.} 
Lens distortions can be decently well estimated for a temporal sequence, while image collections are too severely underconstrained, similar to the focal length ambiguity. 
\newline
\noindent\textit{Undistortion.} In most works that consider lens distortions at all, the forward models are applied to obtain an undistorted pinhole-camera image, although some works instead optimize for the corrected ray directions of the distorted image~\cite{park2021nerfies}. 

\noindent\textbf{Noise.} Noise in the input (\eg RGB images, camera parameters, drifting correspondences) is, by its nature, not easy to detect. 
Even small noise can have negative impacts on the optimization. 
\extended{
When developing a method, it is therefore common to first experiment with synthetic data to avoid these kinds of imperfections. 
}
\newline
\noindent\textit{Correcting Input Estimates.} Some estimated input parameters $\theta_\text{est}$, especially camera extrinsics and intrinsics, tend to be slightly incorrect (noisy) in practice. 
If this noise is too severe, it is possible to optimize for corrective parameter offsets $\Delta \theta$: $\theta_\text{correct} = \theta_\text{est} + \Delta \theta$, where $\Delta \theta$ is usually kept small via an $\ell_1$ or $\ell_2$ loss. 
\newline
\noindent\textit{Robust Losses.} 
Some losses are more robust to input outliers, \eg a Huber loss or an $\ell_1$ loss more so than an $\ell_2$ loss.

\noindent\textbf{On-Camera Processing.} On a very practical level, modern cameras process images, which might lead to undesirable effects (\eg automatic white balancing or varying gamma correction, as well as missing color calibration for image collections can lead to varying measurement results of the exact same real-world color). 
\newline
\noindent\textit{Ignoring It, Modeling It or Turning It Off.} 
Oftentimes, the parameters of these operations are not accessible to the end user, and hence are ignored. 
They can also be estimated afterwards if these kinds of processing cannot be turned off when recording.

\subsection{Soft Priors: Making the Problem Tractable}\label{sec:priors}

A lot of information is lost during the image formation process because, at any time step, we only obtain a \emph{visual} measurement under \emph{one} viewing angle of any \emph{visible} surface point. %
We thus need to fill-in this lost information. 
In addition, we need to prevent undesirable local minima to stabilize the optimization in practice. 
As with other inverse problems, we therefore seek to constrain/regularize the solution space of the shape reconstruction with prior assumptions, ideally forcing the existence and uniqueness of a solution. 
We note that Sec.~\ref{sec:parametrization} discusses hard priors.

\subsubsection{Geometry Soft Priors}\label{sec:geometry-priors}
Geometry priors are solely spatial, \ie they only act on a geometry by itself, regardless of whether it was obtained through deforming some reference geometry or not. 
Typical priors include spatial smoothness, where, for example, a Laplacian loss or a loss on the normals of the geometry encourage locally smooth geometry, and parametrization-specific priors that discourage local minima, like a loss encouraging mesh edges to be short. 
Some methods may exploit symmetry constraints.

\subsubsection{Deformation Soft Priors}

Most deformation priors act on the final shape by introducing one or more reference geometries with respect to which they regularize the current one. 
Alternatively, they can rely on parametric models and regularize their parameters.

\noindent\textbf{Metric-Based Priors.}
Many spatial deformation priors approximate the underlying physical properties of non-rigid objects. 
They are defined locally and follow from the measures of deformation (see Sec.~\ref{subsec:deformations}). 
Physically plausible deformations are assumed to preserve different metric quantities such as lengths, angles, and areas on the surface of the geometry. 
\extended{
Deformation models differ by the choice of such invariant metrics, which subsequently determines how constraining the model is and its applicability to different classes of objects. 
}
The measures are defined on a deformed geometry $g_d$, with respect to a reference geometry $g$. %
Ideally, the reference geometry should be  a physical rest pose so that the deformations are not just geometrically but also physically meaningful. 
\extended{
Then the geometric priors, commonly known as \textit{deformation constraints}, can be defined as surface mappings formulated in the local coordinate frames. 
}
\revision{Then with the definitions from Sec.~\ref{subsec:deformations} and Fig.~\ref{fig:deformable_dynamics},}
\begin{itemize}
\item
\textit{Isometric deformation maps} preserve the geodesic distance between any two points on the surface (\eg consider paper): %
\revision{
\begin{align}
G(\mathbf{x}(\mathbf{m},t)) = G(\mathbf{x}(\mathbf{m},0))
\end{align}
}
It only allows surface bending, but not stretching or shearing. 
A simpler alternative is \textit{inextensibility}, which preserves the \emph{Euclidean} distances instead. 
For real-world extensible objects, \textit{quasi-isometry} prevents large stretching or shrinking and can be implemented as As-Rigid-As-Possible (ARAP) prior~\cite{sorkine2007rigid}.
\item
\textit{Conformal maps} preserve local angles on the surface: %
\revision{
\begin{align}
G(\mathbf{x}(\mathbf{m},t)) = \lambda(\mathbf{m}) G(\mathbf{x}(\mathbf{m},0))
\end{align}
}
Conformality is weaker than isometry and allows for stretching 
(\eg an expanding balloon). 
\item
\textit{Equiareal maps} preserve area on the surface and lead to:
\revision{
\begin{align}
\det(G(\mathbf{x}(\mathbf{m},t))) = \det (G(\mathbf{x}(\mathbf{m},0)))
\end{align}
}
Isometry is equivalent to conformality and equiareality together. %

\extended{
\item
\textit{Volume-preserving maps} preserve local volumes between undeformed and deformed shapes. With the local frame at a point given by $ (\mathbf{t}_1,\mathbf{t}_2, \mathbf{t}_3)$, we have: 
\begin{align}
    {\lVert \overline{\mathbf{t}}_1 \cdot (\overline{\mathbf{t}}_2 \times \overline{\mathbf{t}}_3) \rVert}^2 = {\lVert \mathbf{t}_1 \cdot (\mathbf{t}_2 \times \mathbf{t}_3) \rVert }^2.
\end{align}}
\end{itemize}

\begin{table*}[!t] 
\small{ 
\begin{tabular}{m{0.11\linewidth} | m{0.20\linewidth} | m{0.21\linewidth} | m{0.19\linewidth} | m{0.15\linewidth}} 
\toprule 
\textbf{Object Type} &  \textbf{Solving Strategy} & \textbf{Data Term} & \textbf{Deformation Prior} & \textbf{Temporal Coherency} \\  \midrule 

curve: {\tiny \cite{gallardo2020shape}} 
\newline volumetric: {\tiny \cite{parashar2015rigid, fuentes2018deep, Yu2015}}
\newline thin-shell: {\tiny \cite{kair2022sft, casillas2021isowarp, shimada2019ismo, FuentesJimenez2021}}
& analytical: {\tiny \cite{casillas2021isowarp,casillas2019equiareal, chhatkuli2016stable, Bartoli2015}} 
\newline energy-based: {\tiny\cite{kair2022sft, ozgur2017particle, malti2017elastic, Yu2015}} 
\newline neural (object-specific): \newline {\tiny\cite{fuentes2018deep, shimada2019ismo, Pumarola2018, Golyanik2018}}
\newline neural (generic): \newline {\tiny\cite{FuentesJimenez2021, shimada2019ismo}}
& template-image warp: \newline{\tiny \cite{casillas2021isowarp,Ngo2015} }
\newline per-pixel intensity: {\tiny \cite{kair2022sft}}
\newline per-vertex intensity: \newline{\tiny\cite{ozgur2017particle, Yu2015}}
\newline shading cue: {\tiny \cite{liu2017better}}
\newline pre-trained: {\tiny \cite{shimada2019ismo, FuentesJimenez2021}}
\newline surface micro-structure: {\tiny\cite{habermann2018nrst}}
& isometry: {\tiny \cite{casillas2021isowarp, shimada2019ismo, chhatkuli2016stable}} 
\newline conformality: {\tiny\cite{Bartoli2015}}
\newline equiareality: {\tiny \cite{casillas2019equiareal}}
\newline elasticity: {\tiny \cite{kair2022sft, ozgur2017particle, malti2017elastic}}
\newline ARAP: {\tiny \cite{FuentesJimenez2021, Yu2015}}
\newline Laplacian: {\tiny\cite{ngo2015template}}
\newline low-rank: {\tiny\cite{tretschk2020demea}}
& present: {\tiny\cite{kair2022sft, Yu2015}} 
\newline not present: {\tiny \cite{shimada2019ismo, casillas2021isowarp, FuentesJimenez2021}}\\
\bottomrule 
\end{tabular} 
} 
\caption{Overview and classification of Shape-from-Template methods. %
}
\label{table:overview_sft}
\end{table*} 

\noindent\textbf{Other Reference-Based Priors.} 
In addition\extended{~to these intrinsic priors, which are invariant under local rigid transforms}, we may also favor stricter closeness to the template. 
For example, we may encourage the template offsets to be small\extended{, in which case the template acts as a mean shape}, or, in the case of skeleton-based deformations, the angles of the joints to stay close to the rest pose or within a certain range. 
Furthermore, we can encourage the coefficients or latent codes of a parametric model to be close to zero.

\noindent\textbf{Temporal Priors.}
Unlike single images, videos provide an additional temporal dimension that can be leveraged as a prior. 
Assuming that images are sampled at high enough frame rates, deformation states that are temporally close are, in general, similar to each other. 
We can impose this prior knowledge about temporal smoothness using similarity losses between the deformations of the time steps in question. 
Alternatively, when reconstructing a temporal sequence of multiple time steps, it can be useful to optimize in a sequential manner, starting with reconstructing $t=0$ and then continuing step-wise for $t>0$. 
In particular, the deformation parameters $\theta^t_d \subset \theta$ at time $t$ are often initialized from the previous time step: $\theta^t_d = \theta^{t-1}_d$, which is usually referred to as \emph{tracking}. %

\subsubsection{Appearance Soft Priors}

In the case of texture-less surfaces, a smoother change in appearance or more explicit priors about lighting and reflectance maybe employed to aid reconstruction using shading cues. 
Nonetheless, soft priors are only rarely applied for appearance. 
3DMMs \cite{blanz1999morphable} and similar statistical models can encourage the appearance parameters to stay close to the estimated prior parameter distribution, which is often assumed to be normal distributed.

\extended{
\subsubsection{Material-Force-Elasticity}
Following the continuum mechanics of deformable objects (Sec.~\ref{subsec:deformations}), physics-based approaches build additional force and elasticity priors. %
Among the external forces, gravity can be assumed as a known force. %
Apart from this, most priors concern the material parameters. 
An implicit prior in many SfT and NRSfM methods is the material homogeneity of the deforming object, and consequently, identical geometric or elastic deformation constraints are specified at all points on the surface. 
For specific object categories such as cloth or paper, the material density and surface thickness can be set to known values. 
If a template is supplied in addition to the above, the mass at each node can be computed using the triangulation of the surface.  
For isotropic objects whose material response is independent of the direction of the applied deformation, we may use Young's modulus and Poisson's ratio. %
Real-world values are available for common materials like rubber or clothes. 
From these parameters, one can derive the stretching and shearing stiffness for deformable volumes and additionally bending stiffness for deformable surfaces.  
Methods either use known stiffness values~\cite{ozgur2017particle} or initialize them from other data and refine them on a specific instance~\cite{kair2022sft}. 
}

\subsection{Optimization: Finding the Right Parameters}\label{sec:optimization}
Once we have set up a solution parametrization with a set of parameters $\theta$ and a loss function $\mathcal{L}(\theta)$, containing data terms and priors, we can finally determine the best set of parameters $\theta^*$ as the solution to the 3D reconstruction problem:
\begin{align}\label{eq:optimisation}
    \theta^* = \argmin_\theta \mathcal{L}(\theta).
\end{align}

There is a wide variety of optimization techniques that is used in the literature for this problem that is virtually always highly non-convex. %
While neural methods are almost exclusively optimized via gradient-based techniques (using $\frac{\partial \mathcal{L}}{\partial \theta_i}$ for the $i$-th parameter) that start from an initial guess $\theta_\text{init}$, other methods also employ gradient-free optimization (such as simulated annealing, particle swarm optimization or evolutionary policies). 
A detailed discussion of these techniques, however, is outside the scope of this section. 

\noindent\extended{\textbf{Local vs. Global.} 
Most methods optimize $\theta$ all at once, \ie \emph{globally} or \emph{offline}. 
However, especially when tracking in a real-time context, it is sometimes advantageous to only optimize the parameters $\theta_t$ of the current time $t$. 
This \emph{local} or \emph{online}, causal optimization keeps the processing time per frame constant. 
}

\noindent\extended{\textbf{Coarse to Fine.} 
To smoothen the optimization landscape, some methods, for example \cite{ZuffiSmallr2018, li2021coarsetofine}, factorize the deformations into a hierarchy of coarse and fine deformations. 
Then, the coarse deformations can help in regularizing the problem, while the fine deformations allow to add details. 
}

\section{State-of-the-Art  Methods}\label{sec:works} 
The main axis along which we organize our discussion is the object category that is to be reconstructed. 
After discussing methods for monocular 3D reconstruction of general objects (Sec.~\ref{sec:general-objects}), we describe the state of the art of methods specialized for the human body (Sec.~\ref{subsec:humans}), faces (Sec.~\ref{subsec:face_comp}), hands (Sec.~\ref{subsec:hands}), and animals (Sec.~\ref{sec:animals}). 
We discuss methods using event cameras in Sec.~\ref{sec:discussion}. 

\subsection{General Objects} \label{sec:general-objects} 

We first discuss the established fields of SfT (Sec.~\ref{sec:sft}) and NRSfM (Sec.~\ref{sec:nrsfm}) before moving on to few-scene reconstruction methods that rely neither on template nor correspondences as their core assumption. 
We split these into NeRF-like neural methods (Sec.~\ref{sec:general-nerf}) and others (Sec.~\ref{sec:fewscene}). 
We finally turn to data-driven approaches (Sec.~\ref{sec:manyscene}) that work on large-scale image collections. 

\subsubsection{Shape from Template (SfT)} \label{sec:sft}
\begin{figure}
    \centering
    \includegraphics[width=\linewidth]{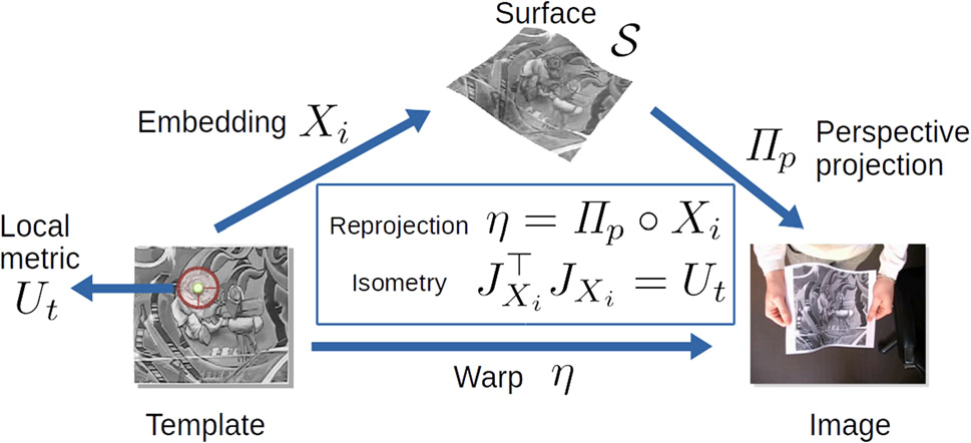}
    \caption{Differential isometric SfT \cite{casillas2021isowarp,chhatkuli2016stable,Bartoli2015} solves for the reconstruction embedding $X_i$, given the warp $\eta$ (which aligns the template with the input image). Image adapted from \cite{casillas2021isowarp}.
    } %
    \label{fig:isowarp}
\end{figure}
\par
Shape from Template (SfT), or template-based reconstruction, comprises monocular non-rigid 3D reconstruction methods that assume a single static shape or \textit{template} is given as a prior. 
It has been an active research area for two decades \cite{salzmann2007deformable, salzmann2010deformable}. 
The name SfT (not to be confused with Shape from Texture) became common after 2015 due to the eponymous work of Bartoli \etal~\cite{Bartoli2015}. 
We next discuss templates, solving strategies and deformation priors in SfT before describing the state of the art in detail. %

\noindent\textbf{Template.} 
Given a template %
in a reference configuration and a calibrated camera, SfT aims to reconstruct the shape of a deformable object in every frame of a video sequence observing the object. 
\extended{Some human and animal reconstruction methods~\cite{DeepCap, Kokkinos_2021_CVPR} may also rely on a template; however, we discuss them in their respective Sec.~\ref{subsec:humans} and Sec.~\ref{sec:animals}, and focus on general objects here.} 
The template often corresponds to the first frame of the  sequence though it is not always a strict requirement. 
The template can be used as the initial state of a physics simulator~\cite{kair2022sft}, to obtain 3D-2D registration as a basis for reconstruction~\cite{casillas2021isowarp}, and to encode prior knowledge in neural network  weights~\cite{shimada2019ismo, FuentesJimenez2021}. 

\noindent\textbf{Solution Strategies.} 
As shown in Tab.~\ref{table:overview_sft}, SfT methods can be classified as \textit{energy-based}, \textit{analytical}, and \textit{neural-based} approaches.
Energy-based methods \cite{kair2022sft, ozgur2017particle, malti2017elastic, Yu2015, brunet2014monocular, malti2015linear} define a non-convex cost function with photometric consistency as the data term and deformation priors acting as the regularization term.
The energy is typically minimized using iterative optimization methods~\cite{kingma2014adam, more1978levenberg}.
One major challenge for this method class is that the energy landscape is often non-linear, and the algorithm can potentially converge to erroneous local minima.
Therefore, careful initialization is required, and the template is often used as an initial shape~\cite{kair2022sft, ozgur2017particle}; alternatively, the solution from the previous frame may also be used as in~\cite{Yu2015}.
Analytical methods\cite{casillas2021isowarp, casillas2019equiareal, chhatkuli2016stable, Bartoli2015, fayad2011automated} formulate re-projection and deformation constraints as PDEs and provide well-posed analytic solutions in a single step. However, these do not match the accuracy of energy-based methods and require a refinement of reconstructions using iterative methods. 
Recently, neural methods~\cite{FuentesJimenez2021, fuentes2018deep, shimada2019ismo, Pumarola2018, Golyanik2018, tretschk2020demea} have been used to learn image to 3D shape mappings by training deep networks on datasets of deforming sequences. Since, at test time, 3D reconstructions are obtained simply by a single feed-forward pass, they usually achieve a higher runtime performance compared to energy-based approaches. However, they tend to be texture- and template-specific and often have difficulty generalizing to 
unseen shapes. 

\noindent\textbf{Deformation Priors.} 
SfT methods can also be classified according to the type of deformation priors. 
Strong ones (\eg isometry) have been extensively  studied~\cite{Bartoli2015}, 
whereas weaker but more accurate elastic priors are becoming increasingly popular~\cite{kair2022sft,  casillas2019equiareal, malti2017elastic}. 
In early works~\cite{chhatkuli2016stable,Bartoli2015}, registration between template and input image, along with their differential structures and isometric constraints, delivered well-posed problems with unique solutions.  %
More recently, neural networks~\cite{shimada2019ismo, Golyanik2018, Nehvi_2021_CVPR} have been used to favor isometry instead of enforcing it. 
For extensible surfaces such as balloons, conformal geometric prior have been similarly used to obtain families of solutions~\cite{Bartoli2015}.
Casillas-Perez \etal~\cite{casillas2019equiareal} provide a theoretical framework for equiareal SfT using Monge's theory for solving first-order nonlinear PDE and show results on stretched fabrics. 
Parashar \etal~\cite{parashar2019local} use Cartan's theory of connections and moving frames, that offers a generic solution to all local (isometric, conformal and equiareal) deformation models. 
While these geometric priors are only approximate, physically exact stretching and bending priors can be derived from the continuum 
mechanics of elastic objects. 
The approach of Malti \etal~\cite{malti2015linear} relies on linear elasticity to minimize stretching energy under reprojection boundary conditions, which was later extended to constrain the set of spatial forces to be sparse~\cite{malti2017elastic}.
Similarly, {\"O}zgur \etal~\cite{ozgur2017particle} specify stiffness parameters describing the stretching and bending behavior of elastic objects, whereas another method uses isotropic material elasticity (Saint-Venant Kirchhoff model) \cite{Haouchine_2017_CVPR}. 
To model non-linear and anisotropic behaviors of challenging cloth deformations, $\phi$-SfT  \cite{kair2022sft} imposes the elastic model of~\cite{wang2011data} as deformation prior. 

\par
\noindent\textbf{State-of-the-Art Methods.} 
Given the \textit{warp} relating template to the input image and their differentials, analytical methods~\cite{casillas2021isowarp,chhatkuli2016stable,Bartoli2015} formulate the SfT problem in terms of a system of non-linear first-order PDEs, as shown in Fig.~\ref{fig:isowarp}.
These equations depend on the unknown reconstruction embedding $X_i$, uniquely defined with the depth function $\rho$, given the warp $\eta$ and the template’s local metric $U_t$. 
Bartoli \etal~\cite{Bartoli2015} directly solve for depth $\rho$ and its derivatives $\nabla\rho$ as independent variables in the isometric SfT system, not related via differentiation, leading to the non-holonomic solution.
Extending this, Chhatkuli \etal~\cite{chhatkuli2016stable} propose to use the non-holonomic depth’s gradient to recover the surface via integration. 
This strategy is significantly more stable and robust to errors in the warp. 
Alternatively, Casillas \etal~\cite{casillas2021isowarp} propose \textit{isowarp} to improve the warp for the analytic depth solutions~\cite{Bartoli2015}. 
They define a set of warp constraints from the 3D isometry equations, and the resulting warp representation improves the accuracy of reconstructions.
\par
A recent real-time SfT approach by Fuentes-Jimenez \etal~\cite{FuentesJimenez2021}, \ie RRNet-DCT, relies on deep neural networks. %
Its architecture has two neural networks: A segmentation module for pixel-based detection of the template and a  registration-reconstruction module to perform SfT. 
RRNet-DCT is texture-agnostic as it adapts to new texture maps at run-time compared to the authors' earlier texture-specific method, DeepSfT~\cite{FuentesJimenez2021}.
Being an object-specific method that encodes the template into the neural network weights, it is highly accurate, unlike earlier object-generic methods such as IsMo-GAN~\cite{shimada2019ismo}. 
However, both DeepSfT and IsMo-GAN are less generic methods than energy-based methods.
\par
In contrast to the \textit{wide-baseline} analytical and neural methods, the recent \textit{short-baseline} method $\boldsymbol{\phi}$-SfT by Kairanda \etal~\cite{kair2022sft} leverages the temporal consistency across frames. 
\begin{figure}
    \centering
    \includegraphics[width=\linewidth]{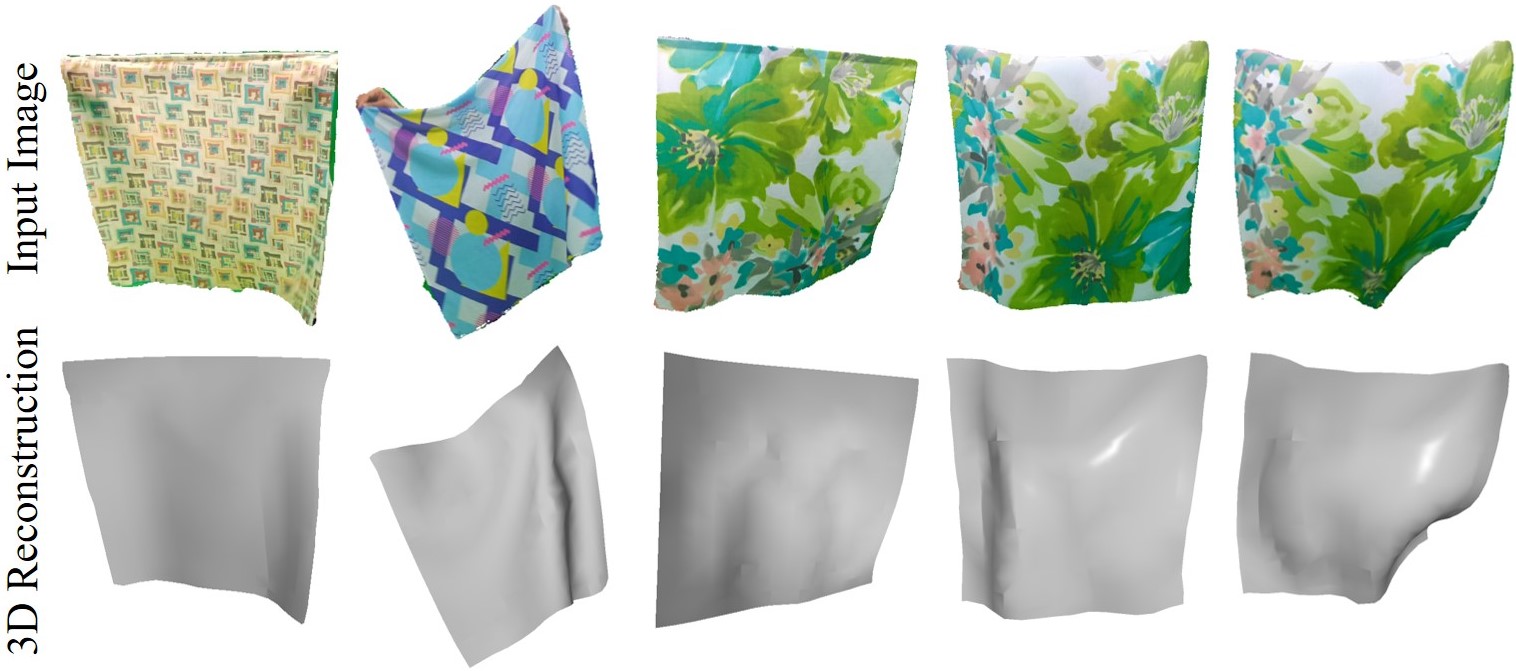}
    \caption{$\boldsymbol{\phi}$-SfT \cite{kair2022sft} explicitly simulates the physical fold formation process. 
    It can thus handle even challenging local folds. 
    Image adapted from \cite{kair2022sft}.
    } 
    \label{fig:PhiSfTResults}
\end{figure}
$\boldsymbol{\phi}$-SfT accounts for 2D observations through physical simulations of forces and material properties. 
They use a differentiable physics simulator \cite{Liang2019} to regularize the surface evolution and to optimize the forces and material elastic properties such as bending coefficients, stretching stiffness and density. 
Following an analysis-by-synthesis approach, a differentiable renderer is employed to minimize the dense reprojection error between the estimated 3D states and the input images; the deformation parameters are recovered by adaptive gradient-based optimization. 
Compared to earlier analysis-by-synthesis solutions with  per-vertex photometric costs~\cite{Yu2015, liu2017better},  $\boldsymbol{\phi}$-SfT's per-pixel approach  uses the full  information in the high-resolution texture map, leading to  accurate reconstruction of challenging local folds; see  Fig.~\ref{fig:PhiSfTResults}.

\noindent\textbf{Datasets.} 
SfT methods require reference templates and image sequences as part of the dataset. 
The template and the respective texture map are generally acquired with SfM~\cite{Yu2015} or an RGB-D camera~\cite{kair2022sft}.
Most works also evaluate on 
real and synthetic datasets that satisfy the assumptions on deformation types of the respective methods. 
We list the real datasets with the most recent ones first: 
\textit{$\boldsymbol{\phi}$-SfT} \cite{kair2022sft}; 
\textit{t-shirt} and \textit{balloon} and \textit{sock} \cite{casillas2019equiareal};
\textit{zooming} and \textit{can} \cite{chhatkuli2016stable};
\textit{face}, \textit{bobby} and \textit{pig} \cite{Yu2015};
\textit{cushion}, \textit{balloon} and \cite{Bartoli2015};
\textit{woggle}, \textit{sponge} and \textit{arm} \cite{parashar2015rigid};
\textit{balloon}, \textit{spandex}, \textit{redchecker} and \textit{cap} \cite{malti2013monocular, malti2015linear}; 
\textit{t-shirt} and \textit{paper} \cite{Varol2012}; and
\textit{face}~\cite{Valgaerts2012}. %
Besides real sequences, a few methods also evaluate on synthetic datasets whose geometries are generated with physics simulation.
Learning-based works~\cite{Golyanik2018, Pumarola2018} often 
train neural networks using lightweight, synthetic training datasets. 
They incorporate various deformations, textures, illuminations and camera poses to ensure generalizability to unseen images. 
\extended{Shimada \etal~\cite{shimada2019ismo} extend the dataset from \cite{Golyanik2018} to include textureless surfaces.}

\par 
\noindent\textbf{Open Challenges.} 
SfT has been successfully applied in the medical domain (\eg to register a preoperative 3D liver model to a laparoscopy image~\cite{espinel2020combining, collins2016robust, koo2017deformable}); 
however, practical applications are still limited, and we list the reasons for the same. %
A set of problems not yet attempted in the field include: background reconstruction, changing object topology, multiple deformable objects and severe self-collisions. 
Next, 
it is common to evaluate SfT on datasets with smooth deformations (\eg the \textit{t-shirt} and \textit{paper} sequences \cite{Varol2012}). 
The physics-based \textit{$\boldsymbol{\phi}$-SfT} approach~\cite{kair2022sft} supports challenging local folds but fails to capture small and frequent wrinkles. 
Implicit surface representations have not yet been studied in the context of the classical SfT problem, but we believe they have potential as in many other sub-fields. 
Non-learning methods use triangular meshes with a fixed resolution, while learning-based SfT techniques have fixed output sizes. 
Despite offering unique and closed-form solutions---as registration with a template is fundamental to analytical SfT---errors in warps propagate to 3D and limit the reconstruction accuracy. 
SfT methods commonly operate on individual images, and although they provide 3D correspondences with a template, the reconstructions can suffer from frame-to-frame jitter. 
Deviating from this, a few works   \cite{Yu2015,habermann2018nrst} employ an explicit temporal regularization term and $\boldsymbol{\phi}$-SfT outputs temporally smooth surfaces %
owing to simulation.
Besides, exploring joint optimization over multiple frames is promising and tractable for SfT due to advances in GPUs. 
\par

\subsubsection{Non-Rigid Structure from Motion (NRSfM)} \label{sec:nrsfm}

Whereas SfT uses the information present in a single image to deform the template, NRSfM relies on motion and deformation cues for 3D recovery of deformable  surfaces \cite{Bregler2000} and is more generally applicable than SfT. %
The input to NRSfM are 2D point tracks across multiple images, also called \textit{measurements} or \textit{measurement matrices}, and the output is a set of per-view camera-object poses and 3D shapes. 
This section, similarly to the entire STAR, focuses on dense NRSfM methods, which operate on (per-pixel) densely tracked 2D points. 
During dense point tracking with optical flow or video  registration methods \cite{Garg2013V}, a single keyframe is selected, and the 3D points visible in it are tracked across all remaining views and subsequently reconstructed. %
While sparse NRSfM approaches treat every input point independently, 
dense approaches assume that the observed surfaces are spatially coherent. 
\extended{Sparse techniques can be applied to densely tracked  measurement matrices, though the results often show various artifacts.} 

NRSfM uses only \textit{weak} prior assumptions about the observed  motions and deformations and no 3D priors. 
\extended{Hence, the history of NRSfM is the history of different weak priors and suitable application scenarios for them.} 
Significant progress was achieved in comprehending and solving this classic ill-posed 3D computer vision problem over the last decades \cite{Brand2005, Torresani2008, Gotardo2011, Paladini2012, Dai2012, Garg2013, Kong_2016, Ansari2017, Kumar2018}. 
State-of-the-art and highly influential NRSfM methods at different times were Bregler \etal~\cite{Bregler2000} (the first NRSfM method), hierarchical approach  \cite{Torresani2008}, trajectory-space method  \cite{Akhter2008}, NRSfM with minimal prior assumptions \cite{Dai2012}, variational approach \cite{Garg2013} 
as well as multi-body NRSfM \cite{Kumar2016}. 
\extended{Many constraints initially proposed for sparse NRSfM were subsequently found to be useful in dense NRSfM. 
Thus, the trajectory-based constraints introduced by Akhter \etal~\cite{Akhter2008} were successfully used in Ansari \etal's method \cite{Ansari2017} and the neural trajectory prior approach (NTP) \cite{Wang2022CVPR}. 
}

\noindent\textbf{State-of-the-Art Methods.} 
The first dense NRSfM methods \cite{Russell2012, Garg2013}  provoked many follow-up works. 
\extended{Several recent approaches achieve state-of-the-art  results on different datasets.} 
Most state-of-the-art methods follow (at least implicitly) the matrix factorization approach of Bregler \etal~and the prior assumption that the deformable shapes span low-rank  subspaces~\cite{Bregler2000}.

\begin{figure} 
    \centering 
    \includegraphics[width=\linewidth]{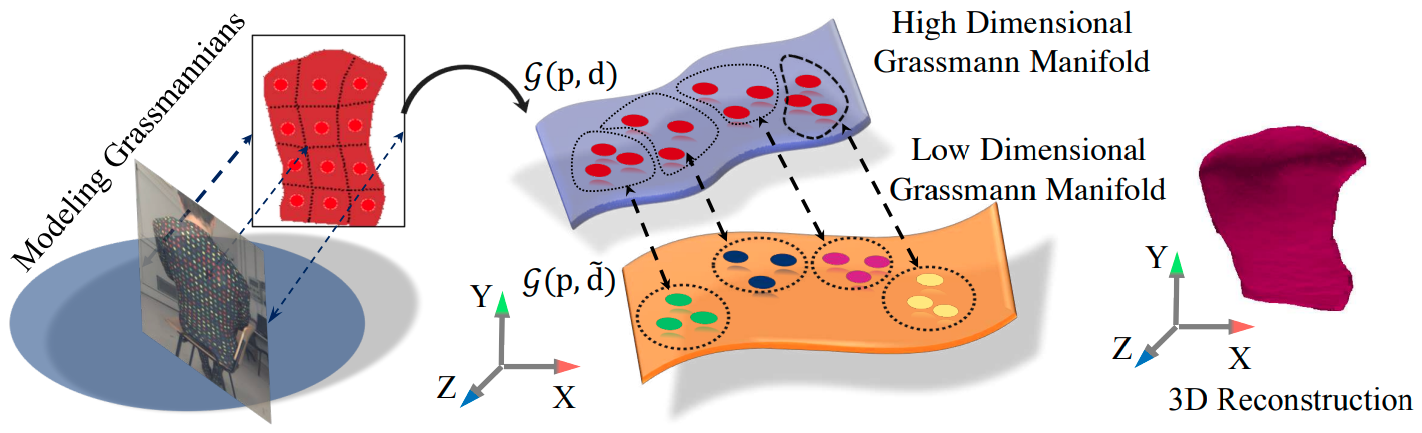}     \caption{Overview of Jumping Manifolds \cite{Kumar2019}.
    Image adapted from \cite{Kumar2019}.
    } 
    \label{fig:JM} 
\end{figure}

Several works \cite{Zhu2014, Ansari2017, Kumar2018, Kumar2019, Kumar2020} were inspired by the Block Matrix Method (BMM) of Dai \etal~\cite{Dai2012}. 
BMM is convex and only assumes the low-rank shape constraint; it showed that NRSfM could be solved unambiguously w.r.t.~the basis unknown during optimization. 
The SMSR method of Ansari \etal~\cite{Ansari2017} updates the input measurement matrix by applying smooth trajectory constraints. 
Differently from Dai \etal, they use the alternating direction method of multipliers (ADMM) to optimize the objective function. 
Moreover, SMSR converges fast and scales well across datasets of different point sizes. 
The jumping manifolds (JM) approach \cite{Kumar2019} is an  extension of Grassmannian NRSfM (GM) \cite{Kumar2018}. 
Both methods follow the ideas of point clustering and unions of linear subspaces \cite{Zhu2014}. 
JM takes into account that local surface deformations depend on point neighborhoods. 
It combines high and low-dimensional Grassmann manifolds for 3D reconstruction and clustering; see Fig.~\ref{fig:JM} for an overview of the method. 
JM currently achieves one of the lowest 3D reconstruction errors on one of the synthetic faces \cite{Vlasic2005, Garg2013}. 
The weaknesses of GM and JM is an excessive number of parameters that need to be set 
\extended{(and that cannot be known for new sequences in advance)} 
compared to many other techniques requiring much fewer of them \cite{Parashar_2020_CVPR, Sidhu2020, Grasshof2022, Wang2022CVPR}.

\begin{figure}
    \centering
    \includegraphics[width=\linewidth]{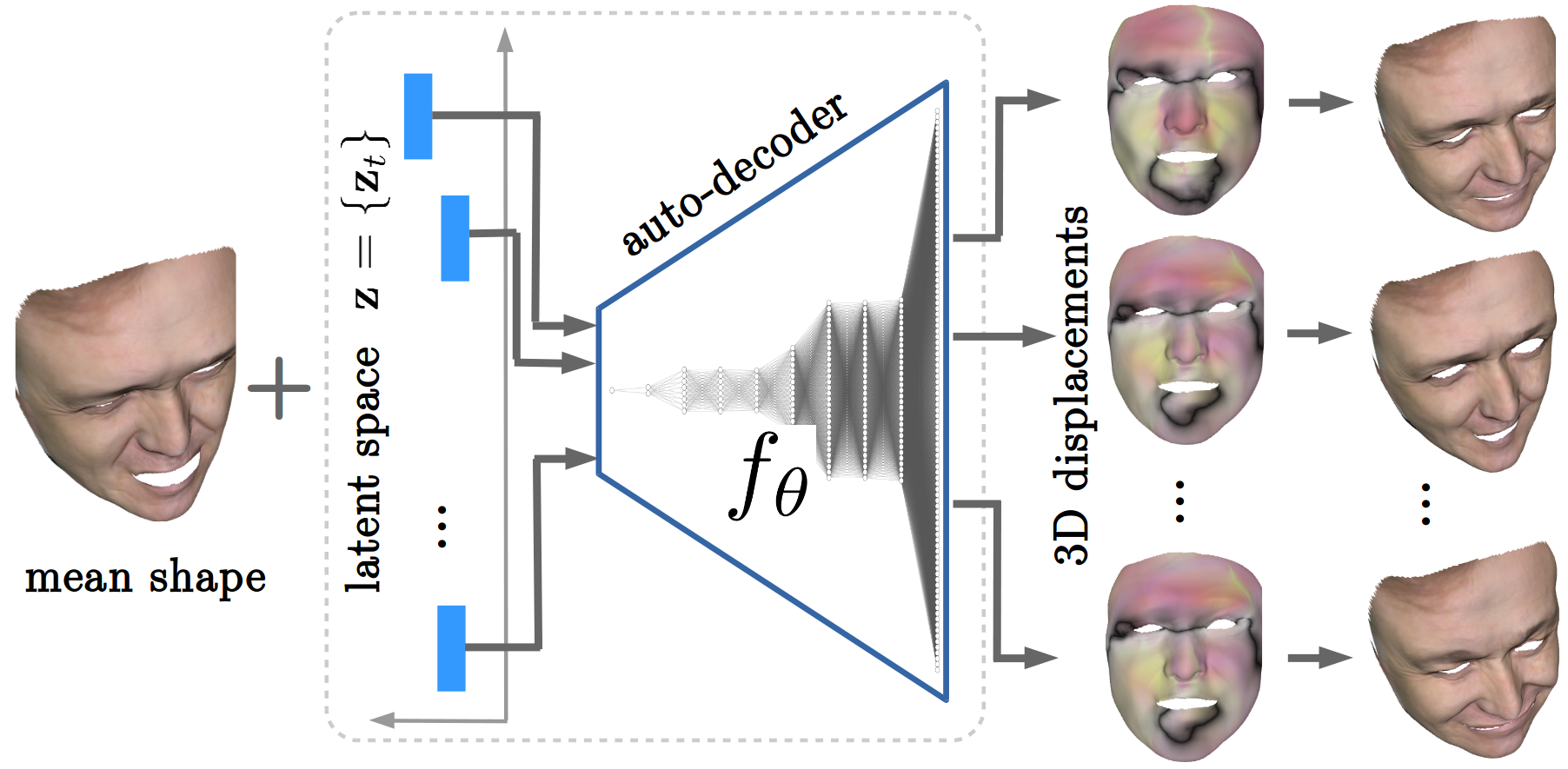}
    \caption{Deformation model of N-NRSfM \cite{Sidhu2020}, the first NRSfM approach with neural deformation model.
    Image adapted from \cite{Sidhu2020}.
    } 
    \label{fig:NNRSfM}
\end{figure}

Sidhu \etal~\cite{Sidhu2020} introduced N-NRSfM, \ie~the first \textit{neural} dense NRSfM approach with a deformation model represented by a neural network. 
They follow the auto-decoder paradigm and assign a latent space variable to each 3D state; see Fig.~\ref{fig:NNRSfM}. 
The deformation model of N-NRSfM provides sufficient  expressiveness
due to non-linearities of the MLP, and the 
latent space function  
(\ie the set of per-shape latent variables) compresses the reconstructions into a lower-dimensional space. 
A new loss Sidhu \etal~impose is the latent space constraint in the Fourier space that forces similar 3D shapes---observed in arbitrary frames---to have similar latent variables. 
It also allows to reveal periods of the input sequences.

Wang \etal~\cite{Wang2022CVPR} proposed a neural trajectory prior (NTP) for motion regularization in different 3D computer vision tasks, including scene flow integration and dense NRSfM. 
NTP relies on the smoothness bias of MLPs and imposes temporal  smoothness and spatial  similarity on continuous point trajectories. 
Similarly to N-NRSfM \cite{Sidhu2020} and PAUL \cite{Wang2021CVPR}, they use a  bottleneck layer in their model, which forces the resulting trajectories to be  compressible (\ie to lie in a low-dimensional space). 

Gra{\ss}hof and Brandt combine tensor-based modeling and rank-one 3D shape basis formulation for NRSfM \cite{Grasshof2022}. 
They recover 3D shapes up to an affine 3D transformation and perform a metric update if camera calibration is known. 
They primarily target 3D reconstruction of faces and achieve accurate results on the BU3DFE dataset \cite{Yin2006} compared to several previous methods.

\begin{figure} 
    \centering 
    \includegraphics[width=\linewidth]{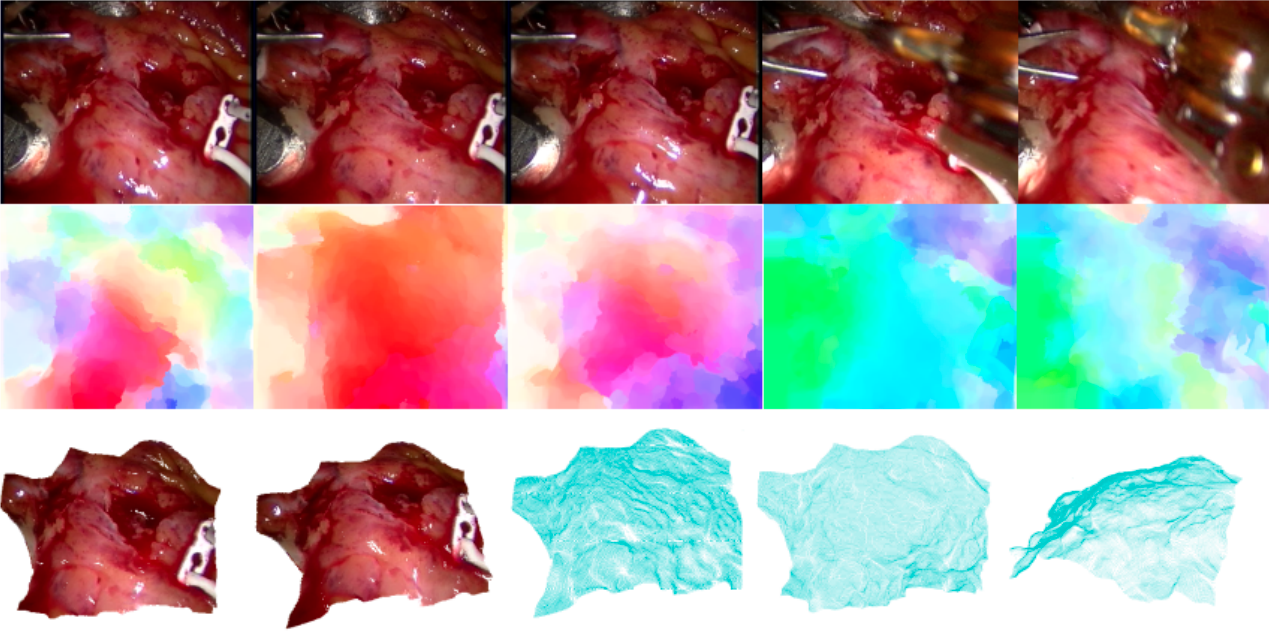} 
    \caption{DSPR's \cite{Golyanik2020DSPR} results (bottom) on the 2D input correspondences (middle) of the bypass surgery sequence \cite{Stoyanov2012} (top). %
    DSPR detects and handles the robotic arm %
    as an occluder. %
    Image adapted from \cite{Golyanik2020DSPR}.
    } 
    \label{fig:DSPR} 
\end{figure}

A different approach for smooth surfaces 
is pursued in Diff-NRSfM 
\cite{Parashar_2020_CVPR}. 
This method assumes local surface diffeomorphism 
associated with specific differential properties of the 3D  points. 
Diff-NRSfM is among the fastest methods achieving competitive  performance in dense scenarios. 
Few works target occlusion handling or restricted  
camera paths in dense NRSfM \cite{Golyanik2017SPVA,  Golyanik2020DSPR, sengupta2021colonoscopic}. 
They are motivated by medical applications (\eg endoscopy), in which relying on 2D matches only is not sufficient. 
SPVA \cite{Golyanik2017SPVA} combines NRSfM with SfT for increased 3D reconstruction stability while handling inaccurate and partially corrupted dense correspondences (\eg due to large external occlusions). 
DSPR \cite{Golyanik2020DSPR} extends this idea to a dynamic shape prior with multiple 3D states obtained on non-occluded parts of the input sequence. 
The dynamic shape prior is then used to stabilize the occluded shape parts (\eg by a robotic arm), 
while the non-occluded regions select the most suitable 3D surface (that was previously observed) for 3D shape inpainting. 
Fig.~\ref{fig:DSPR} shows DSPR's exemplary 3D reconstructions of the heart bypass sequence 
\cite{Stoyanov2012}. 
A proof-of-concept approach with topological shape prior \cite{sengupta2021colonoscopic} assumes that the reconstructed shape is tube-shaped, as expected in colonoscopy. 
It alternates between unconstrained 3D reconstruction assuming isometry and tubular parametrization upgrading the initial point clouds to tubular-shaped smooth  surfaces.

\noindent\textbf{Datasets.} 
NRSfM is a severely ill-posed problem, and no single NRSfM method was shown to reconstruct sequences observing different motion and deformation types with steadily high accuracy. 
Dense NRSfM approaches were tested on different sequences and types of non-rigid objects over the last ten years. 
We summarise most of them in the following by starting with the ones providing 3D ground truth: 
\textit{Synthetic flag} \cite{Garg2013,Garg2013V}, 
\textit{synthetic flag with occlusions} \cite{Golyanik2017SPVA}, 
\textit{synthetic faces} \cite{Vlasic2005,Garg2013}, 
actor \cite{Beeler2011, Ansari2017}, 
\textit{toss} and \textit{pants} \cite{White2007,Agudo_etal_cviu2018}, 
\textit{actor mocap}  \cite{Valgaerts2012,Golyanik_2019,Golyanik2020DSPR}; 
the sequences \textit{t-shirt} and \textit{paper} \cite{Varol2012} coming with reference depth data recorded by a Kinect sensor. 
Widely-used sequences without ground truth are: \textit{Face} (``Nico'') \cite{Garg2013}, \textit{back} \cite{Russell2011}, \textit{heart bypass surgery} (two sequences) \cite{Stoyanov2005, Stoyanov2012}, 
\textit{rabbit laparoscopy}  \cite{agudo2016sequential,Golyanik2017_CDF_INTRO}, 
\textit{liver} \cite{Mountney2010,Golyanik2020DSPR} and 
\textit{barn owl} \cite{NRSFMFLOW2016}. 
Note that a few NRSfM methods for dense reconstruction  \cite{Ansari2017, Golyanik_2019, Parashar_2020_CVPR} were  also tested on the sparse (semi-dense) NRSfM Challenge 2017 dataset of Jensen and colleagues \cite{Jensen2021}.

\noindent\textbf{Field Specifics and Open Challenges.} 
Despite all the progress, 
there remains a significant gap between the theory and practical applications of NRSfM, and several reasons for that can be named. 

\textit{First}, the input 
2D point tracks are usually extracted from the input views 
by dense optical flow techniques \cite{Garg2013V,  Taetz2016}. 
Unfortunately, most NRSfM papers 
ignore the recent progress in optical flow estimation, 
even though 1) modern deep-learning-based methods \cite{Teed2020} can be applied to deformable objects, and 
while 2) it is well known that the accuracy of NRSfM depends on the accuracy of point tracks. 
Many NRSfM datasets, however, provide ground-truth 2D correspondences obtained by re-projecting ground-truth 3D shapes to an image plane by a virtual camera. 
This allows to focus on the 3D reconstruction while  delegating dense point tracking. 
At the same time---even if a method can accurately  reconstruct a scene from accurate point tracks---it is often not known how the same approach performs on  real and deteriorated 2D tracks \cite{Garg2013, Ansari2017, Grasshof2022, Wang2022CVPR}. 
(Only several works evaluate the proposed methods on  noise-contaminated ground-truth measurements \cite{Kumar2018, Kumar2019, Parashar_2020_CVPR, Golyanik2020DSPR}.) 
All that suggests that the reported metrics in most NRSfM papers reflect an upper-bound accuracy that cannot be reached in practice.

\textit{Second}, NRSfM assumptions are often not fulfilled in practice, which results in corrupted shapes even on accurate point tracks. 
Moreover, most (if not all) sequences demonstrated in papers on dense NRSfM can be accurately initialised under the rigidity assumption \cite{TomasiKanade1992}; otherwise, dense NRSfM would not perform well on them. 
Moreover, due to the severe ill-posedness of NRSfM, there is often no unique set of parameters (of the energy terms)
working equally well across multiple datasets. 
Consequently, some recent research addresses scalability \cite{Ansari2017, Kumar2018}.

Noticeable is also the saturation of the field of dense NRSfM\extended{~(in the sense that more works are published for the sparse case)}.
One of the main reasons is that the numbers are improving marginally on the existing datasets, let alone that such improvements can barely be noticed  qualitatively. 
Most datasets contain small motions and are widely considered not challenging enough to boost the progress in dense NRSfM. 
Next, the notion \textit{NRSfM} is being used in other contexts than originally meant. 
Consider so-called ``deep NRSfM'' methods for sparse 3D reconstruction from single images 
\cite{novotny2019c3dpo, Park2020, Wang2021CVPR,  Zeng2021ICCV, Song2022}. 
The underlying neural networks 
are trained on large image collections without 3D supervision 
and do not always use observed object motions as one of the 3D  reconstruction cues. 
Moreover, these 2D-to-3D lifting methods often require different datasets for each object class.

One unsolved problem in the field remains dense NRSfM with shape completion. 
Since a single keyframe is selected for point tracking, only the points visible in it are subsequently reconstructed; the points that become visible in other frames are discarded. 
A na\"{i}ve approach with shape completion would require multiple keyframes and a subsequent 3D surface fusion; no such technique has been demonstrated in the literature yet. 
Only recently first solutions to non-rigid shape estimation and completion from monocular videos were shown in the context of non-rigid neural radiance fields  \cite{tretschk2021non,li2021neural}. 
Thus, NR-NeRF \cite{tretschk2021non} can simultaneously reconstruct a volumetric scene representation of a deformable scene from monocular videos (no 2D point tracks are required) so that all input views  contribute to the canonical volume and complement the already  available 3D densities. 
Ub4D \cite{johnson2022ub4d} specifically targets explicit surface extraction and comes in the setting even closer to dense NRSfM. 
We next look at volumetric rendering methods that reconstruct  non-rigidly deforming scenes using volumetric 3D representations.

\subsubsection{Neural Rendering Methods}\label{sec:general-nerf}

Neural radiance fields have introduced a new area of general dynamic reconstruction methods from a video that do not neatly fall into SfT or NRSfM. 
Crucially, these methods all \emph{combine} \naive~volumetric rendering and a neural scene parametrization, but differ widely in the specific kinds of input annotation used. 
Neither a template nor long-term correspondences are in principle needed as input and, unlike most prior work, they include the static background in the reconstruction, making these methods much more flexible and easier to apply in real-world settings. 
In contrast to most prior work, NeRF-based approaches tend to use density functions for geometry and not hard surfaces, allowing for some slack during optimization. 
While this slack enables almost photo-realistic novel-view synthesis, the underlying geometry is seldom evaluated as it exhibits, in most cases, rather low-quality mid-level and fine details. 
In addition, foggy artifacts may arise. 
Improving the quality of the geometry is thus central to move towards better reconstructions. 
For a detailed discussion, we refer to Tewari \etal~\cite{tewari2021advances}.

\noindent\textbf{State-of-the-Art Methods.} 
Six concurrent works were the first to extend NeRF to the dynamic setting, covering a wide design space by choosing different trade-offs. %
They fall into two broad categories: time conditioning~\cite{li2021neural,xian2021space,du2021neural} and ray bending~\cite{park2021nerfies,tretschk2021non,pumarola2021d}, which correspond roughly to the coordinate-system focus of Eulerian motion formulations and the particle focus of Lagrangian motion formulations in physics, respectively. 
The first category conditions the radiance field (a coordinate-based MLP parametrizing geometry and appearance) on a temporal input, \eg time $t$, and thus loses long-term correspondences, which gives it the freedom to reconstruct large motions and topology changes. 
Consistency across time (via jointly optimized scene flow) is encouraged by warping losses, optical-flow losses, or keypoint losses. 
Thus, information is not propagated well in the long-term, restricting novel-view synthesis to nearby views at any time~$t$. 
The second category disentangles the deformations into a separate, time-conditioned deformation field that acts on top of a static canonical radiance field, effectively bending rays to model deformations via space warping. 
Since this enforces hard correspondences across time (via the canonical model) and hence geometry and appearance information is shared across all time steps by design, it is empirically limited to a much smaller range of motion and does not cope well with topology changes but enables more challenging novel-view synthesis. 
Results from both categories exhibit close to photo-realistic appearance, although artifacts are noticeably more common than in the static setting, due to the more challenging nature of the problem. %

\begin{figure} 
    \centering 
    \includegraphics[width=0.49\linewidth]{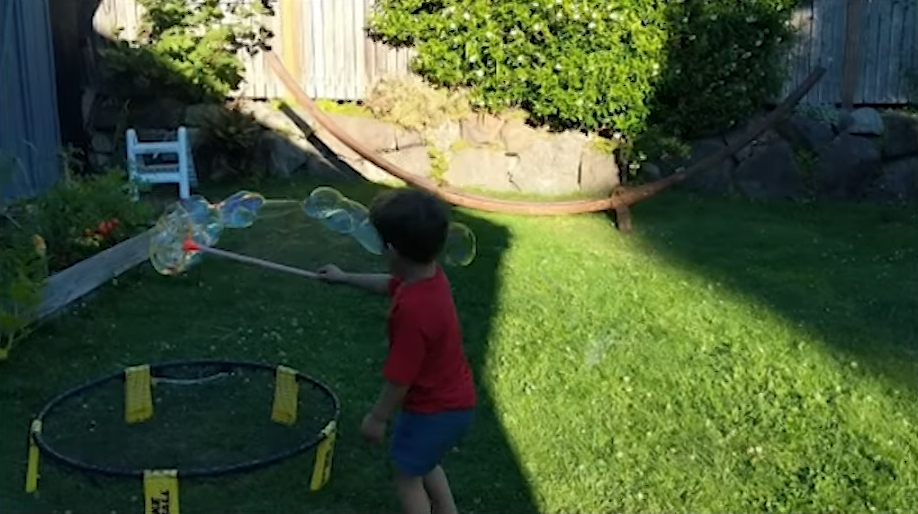} 
    \includegraphics[width=0.49\linewidth]{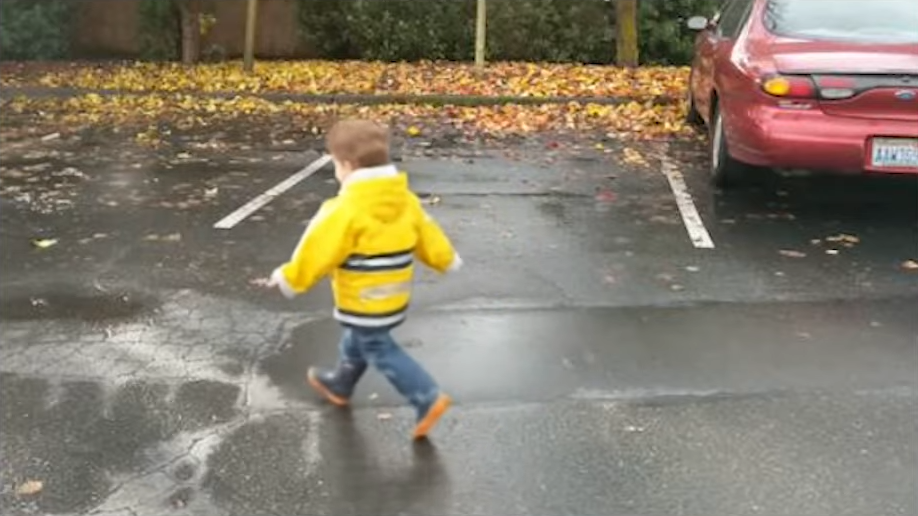} 
    \caption{Neural Scene Flow Fields~\cite{li2021neural} can model topology changes (bubbles) and view dependence (ground reflection). 
    Image adapted from \cite{li2021neural}.
    } 
    \label{fig:nsff} 
\end{figure} 
Neural Scene Flow Fields (NSFF)~\cite{li2021neural} show that complicated real-world lighting effects like shadows and reflections in dynamic scenes can be modeled well by NeRF-like approaches, see Fig.~\ref{fig:nsff}. 
Nerfies~\cite{park2021nerfies} introduce an \emph{SE(3)} deformation parametrization that is well-suited for deformations that are mostly articulated. 
Non-Rigid NeRF (NR-NeRF)~\cite{tretschk2021non} shows that a video captured by a moving camera with associated time stamps and camera parameters (and no other annotations) is sufficient to reconstruct scenes with small deformations. 
Xian \etal \cite{xian2021space} show that recent depth-estimation methods~\cite{LuoVideoDepth2020} offer helpful guidance for reconstruction. 
NeRFlow~\cite{du2021neural} uses a Neural-ODE-based~\cite{chen2018neural} deformation model, which is slow but invertible by construction and avoids self-intersections by design\extended{, unlike other approaches}. 

After this initial wave of works, progress has slowed recently, as noticeable improvements in this challenging setting, beyond mere shifting of trade-offs, have been hard to come by. 
HyperNeRF~\cite{park2021hypernerf} is a follow-up to Nerfies \cite{park2021nerfies} %
with a sophisticated conditioning of the canonical model, 
which is not only temporally but also spatially varying. 
This enables the reconstruction of topology changes \extended{(by ``slicing'' the higher-dimensional canonical model)} and larger deformations than Nerfies but comes at the cost of losing correspondences. 
It is a hybrid of both categories. 
Gao \etal~\cite{gao2021dynamic} introduce a new time-conditioned method that exploits single-view depth in a scale-invariant depth-order loss. 
Unbiased4D~\cite{johnson2022ub4d} steers NR-NeRF towards surface estimation by replacing the commonly used density function for geometry by an SDF, following NeuS~\cite{wang2021neus}. 
Marching Cubes then allows to easily extract high-quality meshes from the reconstruction, although temporal correspondences are lost in that process. 
Fang \etal~\cite{fang2022fast} apply a fast, explicit, voxel-based data structure to reduce training time from many hours to a few minutes. 
\extended{
They use an MLP to map into canonical space, which consists of hierarchical voxel grids. 
The extracted features and a time conditioning are then fed into a small MLP to obtain density and color. }
Guo \etal~\cite{Guo_2022_NDVG_ACCV} speed up training similarly. 
They explicitly handle occlusions. 
Qiao \etal~\cite{qiao2022neuphysics} use differentiable mesh-based physics simulation as a soft constraint on the deformation field. 
Subsequently, they can edit the reconstruction in a physical manner.

\noindent\textbf{Datasets.} 
So far, no standard datasets or benchmarks are established and all works evaluate predominantly on self-recorded scenes or, in some cases, the dataset from Yoon \etal~\cite{yoon2020dynamic}. 
The most recent work, Fang \etal~\cite{fang2022fast}, evaluates on synthetic scenes from D-NeRF and real scenes from HyperNeRF. 
Gao \etal~\cite{gao2022dynamic} thoroughly analyze the limitations of currently used datasets. 

\subsubsection{Other Methods for Few-Scenes Reconstruction}\label{sec:fewscene}

There are a number of other reconstruction works that focus on a single or a few scenes but do not fall into any of the previously discussed categories. 
We group them together here since a per-scene parametrization (\ie auto-decoding~\cite{Park_2019_CVPR}) is still feasible in this problem setting.
All in all, this is a nascent niche with a lot of unexplored potential. 
However, it can merge with the NeRF-style works (Sec.~\ref{sec:general-nerf}) for the foreseeable future, as BANMo~\cite{yang2022banmo} indicates, and ignore non-neural alternatives that could, for example, build on differentiable mesh rendering. 

In their pioneering work~\cite{yoon2020dynamic}, Yoon \etal~primarily work with estimated depth maps to 3D-reconstruct a temporal sequence. %
\extended{
The static background can be reconstructed well through multi-view cues from the moving camera. 
(We note that no other method in this section reconstructs the background.) 
The dynamic foreground, however, only yields per-image depth maps that are scale-inconsistent with each other. 
}
A neural network fuses these depth maps consistently and with small scene flow  %
into a novel view, filling in holes. 
Subsequent image warping of the RGB input images followed by a neural blending network enables novel view synthesis, which, however, does not give correspondences across time. 
\extended{
Their method needs estimated optical flow and manual foreground masks. 
}
The individual networks of the method need to be pretrained on synthetic or larger datasets. %
A few works in Sec.~\ref{sec:general-nerf} evaluate on Yoon \etal's dataset. 

\begin{figure} 
    \centering 
    \includegraphics[width=0.32\linewidth]{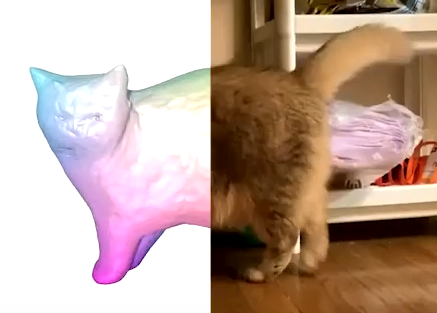} 
    \includegraphics[width=0.32\linewidth]{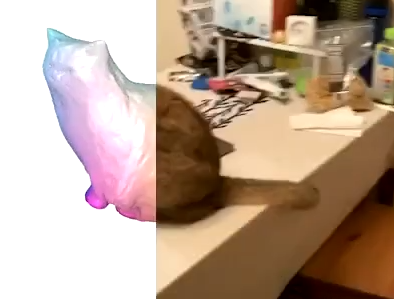} 
    \includegraphics[width=0.32\linewidth]{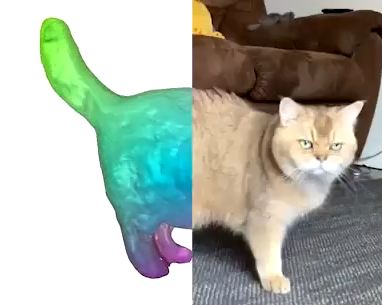} 
    \caption{BANMo~\cite{yang2022banmo} uses a NeRF-style object representation to reconstruct an object from a few monocular videos. 
    Image adapted from \cite{yang2022banmo}.
    } 
    \label{fig:banmo} 
\end{figure} 

Yang \etal's LASR~\cite{yang2021lasr} uses foreground masks and optical flow to reconstruct a general dynamic foreground object as a deforming mesh (initialized to a sphere). 
They are the first to exploit differentiable mesh rendering in the general dynamic per-scene reconstruction setting. 
\extended{
They use it to compare their skinned mesh to the input data, thereby optimizing for all parameters, including camera parameters. 
}
The follow-up ViSER~\cite{yang2021viser} additionally attaches features to the geometry and matches them to image features (the features are learned from scratch), which provides more robustness than LASR against appearance changes. 
Unlike LASR, ViSER can also reconstruct multiple videos of the same object at once. 
In a further follow-up, BANMo~\cite{yang2022banmo}, they merge this line of work with NeRF-style volumetric rendering and neural parametrization. 
It heavily relies on matching pretrained image features to establish correspondences. 
Its results on a wide variety of general objects show decent mid-level details, with slight temporal jitter and unnaturally smooth deformations, see Fig.~\ref{fig:banmo}.

LASSIE~\cite{yao2022lassie} reconstructs animals from a small-scale image collection (${\sim}30$ images of different individuals of the same species) and hence does not exploit temporal information. 
It goes even further than BANMo and completely forgoes any appearance loss, relying entirely on pretrained features. 
\extended{
It uses a fairly generic quadruped skeleton, with one 3D part attached to each bone, leading to an articulated model. 
LASSIE strongly regularizes the reconstruction: for example, the shape of each part is shared across all individuals of that species, with only a similarity transform applied per individual. 
}
Although the results are far from photo-realistic, they are promising given the very challenging setting that does not need input annotations of any kind.

\subsubsection{Methods Using Data-Driven Priors}\label{sec:manyscene}

When working with large-scale image collections instead of a few scenes at most, per-scene parametrization ceases to be practicable. 
Methods in this category thus need to rely on a data-driven prior, see Sec.~\ref{sec:fundamentals-data-driven}. 
While this setting has long been common for category-specific methods, it only recently turned out to be a viable path for general methods as well. 
The main trend in this area is a preference towards reducing the need for involved annotations and exploring alternative annotation settings like video rather than improving the quality noticeably beyond the quality of the initial work (CMR~\cite{cmrKanazawa18}). 
\extended{
In practical image-collection settings with rather loose restrictions on an object's pose and deformations, designing the right priors and input data to disambiguate geometric details from appearance remains very challenging. }
Another testament to the difficulty of the problem setting is that most methods stick with the CUB dataset~\cite{WahCUB_200_2011} of different species of birds, which only have challenging deformations in their wings, which are barely reconstructed by any existing method. 
For an excellent table summarizing methods in this section, we refer to Table~1 in a recent survey~\cite{MarvastiZadeh2022} and to Table~1 in DOVE~\cite{wu2021dove}. 

\begin{figure} 
    \centering 
    \includegraphics[width=\linewidth]{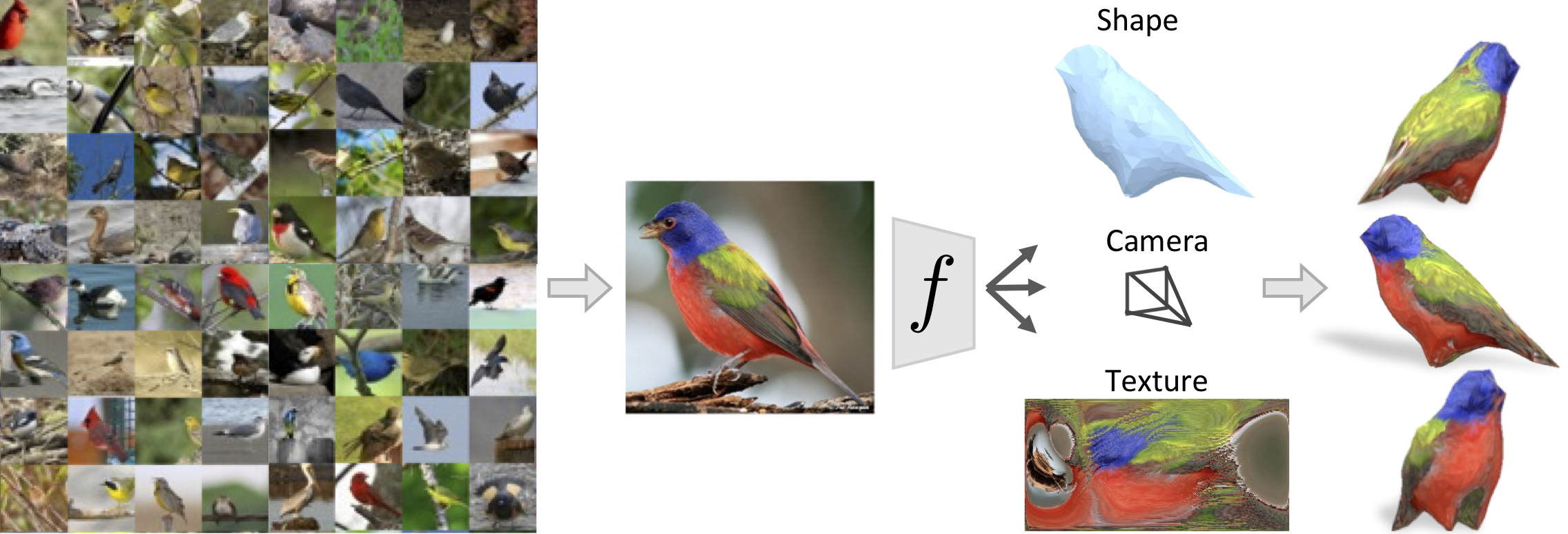} 
    \caption{CMR~\cite{cmrKanazawa18} learns a data-driven prior from a collection of images to reconstruct an object from a single image. %
    Image adapted from \cite{cmrKanazawa18}.
    } 
    \label{fig:cmr} 
\end{figure}

\noindent\textbf{State-of-the-Art Methods.} 
After the earlier work by Tulsiani \etal~\cite{tulsiani2016learning} reconstructing rigid categories by deforming a template, interest in general image-collection approaches has started growing with Kanazawa \etal's~\cite{cmrKanazawa18} CMR method, which mostly shows results on CUB, see Fig.~\ref{fig:cmr}. 
CMR uses foreground masks and manually labeled semantic keypoints as annotations, and they regress camera pose, per-vertex offsets of a mean shape, and appearance. 
Their analysis-by-synthesis method uses differentiable rendering of the mesh. 
To handle occlusions, they exploit the left-right symmetry of birds and only predict one side. 
For appearance, they regress, for every pixel of a UV map, where that pixel should sample the input image to copy its RGB color from, the so-called \emph{texture flow}, a technique that remains in wide-spread use in this line of work. 
This leads to a good appearance quality, while the predicted geometries are of rather coarse quality. 
Fine structures like legs or large deviations from the mean shape like open wings hardly exist\extended{~due to the difficult problem setting requiring strong deformation regularization, namely Laplacian smoothness and small offsets, respectively}. 
In a follow-up, Goel \etal~\cite{ucmrGoel20} (U-CMR) get rid off the need for keypoints. 
Since CMR uses SfM on the keypoints to obtain camera poses, U-CMR no longer has easy access to rough poses and they instead optimize for a per-image set of potential cameras, in auto-decoder fashion. 
The cameras thus estimated are of slightly lower quality than CMR's and, accordingly, the result quality remains, at best, comparable to CMR. 

Building on an idea originally introduced by DensePose~\cite{Guler2018DensePose}, a couple of works start from predictions of (visible) object coordinates in image space, a dense analogue to sparse semantic 2D keypoints: 
In a follow-up to their work on rigid objects, Canonical Surface Mappings (CSM)~\cite{kulkarni2019csm}, Kulkarni \etal~\cite{kulkarni2020acsm} (A-CSM) fit an articulated 3D geometry to object coordinates predicted in image space, with the articulations and coordinate predictions jointly learned without direct supervision on either. 
They require a template shape per category\extended{, for which they learn a skinning to a skeleton of fixed topology,} and can thus handle a wider variety of datasets than just CUB. 
While this makes their geometry inherently detailed, its deformations to fit to the input are rather coarse, often ignoring even legs in the input. 
Similarly, DensePose3D~\cite{shapovalov21densepose3d} exploits a pretrained DensePose model for humans and pretrained Continuous Surface Embeddings (CSE)~\cite{neverova2020cse} for animals to fit a skinned template to 2D object coordinates in image space.
\extended{~Both DensePose and CSE require manually labeled annotations for training.} 
A canonicalization loss handles missing camera poses\extended{, and they regularize the geometry with an ARAP loss}. %
Their result quality is similar to A-CSM, with only coarse deformations being somewhat accurate. 

Tulsiani \etal's IMR~\cite{tulsiani2020imr} applies CSM to CMR's setting. 
Although still unpublished, it is widely considered as a proper member of this line of work. 
They allow for instance-specific offsets \extended{(parametrized by an MLP, not individual vectors anymore)} of the template before applying the skinning, which A-CSM does not. 
T\extended{hey quantitatively outperform prior work and t}heir results contain legs and coarse deformations for a wide variety of animal species, but still severely lack in detail. 
\extended{Like U-CMR and IMR, }Li \etal's UMR~\cite{umr2020} also no longer needs keypoints or camera poses, or even any kind of template\extended{~(instead starting from a sphere)}. 
They obtain the same benefits that keypoints provide by exploiting self-supervised part segmentation in image space from prior work. 
While simplifying the required annotations, this yields quality on par with CMR. 
In their follow-up VMR~\cite{vmr2020}, Li \etal~apply a standard per-image model to a video at test time. 
In addition, they no longer assume symmetry, instead \extended{applying an ARAP loss and }replacing the single template with a linear subspace model obtained from clustering CMR's reconstructions. 
They exploit appearance constancy and the consistency of semantic parts to obtain a temporally consistent reconstruction. 
\extended{
The semantic parts are manually labeled in one frame and then propagated via optical flow. 
}
While their method makes the reconstructions less noisy\extended{, especially under novel views,} and enables asymmetric deformations, the geometry and deformations remain coarse. 

\extended{Not building on the works previously discussed, }Wu \etal~\cite{wu2020unsup3d} mainly exploit the symmetry of certain object categories like faces of humans and cats to reconstruct a canonical mesh, which is then rendered into an estimated camera view. 
Thanks to symmetry, a 2.5D mesh, and restricted input view points, they avoid having to meaningfully handle occlusions, and they hence do not need any kind of input annotation or template. 
\extended{
Unlike all other methods discussed in this section, they technically do not even require segmentation masks because the only work on images cropped to the object. 
}
Their results already exhibit decent mid-level detail, although the image resolution is rather low. 
DOVE~\cite{wu2021dove}, proposed by almost the same authors as the previous work~\cite{wu2020unsup3d}, is the first to use many videos at training time. 
Their goal is a per-image predictor at test time, the opposite of VMR's setting. 
This allows them to exploit temporal information via geometry and appearance consistency, and optical flow. 
\extended{
They do not require a template, starting from a sphere instead. 
}
Since they do not assume camera poses to be given, they argue that, for symmetric shapes, a simple flipping operation akin to their prior work~\cite{wu2020unsup3d} is enough to decide between ambiguous poses instead of optimizing for a set of different cameras. 
Their results are of coarser quality than CMR's, since their input requirements are more relaxed.

Kokkinos \etal~\cite{Kokkinos_2021_CVPR} use a sophisticated deformation model of a template that is based on Laplacian deformations~\cite{sorkine2004laplacian} in an end-to-end differentiable manner. 
As they train their per-image predictor on videos, they encourage consistency with the optical flow between neighboring pairs of frames. 
Despite using keypoints, they find the camera optimization of U-CMR helpful. 
Crucially, at test time, they refine the predictions made by the neural predictor using auto-decoding-style instance optimization, similar to works in Sec.~\ref{sec:fewscene}. 
\extended{
Since they use a template, the geometry of their results is by design somewhat detailed. 
However, only small articulations are reconstructed well, deformations that differ noticeably from the template lead to failure. 
}
On CUB, they can handle open wings but otherwise only barely improve the coarse geometry beyond the quality of CMR. 
In the follow-up TTP~\cite{kokkinos2021point}, they turn around the correspondence regression of A-CSM and IMR, instead regressing the 2D UV coordinate for every vertex of the mesh\extended{, similar to texture flow estimation}. 
\extended{Unlike their previous work, which only performs instance optimization at test time, }TTP trains a shared network for the UV regression task but performs end-to-end differentiable, iterative instance optimization to determine the deformation and camera parameters \emph{at training time}\extended{~as well}. 
This noticeably improves their result quality over prior work, with the coarse geometry mostly correct and hints of mid-level details.

TARS~\cite{duggal2022tars3D} is the first work in this section to use a neural SDF parametrization rather than a mesh with fixed topology for the geometry. 
Similar to HyperNeRF~\cite{park2021hypernerf} (see Sec.~\ref{sec:general-nerf}), TARS handles topology changes by conditioning the canonical model on a latent code\extended{, thereby removing the ability to determine correspondences across instances}. 
This lack of a shared canonical model (unlike prior work) improves the geometry quality on CUB, where open wings are now possible and some mid-level details are discernible. 

\extended{
\subsubsection{Shape from Shading} %
}
\extended{
Shape from Shading (SfS) exclusively uses \textit{shading cues} for the reconstruction of \textit{texture-less} surfaces from a single image~\cite{zhang1999shape}. 
Shading information links surface geometry, surface reflectance and scene illumination to the image brightness. 
SfS is largely unsolved and often difficult to use in practice due to its ill-posed nature outside of tightly controlled lighting environments. 
This area appears to be dormant as we were unable to identify any works published in recent years.
}
\extended{
\begin{figure}
    \centering
    \includegraphics[width=\linewidth]{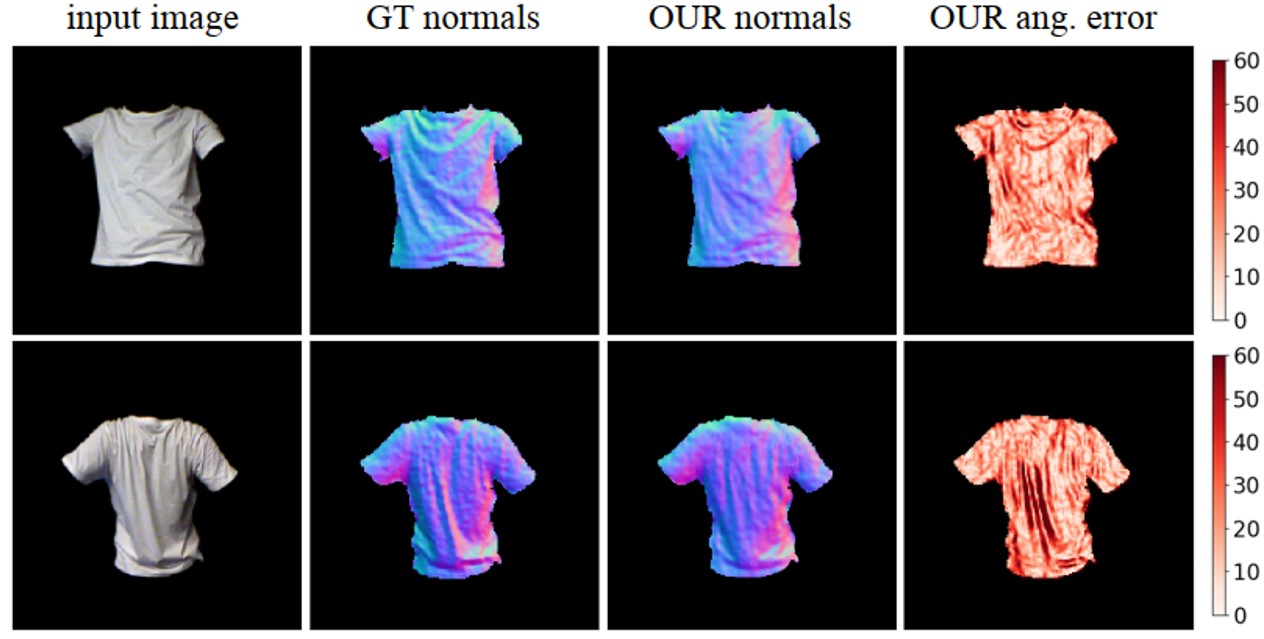}
    \caption{Reconstruction of a \textit{texture-less} deforming t-shirt by a Shape-from-Shading approach \cite{bednarik2018learning}.} %
    \label{fig:SfS}
\end{figure}
}

\noindent\extended{\textbf{State-of-the-Art Methods.} 
Yang \etal~\cite{yang2018shape} train a SfS neural network on synthetic images. %
They combine two synergistic processes: %
Guided by the network training, they evolve complex shapes from simple primitives, while the training improves by using the evolved shapes. 
In contrast to ~\cite{yang2018shape}, which does not use any external shape dataset, Bednarik \etal~\cite{bednarik2018learning} train an image-to-shape network on a real dataset. 
Similar to early classical works in SfS~\cite{horn1989shape}, they demonstrate the effectiveness of training a network for the prediction of dense maps of depth, normals, or both, as opposed to just geometry, see Fig.~\ref{fig:SfS}. 
A challenge with such global learning-based methods is that obtaining a training set consisting of all possible deformations an object may realize is impracticable. 
To address this, Tsoli \etal~\cite{tsoli2019patch} learn \textit{local} models of shape variation from image patches and then combine them into a global reconstruction of the observed object.
}

\noindent\extended{\textbf{Shading Cues in SfT and NRSfM.} 
Poorly-textured surfaces are a challenge for most SfT %
and NRSfM %
methods and this can be improved with SfS techniques.
Shading-based refinement can be used as a post-processing step
\cite{moreno2009capturing, varol2012monocular, garrido2013reconstructing}. 
Alternatively, motion and shading cues can be integrated into a single energy function and jointly optimized as in registration-based~\cite{gallardo2016using} and analysis-by-synthesis~\cite{liu2017better} SfT methods. 
Albedo is required in order to apply shading constraints, although it is usually not given as an input. %
Hence, SfT methods~\cite{gallardo2016using, liu2017better} decompose the template texture map to initialize albedo, while NRSfM methods~\cite{gallardo2017dense,gallardo2020non} estimate surface albedo together with non-rigid 3D deformations. 
All in all, using shading cue leads to more detailed reconstructions as  surface creases and wrinkles are recovered. 
}

\noindent\extended{\textbf{Open Challenges.}
Most SfS approaches struggle to handle depth discontinuities such as sharp edges, self-shadowing, inter-reflections as well as convex/concave ambiguities.
Optimization-based methods for SfS require a complete photometric calibration, \ie known or constant albedo and scene illumination. 
Learning-based methods can operate under much weaker assumptions and are practical for scenes where the lighting and material is complex, but they fail to generalize to novel scenes. 
Notably, shading cues, which were initially used as the sole cue in the SfS paradigm, are insufficient for dense monocular reconstruction. 
Thus, future methods may improve the quality and robustness of reconstructions by combining them with other inputs. %
}

\subsection{Humans} %
\label{subsec:humans}
Capturing the deforming 3D surface of humans from a single RGB camera, also called \textit{monocular (human) performance capture}, has become a very active research area over the last decade. 
It complements and refines concepts initially introduced for the general case (Sec.~\ref{sec:general-objects}). 
A key difference is that those methods introduce human-specific priors because the rough shape and topology remain the same irrespective of gender, age, and clothing type. 
\extended{These human-centered technical advances make it worth discussing them within this survey on monocular non-rigid 3D reconstruction.} 
\par
We categorize existing methods based on how strong their assumptions about the 3D geometry of the person are. 
Template-free methods do not assume prior knowledge of the specific 3D geometry (Sec.~\ref{sec:human_template_free}).
Parametric methods leverage a low-dimensional parametric model of humans obtained by a database of 3D scans of thousands of humans (Sec.~\ref{sec:human_parametric}). 
\extended{
Importantly, we do \textit{not} review works that solely regress pose and/or shape parameters of a parametric body model, \eg SMPL~\cite{SMPL:2015}, since they naturally cannot recover dense and non-rigid 3D deformations. %
}
Finally, template-based methods assume a pre-scanned 3D template of the person is given\extended{~and the deforming states that explain the video are unknown} (Sec.~\ref{sec:human_template_based}). 
Before we review all categories in more detail, we introduce the problem-specific challenges in Sec.~\ref{sec:human_challenges}.

\subsubsection{Challenges} \label{sec:human_challenges} %
On top of the general challenges (Sec.~\ref{sec:difficulties}) of this inherently ambiguous setting, there are also human-specific ones. %
Humans are composed of individual body parts, \eg arms and legs, which can move in a highly articulated and fast manner, leading to large displacements\extended{~in the observed images}. 
This makes finding correspondences between neighboring frames non-trivial, and photometric consistency between a model and the input image can be challenging due to the local nature of image gradients.
Moreover, the articulated structure can lead to severe self-occlusions, \ie one body part is occluding another, and a sudden change in visibility can occur. 
This change in visibility is not only hard to track but also non-differentiable, and occluded body parts can only be tracked using priors.
Last, there are two types of deformations for humans: 
The piece-wise rigid deformation induced by the skeletal pose and the non-rigid deformation of the surface, \eg of the clothing. 
Both of them require special care and, at the same time, can only be considered jointly. %
\subsubsection{Template-Free Methods} \label{sec:human_template_free} %
Template-free methods do not assume the availability of a known template or a parametric model. 
Initial methods like BodyNet~\cite{varol18_bodynet} and DeepHuman~\cite{Zheng2019DeepHuman} learned to reconstruct the human at the voxel level. 
However, such a representation is memory-intensive and  suffers from quantization issues. 
To mitigate the memory issue, Moulding Humans~\cite{Gabeur_2019_ICCV} reconstructs humans by estimating the front and back depth maps\extended{~(thus, the name 'moulding' humans)}.
In a similar fashion, PeeledHuman~\cite{jinka2020peeledhuman} represents a human shape as a set of depth maps at the points of intersection of the camera rays with the human surface. 
While such representations use less memory compared to voxel-based reconstruction, they cannot account for high-frequency details due to the finite resolution of depth maps. %
\par
\begin{figure}
	\includegraphics[width=\linewidth]{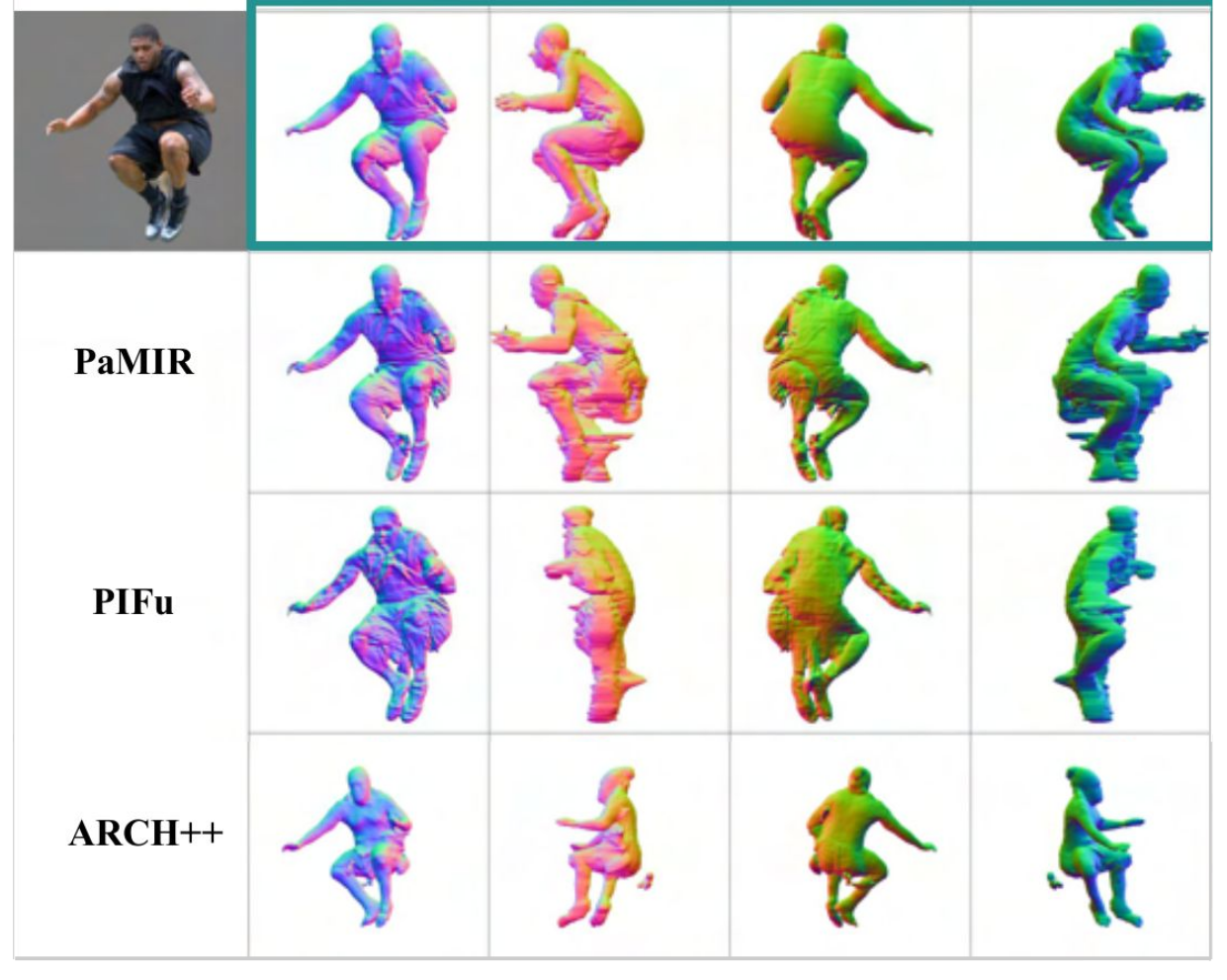}
	\caption
	{
	   Comparison of several monocular human reconstruction methods as illustrated in~\cite{xiu2022icon}. The top row corresponds to ICON~\cite{xiu2022icon}. Template-free methods like PIFu tend to struggle when tested on challenging poses.
	   Image adapted from \cite{xiu2022icon}.
	   }
	\label{fig:humans_comparison}
\end{figure}

These issues motivated work towards learning implicit models to represent a human.
Thus, PIFu~\cite{saito2019pifu} and PIFuHD~\cite{saito2020pifuhd} learn a zero-level set of the surface that can represent high-frequency details, as the model learns a continuous representation. 
This allows for improved handling of hair and clothing deformations (see Fig.~\ref{fig:teaser}, bottom right). 
However, generalization to arbitrary poses  (Fig.~\ref{fig:humans_comparison}) 
remains challenging 
because the only global context provided to the network comes from the image features at the query point. %
To mitigate this, Geo-PIFU~\cite{he2020geopifu} 
adds a 3D U-Net branch that provides geometric features for a query point. 
\par
Several NeRF-based methods were recently proposed to reconstruct humans in 4D from a monocular video.
Human-NeRF \cite{weng_humannerf_2022_cvpr} learns appearance as a continuous field in a canonical space and learns a mapping from the motion field to canonical space using two modules. 
The first module learns the skeleton-level deformations and the second one accounts for non-rigid deformations by learning corrective offsets on top of those deformations. %
Also related is PHORHUM~\cite{phorhum} that learns an SDF for the human body and based on pixel-aligned features, similar to PIFu. 
The method also estimates albedo with the same network and shading is estimated by a separate network using illumination features and the surface normals.
\subsubsection{Approaches Using Parametric Models}  \label{sec:human_parametric} %
Several 
methods use parametric models SMPL, SMPL-X, GHUM(L), or imGHUM~\cite{SMPL:2015, SMPL-X:2019, xu2020ghum, imghum} to estimate coarse pose-dependent geometry. 
They provide a topographically consistent canonical space and skinning weights. 
 
Methods like MonoClothCap~\cite{monoclothcap} optimize for per-vertex offsets from the SMPL template mesh.  
\extended{The deformations are estimated using photometric and geometric constraints.} 
However, the optimization-based pipeline requires up to five minutes to reconstruct one frame. 
Methods like Tex2Shape~\cite{alldieck2019tex2shape} and Alldieck \etal~\cite{alldieck19cvpr} instead learn the per-vertex deformations and normals either directly from a UV-unwarped texture map obtained from the estimated SMPL mesh or from part-wise segmentation images. 
They can thus learn geometry in an image-to-image translation fashion, significantly reducing the inference time. 
\par 
However, estimating the per-vertex deformations of a parameterized mesh inherently limits the level of high-frequency details that can be retrieved.  
This motivated several methods that learn an implicit representation based on a parametric mesh.
Such methods first learn to map a point in the observation space to the canonical space of the parameterized mesh. 
Thus, piecewise rigid deformations are modeled using the skinning weights of the parametric model, and the non-rigid deformations are typically learned using a separate network.
ARCH~\cite{Huang_2020_CVPR:ARCH} proposes to learn the surface deformations as an implicit surface based on image features and a semantic deformation field that warps a posed mesh to the canonical space.
ARCH++~\cite{He_2021_ICCV:ARCH++} improves this by sampling a point cloud from the corresponding parametric mesh in a canonical space and then extracting spatial features using a PointNet++ Encoder.
These spatially-aligned features, along with the pixel-aligned features from a UNet, are then fed to an occupancy network to learn the occupancy field.
However, noisy observations can make it difficult to estimate the warping function. %
Alternatively, one can voxelize the estimated parametric mesh and extract the 3D voxel-aligned features from a 3D network, as done by PaMIR~\cite{zheng2021pamir}. 
Similar to the geometry-aligned features of Geo-Pifu, PaMIR proposes to use these 3D features in conjunction with the image features to learn an implicit 3D surface.
ICON~\cite{xiu2022icon} learns an implicit 3D surface as a function of the front and back surface normal features and the SDF of the corresponding SMPL mesh. 
\extended{The surface normal features are extracted from two normal estimation networks that are trained to estimate the normal maps of the front and back body.}
Recently, HF-Avatar~\cite{hfavatar} proposes to produce high-fidelity by learning a reference-based neural rendering network and using it to refine the neural texture of the human in a coarse-to-fine manner.

Some methods split human performance capture into human body reconstruction and clothing reconstruction~\cite{bhatnagar2019mgn, physics_clothcap}.
MulayCap~\cite{su2022mulaycap} estimates the parameters of the garments and uses them to re-dress the naked-body SMPL mesh using a simulator. 
The textures are reconstructed by rendering using the regressed albedo maps and shading images.
There are also attempts towards learning a parameterized model of human clothing like SMPLicit~\cite{corona2021smplicit}. 
However, parametric clothed human reconstruction is often restricted to a few clothing types and, generally, cannot span the space of all clothing items.

NeuMan~\cite{neural-human-radiance-field:neuman} reconstructs the scene as well as the human by training a separate NeRF for each and jointly integrating samples from each in a ray.
For human-specific deformations, NeuMan transforms the points in the observation space to a canonical space of an SMPL mesh using pose-dependent transforms. 
There are also multi-view neural-rendering-based methods like H-Nerf~\cite{NEURIPS2021_7d62a275} and Neural Body~\cite{peng2021neural} that use the latent codes of imGHUM and SMPL, respectively, to train a NeRF (or SDF). 
However, their reconstruction quality significantly degrades when tested with a monocular input, even with relatively simple articulation\extended{, \eg from the People Snapshot dataset~\cite{alldieck2018video}}. %
\subsubsection{Template-Based Methods} \label{sec:human_template_based} %
Finally, template-based methods assume a textured 3D geometry of the subject to be given. 
In contrast to the general case (see Sec.~\ref{sec:sft}),  \emph{template} here refers to a textured 3D mesh of a clothed human.
Typically, such a template is acquired by moving around the subject standing in a static T-pose and recording a monocular RGB video~\cite{monoperfcap, livecap}. 
In an additional semi-automated step, the character mesh is then rigged and skinned to a kinematic skeleton.
Given such a template and the RGB video of the person in motion, the goal of these works is to estimate the space-time coherent, dense, and non-rigid deformation of the 3D template.
\par
\begin{figure}
	\includegraphics[width=\linewidth]{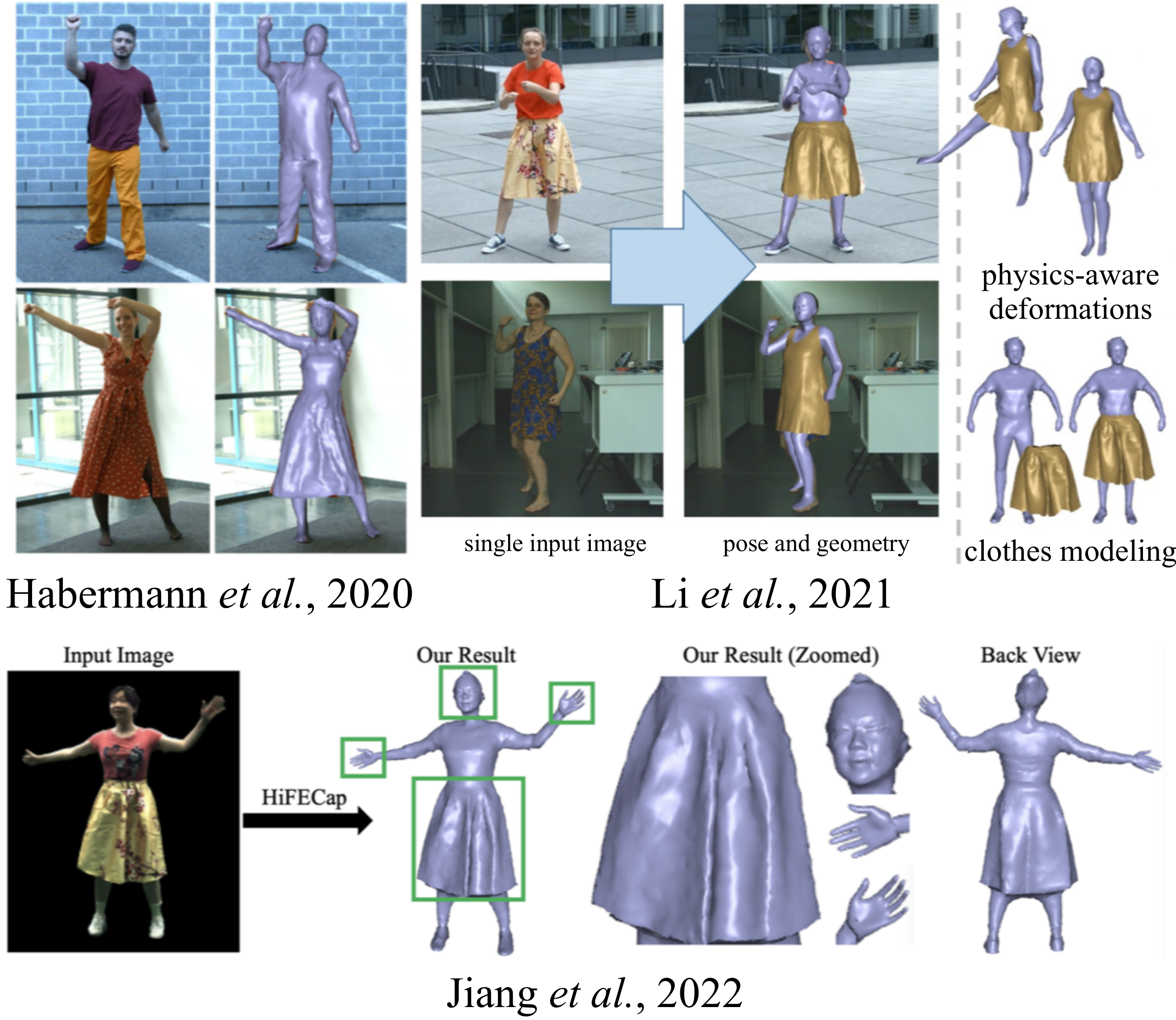} 
	\caption
	{
	    Evolution of template-based human performance capture methods. 
	    Images adapted from \cite{DeepCap,Li2021,jiang2022hifecap}.
	}
	\label{fig:humans_template_based}
\end{figure}
The pioneering work MonoPerfCap~\cite{monoperfcap} is the first method that jointly tracks the skeletal pose and the dense surface deformation of the template from a single RGB video. 
In the first stage, they estimate the skeletal pose by regressing 2D and 3D joint predictions. 
Then, they fit the skeletal motion represented by a discrete cosine transform to the predictions by optimizing a non-linear energy function. 
The obtained skeleton motion is then used to coarse-deform the template geometry and, in a second stage, further refine the surface deformations by fitting the geometry to the human silhouette. 
Although this work can achieve robust and temporally smooth results, the overall method requires more than a minute to optimize a single frame. 
LiveCap~\cite{livecap} is a more efficient approach for optimizing motion and surface deformation.
It introduces an updated energy formulation and skeleton motion representation in conjunction with dedicated GPU solvers and a multi-threaded CPU pipeline to become the first real-time method. %
\par
Nonetheless, it remains challenging to achieve high 3D accuracy due to the depth ambiguity, and occluded surface parts are mostly driven by geometric priors and not any data terms. %
To overcome this limitation, DeepCap~\cite{DeepCap} proposes to train skeletal pose and surface deformation networks, which take as input a single RGB image. 
During training, these networks are weakly supervised on multi-view images, which allows supervising surface areas that are occluded in the input view and also improves 3D accuracy.
\par 
The methods discussed so far all treat the template as a single connected surface, which does not reflect reality since clothing and the human body are separate.
Thus, shifts of clothing along the body cannot be tracked well, and the cloth deformations usually contain baked-in wrinkles from the static template and do not look physically plausible as can be seen in Fig.~\ref{fig:humans_template_based}. 
Li \etal~\cite{Li2021} propose to separate the geometry into two layers, \ie clothing and the human body.
While the pose and deformation networks are leveraged from DeepCap~\cite{DeepCap}, they introduce a physics simulation layer, which enforces more physically plausible deformations during training and prevents cloth-body surface penetrations. 
\par 
HiFECap~\cite{jiang2022hifecap} is the first method that jointly tracks the deforming clothing, the body pose, hand gestures, and facial expressions.
They introduce a hybrid neural network architecture consisting of image and graph convolutions to better recover surface details. 
They demonstrate how existing parametric hand and face models can be fit onto a 3D template, and how those can be jointly deformed with the surface deformation of the clothing.
\subsubsection{Future Directions} \label{sec:future_directions_humans}
While parametric models of clothing \textit{geometry} have been studied recently, 
creating a parametric geometry and appearance model of the whole human body remains an open challenge. 
This is due to a large amount of data necessary to sufficiently sample the model space. 
However, recent progress in dataset~\cite{zhu2020deep, humman}  acquisition may now enable the building of such a model. 
Another unsolved fundamental problem is the tracking of topological changes (\eg the person is taking off their jacket), while maintaining correspondences over time. 
Recently, implicit human models have been extensively researched, 
which can deal with topological changes due to their implicit representation. 
However, they lack space-time coherent correspondences.
Complementary, explicit mesh models have also been studied.
While maintaining correspondence naturally, they fail to faithfully track topological changes or surface details. 
Thus, in the future, a combined representation could lead to the best of both worlds.
Moreover, the joint capture of all aspects of the human is still in its infancy, \ie tracking of hands, face, body pose, clothing, hair and eye gaze. 
While the solution for individual body parts exists, it remains an open question of how they can be efficiently and effectively combined for real-time performance. %
The detailed tracking of hair is another open challenge since its thin and highly dynamic structure is not suited for surface-based methods. 
Thus, future research may involve alternative representations for hair that enable space-time coherent tracking.
Finally, the robustness and interpretability of results are still a problem for learning-based approaches. 
Here, physics could improve the performance further, 
as seminal works already show~\cite{Li2021, PhysCapTOG2020, santesteban2022snug, PIPCVPR2022}. 

\subsection{Faces}
\label{subsec:face_comp}
\begin{figure}
\includegraphics[width=\linewidth]{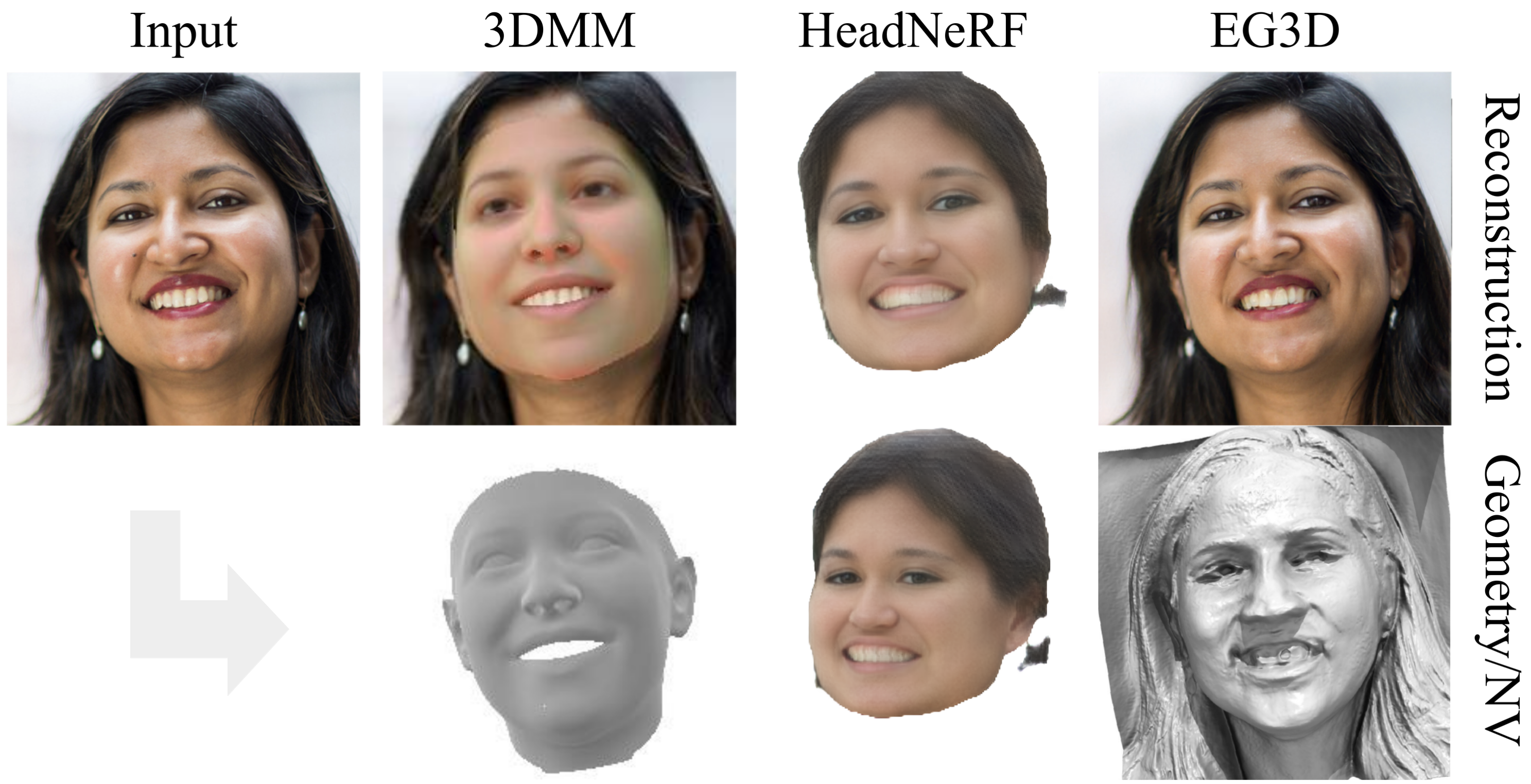}
\caption{
Results of representative methods using an explicit 3DMM~\cite{li2021fit}, trained on multi-view data (HeadNeRF~\cite{hong2021headnerf}) or in adversarial manner on large-scale monocular dataset  (EG3D~\cite{Chan2021}). 
We show (top) the reconstruction and (bottom) its geometry or novel view synthesis (for HeadNeRF). 
Image adapted from \cite{Chan2021}.
}
\label{fig:faces_reconsruction} 
\end{figure}

3D reconstruction of faces from monocular images is a heavily researched topic. 
In contrast to many other object types, faces have such advantageous properties as symmetry, small deformations and well-defined keypoints 
that can be exploited in the ill-posed 3D reconstruction setting. 
Facial shapes can be modeled in low-dimensional spaces and with linear models. 
Simple and effective models like PCA-based ones lead to reasonable results and are currently state-of-the-art in the monocular inverse rendering setting. 
Fueled by the availability of large amounts of data, this leads to faces being one of the dominant applications in the community. 

This section provides an overview of recent developments, datasets, and applications of monocular 3D face reconstruction. 
Whilst some methods focus on photorealism and learned representations, other applications  exploit a parametric representation and benefit from a classical statistical prior. 
The following parts are structured focusing on the distinction between classical explicit (Sec.~\ref{sec:faces_morphable_models}) and modern implicit models (Sec.~\ref{sec:faces_implicit_morphable_models}). 
We also cover specialized models for facial parts and the recent new dataset (Sec.~\ref{sec:faces_datasets}). 
Moreover, Tab.~\ref{table:overview_faces} categorizes the covered methods.

\subsubsection{Explicit Morphable Models}\label{sec:faces_morphable_models}
3D Morphable Models (3DMMs)~\cite{blanz1999morphable} are statistical models of face shape and appearance variation with an explicit surface representation. 
They are built from a (comparably) small set of hundreds of faces and can be used as a prior for non-rigid 3D face reconstruction. 
There are recent surveys on 3DMMs \cite{egger20203d} and monocular 3D face reconstruction and tracking \cite{zollhofer2018state}, and we here focus on the recent developments arising after these surveys. 

The core application area of explicit 3DMMs is the 3D reconstruction of faces from single 2D images through inverse rendering. 
 While this problem has been studied extensively  \cite{egger20203d}, most evaluations were performed qualitatively and in highly constrained scenarios. 
One current trend in 3D face reconstruction with 3DMMs is their application to in-the-wild images. 
 The NoW challenge \cite{RingNet:CVPR:2019} provides, for the first time, a way to quantitatively evaluate dense 3D reconstructions on in-the-wild images. 
 The NoW challenge contains 2054 2D images of 100 subjects and  ground-truth 3D scans of each person.  
 Notably, the 3D scans were captured in a studio and not at the same moment as the images. 
Most methods participating in the NoW challenge are unsupervised or weakly supervised methods that do not exploit pairs of 2D images with 3D geometry. 
Whilst such data exists for controlled lab conditions, it is not publicly available for the in-the-wild setting.

The state-of-the-art methods on the NOW challenge and in the unsupervised setting are DECA \cite{Feng:SIGGRAPH:2021} and  FOCUS \cite{li2021fit}, respectively, 
both 3DMM-based. 
Notably, no implicit modeling approach has participated in 
NOW 
so far. 
DECA exploits weak identity supervision and learns on videos and images. 
It goes beyond the simplistic linear 3DMM space and adds fine details to the 3D reconstructions through a displacement map  trained with a detail-consistency loss (to separate person-specific details from a generic learned expression model). 
FOCUS, instead, learns without identity supervision and achieves similar performance by learning a robust model estimation and being robust to occlusions. 
The recent MICA approach \cite{zielonka2022mica} is trained on paired 3D and 2D data in a fully supervised way. 
Whilst the original NoW challenge ignores the scale of the reconstructed face, MICA can reconstruct the face shape well due to the 3D supervision; it outperforms unsupervised methods by a large margin under the metrical evaluation protocol.

TRUST \cite{Feng:TRUST:ECCV2022} by Feng \etal~(see Fig.~\ref{fig:teaser}, bottom row; second on the right) is the first method explicitly aiming at correct skin tone estimation based on weak supervision through multiple faces in an image and assuming a constant illumination condition. 
Their FAIR dataset addresses biases in 3D face reconstruction  regarding skin tones and ethnicity \cite{Feng:TRUST:ECCV2022}. 
It provides ground-truth albedos of synthetic faces to evaluate albedo reconstruction, focusing on disentangling diverse skin tones and lighting conditions. 

Notably, explicit 3DMMs based on PCA \cite{blanz1999morphable} are still state-of-the-art for both the NoW and the FAIR benchmark, despite the extensive research done in the area of implicit models in recent years  (Sec.~\ref{sec:faces_implicit_morphable_models}). 
We assume this is due to the difficulty of inverting implicit models in challenging scenarios that include extreme poses,  illumination, or occlusions and their comparably expensive rendering. 
Whilst model inversion was trivial for the Eigenfaces  \cite{sirovich1987low,turk1991eigenfaces} approach, it became increasingly difficult for active appearance models 
\cite{Cootesetal98} and is still a research topic for explicit and implicit 3DMMs.

\subsubsection{Implicit Morphable Models}\label{sec:faces_implicit_morphable_models}
Recently, there has been a flurry of methods based on implicit representations. 
As compared to mesh-based representations, implicit models are not restricted to a fixed topology and, as a result, can model the entire head, including hair. 
As these methods also model the face appearance using neural networks (which generally have a much better capacity than simple linear models of 3DMMs), they can synthesize photorealistic faces. 
It is observed that most of the methods based on implicit representations target applications that require photorealistic renderings, with very few methods targeting accurate geometry estimation~\cite{ramon2021h3d, imFace2022}. 

We can broadly categorize implicit models based on the type of training data. 
Some methods use posed multi-view image sets of several identities~\cite{ramon2021h3d, hong2021headnerf, wang_morf, zhuang2022mofanerf}, the others use monocular images of several identities without paired camera poses~\cite{wu2020unsup3d, chanmonteiro2020pi-GAN, Chan2021, deng2022gram, gramHD, gu2021stylenerf, orel2022stylesdf}, and several approaches work with video data~\cite{zheng2022IMavatar, grassal2021neural, gafni2021nerface, athar2022rignerf}. 
Methods that use data of multiple identities at training time usually build a prior of face shape and appearance with implicit representation using a neural network to parameterize the face space. 
These models act as strong priors at test time to reconstruct any face, also from a single image.
Person-specific methods require video data of a single person's face and can reconstruct the entire scene through time.
Very few implicit-based methods address the correspondence problem  ~\cite{wang_morf,tewari2022d3d}; correspondences are necessary for  downstream applications such as texture transfer or registration.

\noindent\textbf{Multi-View Supervision.} 
Several approaches  
use multi-view data of multiple identities to learn face priors, 
mostly using neural networks. 
These methods draw inspiration from the auto-decoder architecture of DeepSDF~\cite{Park_2019_CVPR}. 
Instead of a separate neural network representing each sample, DeepSDF showed that it is possible to learn the entire space of an object category by conditioning the network output with a latent vector specific to each object sample. 
During training, the latent vectors are also learned along with the network parameters.
Then, at test time, an unseen sample is reconstructed by optimizing for the latent vector. 

HeadNeRF~\cite{hong2021headnerf} extends this strategy to face data and builds autodecoder models using data from multiple identities. 
As their training dataset also contains each identity in multiple expressions, they can disentangle deformations due to facial expressions from identity-specific deformations 
by having a separate latent vector for expressions. 
\extended{In order to obtain semantic correspondences with existing 3DMMs, they initialize the latent vectors to parameters obtained by fitting a 3DMM to a given sample. Since the 3DMM parameters are limited by the linear model, they let the initialized parameters deviate slightly to accommodate person-specific high-frequency details.}
MoFANeRF~\cite{zhuang2022mofanerf} also has similar design choices with one exception, \ie the training is performed in a feed-forward manner with identity and expression parameters estimated with the help of 3DMM. 
\extended{At test time, they optimize for these parameter in an auto-decoder manner.} 
Since this model is trained on data without hair and relies on estimated 3DMM parameters only, they cannot model hair.
MoRF~\cite{wang_morf} extends a similar approach to include more features: They learn to map each training identity into a canonical space, use registered meshes for guidance and model diffuse and specular components explicitly.

\begin{figure}
    \centering
    \includegraphics[width=\linewidth]{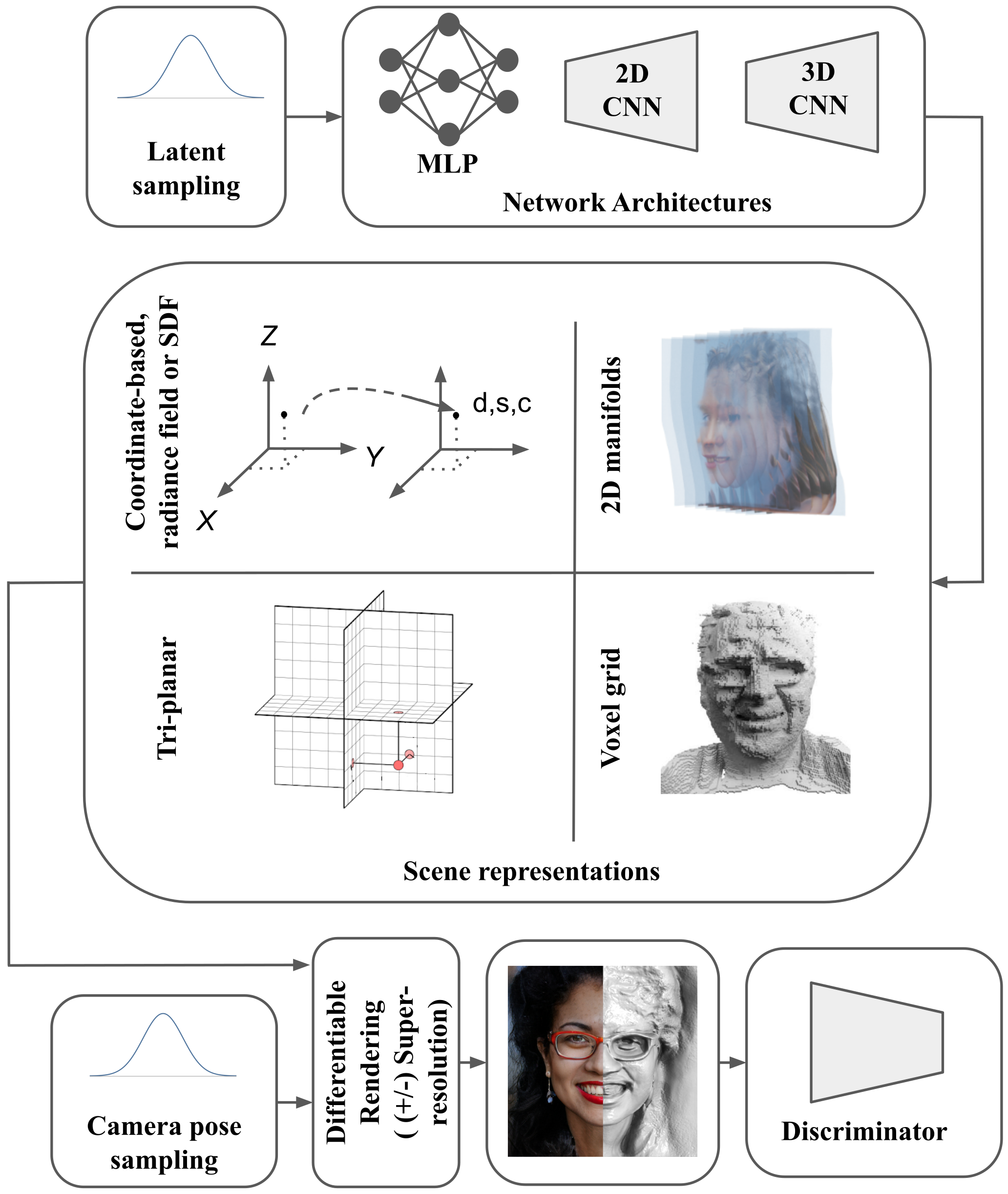}
    \caption{
    A typical pipeline for 3D-aware GANs. 
    The generator can be parameterized by an MLP~\cite{Schwarz2020NEURIPS,chanmonteiro2020pi-GAN}, a 2D CNN~\cite{Chan2021} or a 3D CNN~\cite{xu2021volumegan}, to regress a radiance field~\cite{Schwarz2020NEURIPS,chanmonteiro2020pi-GAN}, SDF~\cite{orel2022stylesdf}, 2D manifolds~\cite{gramHD}, tri-planar~\cite{Chan2021} or voxel-based scene representation~\cite{xu2021volumegan}, which is trained with an image discriminator. Images adapted from \cite{gramHD,Chan2021,schwarzvoxgraf}.
    }
    \label{fig:3d_aware_gan}
\end{figure}

\begin{table*}[!htbp]  
\small{
\begin{tabular}{m{0.22\linewidth} | m{0.19\linewidth} | m{0.18\linewidth} | m{0.17\linewidth} | m{0.11\linewidth}} 
\toprule 
\textbf{Object Type} &  \textbf{Representation} & \textbf{Prior} & \textbf{Training Data} & \textbf{Other Features} \\  \midrule 

only face: {\tiny \cite{Feng:SIGGRAPH:2021,li2021fit}}
\newline full head: {\tiny \cite{hong2021headnerf,zhuang2022mofanerf,wang_morf,ramon2021h3d,Schwarz2020NEURIPS,chanmonteiro2020pi-GAN,deng2022gram,rebain2022lolnerf,orel2022stylesdf,gu2021stylenerf,gramHD,xu2021volumegan,tewari2022d3d,pan2022gan2x,mbr_gcorf} }
\newline hair: {\tiny \cite{Chai2016AutoHairFA,2018arXiv180904765L,hu_avatar_2017,hairnet_2018,Saito:2018:3HS,yang2019dynamic} }
\newline eyes: {\tiny 
\cite{berard2016lightweight,wood16_eccv,ploumpis2020towards} }
\newline ear: {\tiny \cite{9d7119238bb34585ab26564a124d986b,8373858,ploumpis2020towards}} &

implicit: {\tiny\cite{hong2021headnerf,zhuang2022mofanerf,wang_morf,ramon2021h3d,Schwarz2020NEURIPS,chanmonteiro2020pi-GAN,deng2022gram,rebain2022lolnerf,orel2022stylesdf,gu2021stylenerf,gramHD,xu2021volumegan,tewari2022d3d,pan2022gan2x,mbr_gcorf}} 
\newline explicit: {\tiny \cite{Feng:SIGGRAPH:2021,li2021fit,Chai2016AutoHairFA,2018arXiv180904765L,hu_avatar_2017,hairnet_2018,Saito:2018:3HS,yang2019dynamic,berard2016lightweight,wood16_eccv,ploumpis2020towards}} & 
adversarial:~{\tiny\cite{Schwarz2020NEURIPS,chanmonteiro2020pi-GAN,deng2022gram,rebain2022lolnerf,orel2022stylesdf,gu2021stylenerf,gramHD,xu2021volumegan,tewari2022d3d,pan2022gan2x,mbr_gcorf}}
\newline PCA: {\tiny \cite{Feng:SIGGRAPH:2021,li2021fit,berard2016lightweight,wood16_eccv,ploumpis2020towards,9d7119238bb34585ab26564a124d986b,8373858,ploumpis2020towards}} 
\newline exemplar: {\tiny \cite{Chai2016AutoHairFA,2018arXiv180904765L,hu_avatar_2017}}  &

multi-view: {\tiny \cite{hong2021headnerf,zhuang2022mofanerf,wang_morf} }
\newline monocular images: {\tiny \cite{Schwarz2020NEURIPS,chanmonteiro2020pi-GAN,deng2022gram,rebain2022lolnerf,orel2022stylesdf,gu2021stylenerf,gramHD,xu2021volumegan,tewari2022d3d,pan2022gan2x,mbr_gcorf}}
\newline  monocular video: {\tiny \cite{gafni2021nerface,athar2022rignerf,zheng2022IMavatar,grassal2021neural}} &

canonical space: {\tiny \cite{wang_morf,tewari2022d3d,Feng:SIGGRAPH:2021,li2021fit} }
\newline compositionality: {\tiny \cite{ploumpis2020towards,mbr_gcorf}}
\newline reflectance properties: {\tiny \cite{wang_morf}} \\ %

\bottomrule 
\end{tabular} 
}
\caption{Overview and classification of face-specific methods.} %
\label{table:overview_faces} 
\end{table*}

\begin{table*}[!t]
\small{
\begin{tabular}{m{0.08\linewidth} m{0.33\linewidth} m{0.17\linewidth} m{0.12\linewidth} m{0.17\linewidth}} %
\toprule
\textbf{Dataset} & \textbf{Format and Resolution} & \textbf{Coverage} & \textbf{Samples} & \textbf{Scanner} \\ \midrule

FaceScape \cite{yang2020facescape} & triangle mesh (2M vertices), texture images (resolution $4096\times4096$), raw camera images (359 id $\times$ 20 ex $\times$ ~60 views in 4M-12M pixels) & full head including face, neck, ears, excluding eyes & 938 individuals $\times$ 20 expressions & multi-view system with 68 cameras \\ \hline

Multiface \cite{wuu2022multiface} & triangle mesh (7306 vertices), texture images (resolution $1024{\times}1024$), raw camera images ($2048{\times}1334$), including audio & full head including face, neck, ears & 13 individuals $\times$ 65 (v1), 118 (v2) expressions&  multi-view system with 40 (v1) to 160 (v2) cameras \\\hline

H3DS \cite{ramon2021h3d} & triangle mesh (~120k vertices), texture images (resolution $2048\times2048$), raw camera images ($512\times512$) &  full head including face, neck, ears, eyes closed &23 individuals  &  structured light, multi-view (68 cameras) \\
\hline

CelebV-HQ \cite{zhu2022celebvhq} & monocular video dataset ($512\times512$), with audio, manually annotated 83 facial attributes &  full head including face, neck, ears, eyes &15653 individuals  &  monocular camera \\

\bottomrule
\end{tabular}
}
\caption{Overview of publicly available human face datasets (extends Table 1 from Egger \etal~\cite{egger20203d}).} 
\label{table:faceDatasets}
\end{table*}

\noindent\textbf{2D Supervision.} 
As obtaining large-scale multi-view data is challenging, most methods mentioned above are trained with ${<}410$  identities~\cite{hong2021headnerf}, which impacts generalizability. 
We next discuss methods for building face priors from large-scale monocular data. 
Obtaining monocular data is easier than collecting multi-view data, these methods are often trained with more than $7 \cdot 10^4$  identities~\cite{egger20203d} and, as a result, can generalize better. 

Some methods do not assume the camera poses to be given as input  ~\cite{Schwarz2020NEURIPS,chanmonteiro2020pi-GAN}, while others require them~\cite{rebain2022lolnerf}.
The latter methods learn the face model in an adversarial manner (Sec.~\ref{sec:dataterms}) and often use a generative scene model with a 3D representation (\eg radiance fields or SDF) parameterized by a latent space. During training, they assume a known distribution for camera poses and a fixed latent space---that they sample in every iteration---and render the scene to synthesize 2D images. The models are then trained with the help of a discriminator. Please refer to Fig.~\ref{fig:3d_aware_gan} for a typical 3D-aware GAN pipeline. 

GRAF~\cite{Schwarz2020NEURIPS} and pi-GAN~\cite{chanmonteiro2020pi-GAN} are the first methods to build a generative model with NeRF~\cite{mildenhall2020nerf} in an adversarial manner. 
The sampled images are not as high-quality as image-based generative models~\cite{Karras2019} because of deficient sampling in the volumetric integration of rays. 
Moreover, the Monte-Carlo-based sampling results in ineffective training~\cite{deng2022gram}. 
GRAM~\cite{deng2022gram} overcomes this limitation by learning the radiance fields only on a set of 2D manifolds---which are common across different identities--- 
improving the quality of rendered images. 
However, the learned manifolds are biased toward frontal images, as the dataset primarily consists of frontal-looking images. 
Extreme novel views contain severe artifacts. 
LOLNeRF~\cite{rebain2022lolnerf} shows that 
learning 3D head models from large-scale monocular image collections is also possible using image reconstruction loss instead of purely adversarial loss. 
It expects paired camera pose as input, which they obtain from predicted keypoints. 
However, the random samples from the learned model are not as photorealistic as the ones trained in an adversarial setting.

The methods discussed above need to query coordinate-based MLPs for many points on all the rays to render the full image; they can train models with up to $256{\times}256$ resolution. 
Other recent methods~\cite{orel2022stylesdf,gu2021stylenerf,gramHD,xu2021volumegan} try to overcome this limitation, \ie they maintain 3D representation at a lower resolution and apply a super-resolution module that takes the rendered 3D data to synthesize high-resolution 2D images. 
Still, this policy is not truly multi-view consistent as the  super-resolution module operates in 2D. 
To overcome the limitation of not being able to train implicit models at high-resolution because of computational complexity, EpiGRAF~\cite{epigraf} proposed a novel space and scale aware discriminator which enable patch-based training of the generator model.
Recently proposed D3D~\cite{tewari2022d3d} learns a canonical space of faces without supervision, 
which helps in downstream tasks like color and segmentation transfer between faces. 
\extended{D3D learns a deformation network that brings every point in observed space to a canonical space, where the output of a radiance field is regressed. This forces the network to decouple the shape from the color. }
GAN2X~\cite{pan2022gan2x} utilizes StyleGAN2 \cite{Karras2020stylegan2} to create pseudo-multi-view
images for a given input image and has an explicit 3D-to-2D image formation model. 
The obtained labels are used to learn geometry, appearance, and illumination parameters in an iterative manner. 
The discussed methods model face as a single entity, although it has multiple semantic parts. %
gCoRF~\cite{mbr_gcorf} addresses this concern by explicitly representing each part of the face by a separate 3D representation.
This enables exciting applications, such as editing facial regions in volumetric space. 

\noindent\textbf{Monocular Video.} 
Several methods require a monocular video of a person during training to recover geometry and appearance. 
As they can learn from multiple video frames, they typically can capture high-quality face geometry and appearance. 
However, the quality comes at the cost of collecting a person-specific video. 

NerFAC~\cite{gafni2021nerface} models videos with dynamic faces with a 3DMM. 
3DMM helps them bring the rays to a canonical space with a rigid transformation, and they learn a neural network in this space conditioned on tracked expression parameters to regress the radiance field for each frame. 
RigNeRF~\cite{athar2022rignerf} uses a similar approach---but with an explicit deformation field as a function of expression parameters---to bring to the canonical space instead of naively conditioning the neural network as in NerFACE~\cite{gafni2021nerface}. 
Recently proposed I M Avatar~\cite{zheng2022IMavatar} utilizes occupancy fields to model geometry. 
The critical contribution of this method is an analytical gradient formulation for the iteratively located surface intersection via implicit differentiation, which allows for end-to-end training. 
It also makes use of a tracked mesh using a 3DMM along with per frame delta blendshapes and skinning weights to bring the points to canonical space, in which the texture is modeled. 
The method of Grassal \etal~\cite{grassal2021neural} utilizes an explicit mesh-based model to address a similar problem. 
Along with a base geometry, which they get from a tracked face using a  3DMM, they also predict vertex offsets as a function of the head  pose. 
This makes the method more compatible with the traditional graphics pipeline.

\subsubsection{Specialized Models of Face Parts} 
Faces are complex, and some facial components that are hard to  model with global models are targeted with specific models. 
The data availability is a key difference between such models in contrast to whole-face models. 
Whilst various datasets are available for faces, there are very few shared datasets for individual facial regions. 
This results in slower development of specialized models, and we observe that state of the art in the monocular setting is not yet using modern learning and neural rendering techniques. 

The initial methods~\cite{Chai2016AutoHairFA,2018arXiv180904765L,hu_avatar_2017} for monocular hair reconstruction relied on a database retrieval. 
In contrast, recent methods train neural architectures to regress hair shape  directly~\cite{hairnet_2018,Saito:2018:3HS,yang2019dynamic}. 
The approaches targeting high-quality eye and ear reconstruction follow face 3DMM methods by building separate 3DMMs for  eyes~\cite{berard2016lightweight,wood16_eccv,ploumpis2020towards} and ears~\cite{9d7119238bb34585ab26564a124d986b,8373858,ploumpis2020towards}.

\subsubsection{Data}
\label{sec:faces_datasets}
A recent survey \cite{egger20203d} summarized the publicly shared face datasets at the time. We, therefore, focus on the datasets that arrived since (following the format of Table~1 in \cite{egger20203d}) and present our extension in Tab.~\ref{table:faceDatasets}. Along with 3D data (multi-view images), we also discuss a monocular video dataset~\cite{zhu2022celebvhq}, as there have been methods in the past taking advantage of video datasets for learning 3DMMs of faces~\cite{tewari2019fml,mbr_CFML}.

\subsubsection{Limitations and Outlook}

Most methods that rely on explicit 3DMMs fail to capture fine-scale geometric details of faces that are perceptually important. Recently proposed implicit methods trained in an adversarial manner (3D-aware GAN) show promising results in obtaining some fine-scale details~\cite{Chan2021}. We show qualitative comparisons in Fig.~\ref{fig:faces_reconsruction}. 
Note that no existing monocular implicit methods quantitatively evaluate the 3D reconstruction accuracy w.r.t~ground-truth 3D shapes. 
Moreover, since all the implicit methods are generally over-parameterized with neural networks, they can overfit to test images by baking in geometric details into texture space. 
They also can converge to inaccurate geometry if the  initialization during test time is incorrect. 
It is also observed that methods with implicit representations,  which take advantage of a large-scale video dataset~\cite{zhu2022celebvhq}, are under-explored compared to methods with explicit representation. 
This could be an interesting direction to achieve higher performance in capturing especially better facial expressions by exploiting the nature of the data.

\subsection{Hands}
\label{subsec:hands}

\begin{figure}
    \centering
    \includegraphics[width=\linewidth]{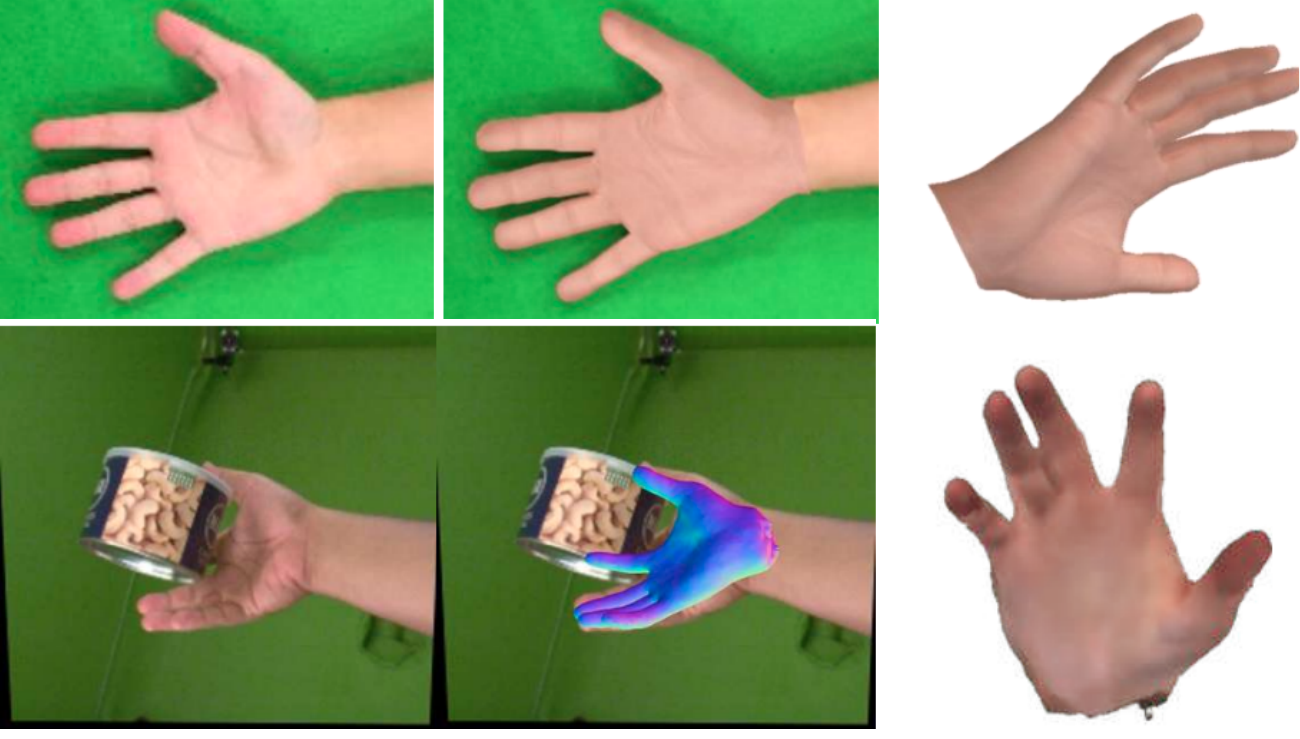}
    \caption{Parametric hand models HTML \cite{Qian2020} (top row) and LISA \cite{corona2022lisa} (bottom row) support hand texture. Row-wise from left to right: The input image, the 3D hand reconstruction overlaid on the input, and a novel view of the reconstruction. %
    Images adapted from \cite{Qian2020, corona2022lisa}.
    } 
    \label{fig:HTML_LISA} 
\end{figure}

Similar to human bodies, human hands are articulated objects with pose-dependent deformations on a fine scale. 
In contrast to human bodies, hands cause more severe self-occlusions and do so more often, especially in the monocular setting.  %
Simple hand movements can be densely tracked in 3D by SfT methods \cite{vicente2013balloon, Yu2015}\extended{~(see Sec.~\ref{sec:sft} for an overview of SfT methods)}. 
However, this requires a known 3D template of the observed hand in advance. %
Moreover, SfT methods are not robust to large self-occlusions that are typical for hands\extended{~(\eg working with various instruments or showing gestures)}. 
A stronger 3D shape prior can help to mitigate these challenging self-occlusions and appearance variations, \ie a statistical parametric hand model covering the entire space of hand shapes \cite{MANO:SIGGRAPHASIA:2017, Qian2020}.

\noindent\textbf{Single Hands.} 
There are several approaches for 3D shape and pose estimation from monocular inputs \cite{boukhayma20193d, Baek2019, Zhang2019, Ge2019,  Zimmermann2019,zhou2019monocular}. 
They all regress parameters of the MANO model \cite{MANO:SIGGRAPHASIA:2017} and differ in their architectures, supervision, and fine-tuning policy for in-the-wild data. 
These methods rely on a differentiable mesh renderer \cite{Baek2019} or depth map rendering \cite{Ge2019}, differentiable re-projection loss \cite{Zhang2019} or re-projection of 3D hand joints to images (2D keypoints) \cite{boukhayma20193d, Zimmermann2019, zhou2019monocular}. 
All of them train on synthetic or mixed datasets with ground-truth 3D hand meshes and poses and 
some fine-tune on in-the-wild images using either 2D annotations only \cite{boukhayma20193d} or rendered depth maps \cite{Ge2019}. 
Moreover, MANO differentiability enables end-to-end trainable architectures \cite{boukhayma20193d, Zhang2019}. 
Further characteristics of these methods are: Boukhayma  \etal~\cite{boukhayma20193d} employ a convolutional encoder and a fixed MANO-based decoder; 
Baek \etal's hand mesh estimator (HME) \cite{Baek2019} is  supervised by 3D skeletons and hand silhouettes; 
Ge \etal~\cite{Ge2019} use a GCNN for mesh generation; 
Zhang \etal~\cite{Zhang2019} regress camera and mesh parameters with an iterative regression module; 
finally, Zhou \textit{et al.}~\cite{zhou2019monocular} apply an inverse kinematics network, for the first time in the context of hands. 
Moreover, their proposed decoupling of image-to-keypoint and keypoint-to-angles regression allows them to train on all available data modalities, \ie 2D and 3D annotated image data as well as pure motion capture data without paired images.

The methods discussed above 
estimate hand shapes from single images independently; their results on videos can be jittery. 
In contrast, the SeqHAND approach \cite{Yang2020} integrates temporal  consistency constraints by learning visual and temporal features from a synthetic dataset mimicking hand movements. 
Noteworthy is their synthetic-to-real fine-tuning policy involving  detaching the recurrent layer from the core architecture and replacing the video input with single real images. 
A recent transformer-based work by Park \etal~\cite{Park2022} on 3D mesh estimation targets robustness against occlusions. 
They are interested in scenarios with hand-object interactions and treat objects as occluders. 
\extended{The main idea of their architecture (operating on a  per-frame basis) is to inject hand information into occluded regions with the subsequent refinement of the final 3D estimates.}

Several works improve upon different aspects of the MANO model. 
HTML \cite{Qian2020} is the first parametric hand texture model %
and it is learned from over one hundred SfM 
scans 
representing people of different genders, ages, and skin colors. 
HTML can regress both shapes and texture thanks to an analysis-by-synthesis photometric loss. %
Such a loss affects the shape estimates due to the additional supervision signal (the re-projected texture). 
LISA~\cite{corona2022lisa} is another hand model based on MANO that supports hand textures, but it uses different shape parameters and the shapes are represented by implicit functions.   
It has a disentangled parameter spaces for texture, shape, and poses learned from multi-view RGB videos annotated with 3D joints. 
Like HTML, LISA can reconstruct hands from monocular RGB images; see Fig.~\ref{fig:HTML_LISA}. 
DeepHandMesh \cite{Moon2020_DeepHandMesh} is a neural encoder-decoder 
that leverages a personalized hand model (\ie assuming the same subject at training and test) trained in a weakly-supervised manner from multi-view depth maps. 
It addresses MANO's limited resolution and implements a penetration avoidance loss to make hand-surface interactions more physical plausible. %
However, more identity-specific geometric details require a new training dataset for each identity and limit the generalizability to other identities.

\begin{figure} 
    \centering 
    \includegraphics[width=\linewidth]{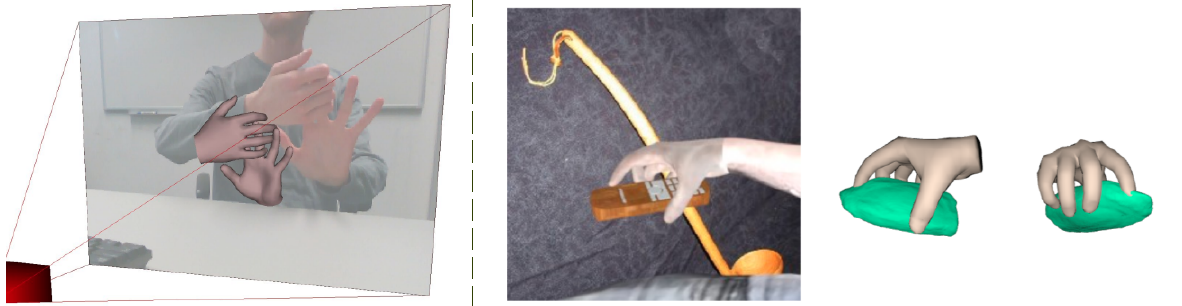} 
    \caption{Monocular 3D reconstruction of two  interacting hands (RGB2Hands \cite{Wang2020}) and joint hand-object 3D reconstruction (Grasping Field  \cite{Karunratanakul2020}) are emerging directions.
     Images adapted from \cite{Wang2020, Karunratanakul2020}.
     } 
    \label{fig:TwoHandsObjects} 
\end{figure}

\noindent\textbf{Two Hands; Hands and Objects.} 
Reconstructing two interacting hands adds complexity to the problem due to mutual hand occlusions\extended{~(in addition to self-occlusions)} and hand interactions that affect the surfaces of both hands. 
Only recently, the first solution to this challenging task %
has been introduced by Wang \etal \cite{Wang2020} (Fig.~\ref{fig:TwoHandsObjects}-(left)). 
Their RGB2Hands method takes inherent depth ambiguities and mutual hands occlusions into account. 
It intermediately estimates %
segmentation for the  handiness, inter-hand relative depth, and inter-hand distances. 
However, intertwined fingers can lead to hand-hand penetrations. 
The follow-up HandFlow \cite{Wang2022VMV} predicts a distribution of plausible hand poses instead of a single estimate. %
The authors highlight that current evaluation schemes assuming a single correct hand pose are deficient. 
Zhang \textit{et al.}~\cite{Zhang2021ICCV} leverage a hand-pose-aware attention block for per-hand feature extraction and a cascaded refinement block. 
The latter improves the initially estimated hand poses and shapes in the MANO space taking into account the interaction context between two hands. 
The proposed method achieves state-of-the-art accuracy and improvements in scenarios with inter-hand occlusions. 
Keypoint Transformer \cite{Hampali_2022_CVPR_Kypt_Trans} predicts 3D poses of objects and hands observed in a single RGB image. 
The method includes three stages: It first detects and disambiguates the hand keypoints using a self-attention mechanism and then estimates the 3D hand poses with a cross-attention module. 
Another recent work \cite{Li2022hands} further advances the two-hand case with the help of GCNNs and two attention blocks and shows a live demo of the proposed method. 
It accurately reconstructs challenging in-the-wild images with inter-hand occlusions. 
As of this writing, the last two discussed methods are the most accurate on the InterHand2.6M benchmark in the literature.

The joint tracking of hands and objects is an emerging area. 
Karunratanakul~\etal's Grasping Field \cite{Karunratanakul2020} is a new joint representation for hands, objects and the contact areas using implicit surfaces. 
They propose a neural method for hand-object reconstruction, assuming that a 3D model is given as input; see Fig.~\ref{fig:TwoHandsObjects}-(right). 
Hasson \etal~\cite{Hasson2019} jointly reconstruct the shapes of a hand and an object after training on a new synthetic dataset. 
They argue that object manipulation simplifies the problem by providing more constraints and show that it improves grasp metrics. 
\extended{Their contact loss penalizes penetrations between the object and the hand.} 
A follow-up work \cite{Hasson2020} assumes that a 3D model of the observed model is given. 
Ye \textit{et al.}~\cite{Ye2022hand} make a related observation that hand articulations are driven by local object shapes. 
Starting from the input image and hand and camera poses estimated by an off-the-shelf system  \cite{Rong2021frankmocap}, they reconstruct the object shape with an SDF decoder for the object shape. 
Like Hasson \etal~\cite{Hasson2019}, they encourage contact between the hand shape and the object at pre-defined regions \cite{Hasson2019}.

\noindent\textbf{Datasets.} 
Only a few datasets in the literature provide RGB images and corresponding 3D shape annotations. 
FreiHAND is the benchmark for 3D hand pose and shape estimation of a single hand \cite{Zimmermann2019}. 
ObMan \cite{Hasson2019} and DexYCB \cite{Chao_2021_CVPR} contain shape annotations for single hands and objects. 
While ObMan provides single synthetic images, the more recent DexYCB includes videos of real grasping scenes recorded from multiple views. 
Moon \etal~\cite{Moon2020} introduce the InterHand2.6M dataset. %
Mesh annotations for it are also available thanks to NeuralAnnot \cite{Moon2022_NeuralAnnot}. 
MultiHands \cite{Wang2022VMV} is an extension of InterHand2.6M with 100 additional annotations per image, which allows quantifying pose  ambiguities as a distance between the predictions and ground-truth pose distributions. 
Finally, H2O is a popular dataset with shape annotations for two hands manipulating rigid objects \cite{Kwon2021_H2O}, and $\text{H}_2\text{O}$-$\text{3D}$ is currently the most challenging dataset with accurately annotated videos of two hands manipulating objects (due to large mutual occlusions caused by hands and objects).

\noindent \textbf{Future Directions.} 
\extended{There is much room for improvement in parametric hand shape modeling.}  
Existing models lack geometric and pose-dependent texture details  (\eg nails, hair and blood vessels). 
We will soon see new 
methods for the 3D shape estimation of 1) hands and articulated objects and 2) hands and deformable objects. 
Moreover, reconstruction under various illumination conditions remains not solved satisfactorily, and 
reconstruction of interacting hands can advance further by improving mesh collision handling.

\subsection{Animals}\label{sec:animals}

Unlike Sec.~\ref{sec:fewscene} and \ref{sec:manyscene}, this section discusses animal reconstruction methods that use parametric models. 
Apart from the seminal work by Cashman \etal~\cite{Cashman2013Dolphins} reconstructing dolphins, interest in animal-centered reconstruction has started growing only recently with the introduction of the SMAL model~\cite{ZuffiSmal2017}, a SMPL-style model for quadrupeds learned from 3D scans of animal toys. 
It enables sufficient regularization to cope with the lack of large, high-quality datasets as are widely used in the mature areas of face and human reconstruction, which is both due to less \apriori~interest and the difficulty of capturing a wide variety of animals in a highly controlled setting. 
Biggs \etal~\cite{biggs2018creatures} fit the SMAL model to videos instead of images. 
Zuffi \etal's~\cite{ZuffiSmallr2018} SMALR uses keypoints and silhouettes to deform the SMAL model with per-vertex offsets beyond the parametric shape space. 
In the follow-up work 3D Safari~\cite{Zuffi_2019}, they train a regression network on synthetic Zebra images and apply it to real data without annotations at test time. 
Dogs have also received some attention: Biggs \etal~\cite{biggs2020wldo} add limb scaling to create a dog-specific SMAL model from internet images annotated with 2D keypoints and silhouettes. 
Li \etal~\cite{li2021coarsetofine} use graph convolutions in a hierarchical manner to refine a regressed SMAL mesh with per-vertex deformations. 
Most recently, BARC~\cite{ZuffiBarc2022} turns SMAL into a breed-aware dog model by exploiting breed labels in a triplet loss. 
Another line of work focuses on birds:  %
Badger \etal~\cite{badger2020} build a SMPL-style parametric model of cowbirds without access to 3D scans, which is then applied to monocular regression of its parameters. 
Wang \etal~\cite{wang21aves} generalize this model to multiple bird  species (see Fig.~\ref{fig:teaser}, top row; second from the left). 
Data-driven general methods from Sec.~\ref{sec:manyscene}, like CMR~\cite{cmrKanazawa18} in Fig.~\ref{fig:cmr}, often evaluate on the CUB birds dataset and do not use a parametric model, which leads to very coarse reconstructions. 
For more discussion on bird reconstruction, we refer to \cite{MarvastiZadeh2022}. 
Wu \etal~\cite{wu2022casa} generalize across species by first retrieving a rigged template mesh via CLIP features~\cite{radford2021learning} from a template database. 

\section{Discussion and Open Challenges}\label{sec:discussion} 
We next elaborate on current challenges in the field and 
discuss two nascent but promising future directions: methods using event cameras and physics-aware approaches.

\noindent\extended{\textbf{Non-Solid Objects.} %
As of yet, no monocular RGB method focuses on objects that are not solid. 
However, there is some sparse multi-view work \cite{Franz_2021_CVPR, chu2022physics} that handles fluids, like smoke. 
Unlike the object-particle-focused Lagrangian formulation, the time evolution of fluids in the reconstruction setting is more naturally expressed in the coordinate-system-focused Eulerian representation. 
}

\noindent\textbf{Large Scale.} %
While static methods~\cite{zhang2020nerfplusplus} can handle large scenes, only a few recent dynamic methods cope with a static background \cite{li2021neural, park2021nerfies, yoon2020dynamic, neural-human-radiance-field:neuman}. 
Distant background, even if static, is hardly reconstructed by current methods. %

\noindent\textbf{Multiple Objects.} %
Static methods already scale to 
scenes with multiple objects \cite{Ost_2021_CVPR}. %
However, handling multiple dynamic objects in the same scene is still in its infancy \cite{Menapace_2022_CVPR, Menapace2023arXiv}.

\noindent\extended{\textbf{Invertible Deformation Models.} %
Most deformation models are purely forward or purely backward, and do not admit an easy inversion to obtain the other. 
However, especially volumetric-rendering methods with their backward models would benefit from invertible deformations, \eg for editability (see below). 
Some first attempts already exist~\cite{chen2021snarf}. %
Invertibility by construction in gradient-based settings leads to \emph{diffeomorphisms}, differentiable homeomorphism. 
In practice, such mappings coincidentally tend to allow for easily computable forward and backward deformations. 
For example, the Neural-ODE-based~\cite{chen2018neural} OccupancyFlow~\cite{OccupancyFlow} and NeRFlow~\cite{du2021neural} trade-off less expressibility and slower runtime for guaranteed invertibility. 
}

\noindent\textbf{Editability.} %
Beyond mere reconstruction, the ability to edit the scene's deformations, geometry, and appearance would enable the easy creation of digital assets (\eg for interactive AR/VR). 
Classical geometry and appearance representations already  possess an extensive toolbox for editing. 
However, deformations remain challenging to manipulate,  especially for non-expert end users. 
While driving coarse deformations by re-posing the skeleton of a skinned template is relatively straightforward, creating the corresponding finer deformations (\eg of cloth), remains  difficult. 
Scene editing becomes even more challenging when using  volumetric representations, like modern neural  parametrizations for geometry and appearance: 
The latter use backward deformation models, where manipulation is less intuitive and more involved. 
We refer to a recent survey \cite{tewari2021advances} for  progress on editing neural representations.

\noindent\textbf{Real-Time Performance.} %
Some category-specific methods \cite{tewari2017self} are  already capable of real-time performance. 
In the general setting, real-time performance comes at the cost of noticeably lower quality~\cite{Yu2015}. 
Related single-camera settings pave promising paths towards real-time high-quality general dynamic reconstruction: general dynamic RGB-D reconstruction \cite{newcombe2015dynamicfusion,lin2022occlusionfusion} has a long history of real-time speed, and classical sparse RGB SLAM~\cite{ORBSLAM3_TRO} and neural dense RGB-D SLAM~\cite{Zhu2022CVPR} also run at real-time rates, with neural dense RGB SLAM very recently achieving the same \cite{chung2022orbeez,rosinol2023nerfslam}.  %

\noindent\extended{\textbf{Higher-Level/Downstream Tasks.} %
While this STAR focuses on the reconstruction problem itself, the reconstruction can be useful for downstream tasks like editing (see above), higher-level scene understanding (\eg, semantic labeling or affordance estimation), active learning (\eg, best-next-view selection), or motion planning. 
These could also be integrated into the reconstruction algorithm, allowing for joint optimization and potential mutual improvement. 
}

\noindent\textbf{Data Bias.} %
Reducing biases in the data %
is an open challenge not only in 3D reconstruction but in computer vision in general. 
Different ethnic groups and minorities are underrepresented in most existing datasets, which makes them unbalanced. 
As a result, methods trained on them (\eg to estimate texture or albedo) %
are often biased towards statistically expected skin colors (\ie light tones). 
Special care should be taken when acquiring data so that as many ethnicities as possible are represented in the samples \cite{Qian2020}. 
Moreover, benchmarks that quantify biases are of great help \cite{Feng:TRUST:ECCV2022}. 

\noindent\textbf{Model Variety.} 
Morphable and parametric models assume able-bodied  individuals; missing limbs are seldom modeled. 
The same holds for highly individualistic appearance  variations like tattoos.

\noindent\textbf{Event Cameras.} %
As monocular non-rigid 3D reconstruction from event cameras is an emerging domain  with only a few published works, we discuss them jointly here instead of in their respective sections. 
Their design 
is challenging as existing RGB-based 
techniques 
are not directly applicable to event streams. 
Event cameras provide an ultra-high temporal event resolution ($\approx$1${\mu}s$)
and record with high dynamic range (see Sec.~\ref{sec:rendering-into-2d}). 
Hence, they are well-suited for high-speed motions in challenging lighting conditions.

\extended{
\begin{figure} 
    \centering 
    \includegraphics[width=\linewidth]{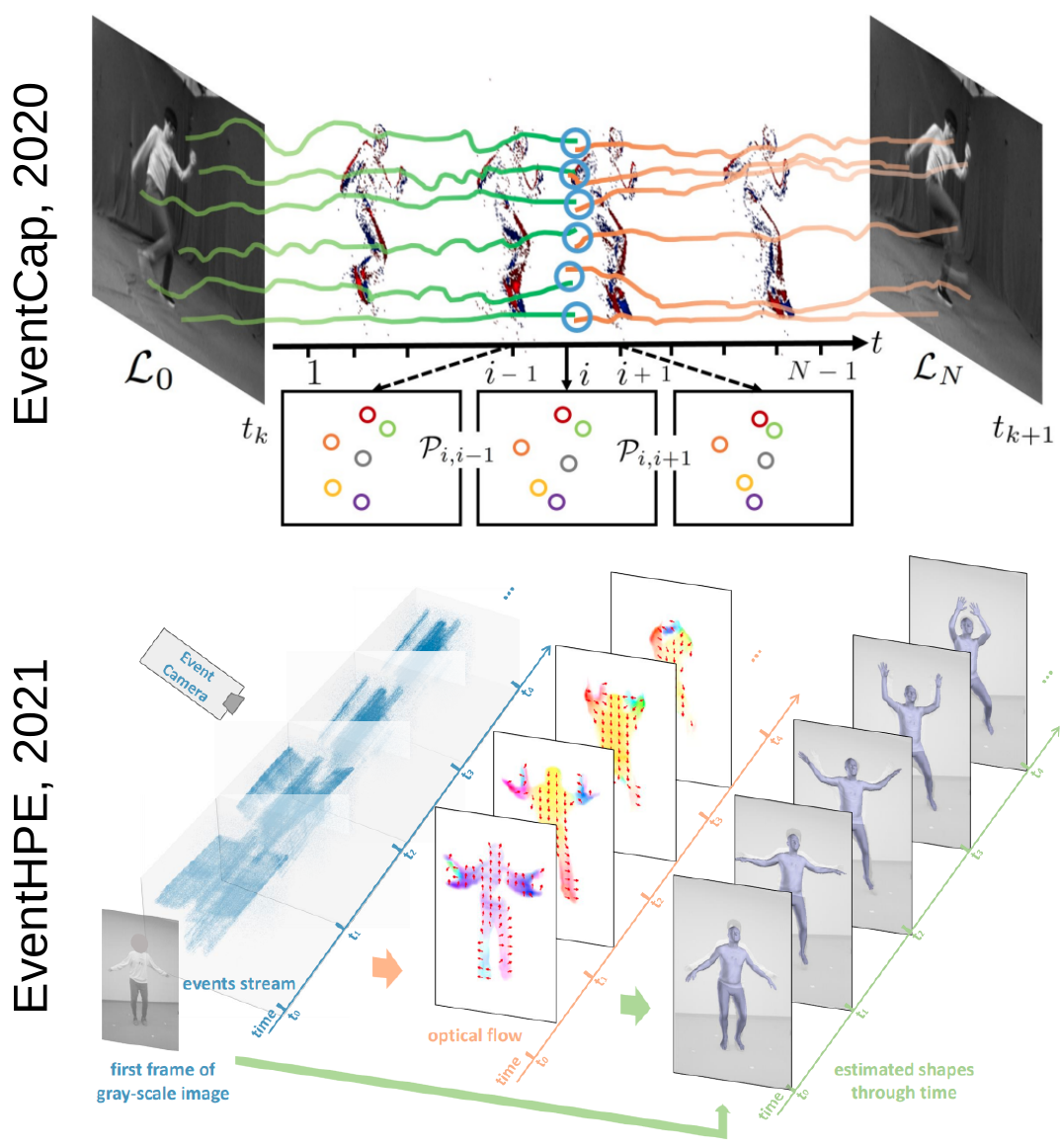} 
    \caption{\extended{While EventCap \cite{eventCap2020CVPR} tracks 2D features between keyframes using events relying on Levenberg-Marquardt, EventHPE \cite{Zou2021} is a learning approach that requires a single intensity image for initialization.
    } 
    } 
    \label{fig:event_based} 
\end{figure} 
}

EventCap \cite{eventCap2020CVPR} tracks a human in 3D from a hybrid input of events and synchronous greyscale images captured at 
${\leq}25$ fps. 
For its highest accuracy, EventCap requires a rigged and skinned 3D human template
but also supports
SMPL \cite{SMPL:2015}. 
Note that 
the events and images are captured by the same sensor, 
\ie 
\textit{the scene is observed from a single view.} 
EventCap uses events to track 2D features and establish  correspondences between the greyscale keyframes. 
That is because, for high-speed motions, the 2D point trajectories guided by events can differ significantly from linear feature interpolation between the keyframes. %
EventHPE~\cite{Zou2021} 
relies only on a single greyscale frame for the 3D human pose  initialization. 
It is a learning-based human-specific approach trained on a new dataset with event streams and corresponding SMPL annotations. 
It uses an unsupervised warping loss with events-based optical flow. 
Nehvi \etal~\cite{Nehvi_2021_CVPR} track general objects from an event camera. 
Their analysis-by-synthesis SfT approach searches for 3D states 
obeying 
the deformation model (such as ARAP or a parametric shape model  \cite{MANO:SIGGRAPHASIA:2017}) and inducing synthetic events that resemble the observed events. 
The data term accumulates the events into \textit{event  frames}, a 2D representation of accumulated events in short time  intervals. 
EventHands \cite{Rudnev_2021_ICCV} is a data-driven approach for 3D hand pose estimation from a single event stream trained with a synthetic dataset. 
It neither uses greyscale images nor a 3D template\extended{, and thus  supports fewer shape details than Nehvi \etal~\cite{Nehvi_2021_CVPR}}. 
Both EventHPE and EventHands use parametric models %
\cite{SMPL:2015, MANO:SIGGRAPHASIA:2017}. 
EventHands enables the tracking of high-speed hand movements at $1000$ equivalent fps, \ie the number of discretely reconstructed 3D shapes per second. 
A Kalman filter stabilizes the results via temporal smoothness. 

\noindent\textit{Observations.} 
EventHands demonstrates that events are more abstract signals than RGB or greyscale pixel values. 
Thus, the model trained on synthetic data generalizes well to real events. 
Furthermore, all discussed methods %
show that high-speed motions could be reconstructed using much lower bandwidth compared to high-speed RGB recordings. 
Furthermore, they all convert the raw event streams to more suitable 2D representations. %
Finally, a single or a few events are not expressive enough; a critical mass of events is necessary to regress changes in the estimated 3D poses and shapes.

\noindent\textbf{Physics.} 
Physically-based simulation of soft body  dynamics~\cite{bender2017survey} has been well and actively 
studied 
for more than 30 years~\cite{terzopoulos1987elastically}. 
However, unlike this well-posed forward problem, non-rigid reconstruction is inverse and ill-posed, and physics-based methods only form an emerging field. 
In addition to higher computational load, physics-based models are also harder to optimize in practice.
For example, physically meaningful material parameters (which determine the deformations) are often time-invariant, which leads to hard-to-escape local minima due to the strong path dependence in the forward simulation. %
Therefore, physics-based reconstruction methods primarily target simple objects and only elastic phenomena. 
They ignore complex physics such as human skin, muscles, hair and clothing, which are all non-rigid but have different physical properties and varied deformation behavior.  
They also do not account for collisions, contacts, fractures, %
or plasticity. 

Along with that, non-learning methods that model physics and data-driven learning-based methods have shown first success. %
The former necessarily employ some intuition or approximation of physics, %
and they primarily differ by how accurately they model the   physics of deformable objects. 
Thus, a few earlier SfT and NRSfM methods apply 
continuum mechanics as hard constraints by representing surfaces and tracking deformations with finite elements (FEM) \cite{malti2017elastic} or particle-based models \cite{agudo2015simultaneous,ozgur2017particle}. 
Recent advances in differentiable simulation~\cite{li2022diffcloth, Liang2019} and differentiable rendering (Sec.~\ref{sec:rendering-into-2d}) enable physics-based analysis by synthesis: %
$\boldsymbol{\phi}$-SfT~\cite{kair2022sft} reconstructs 3D geometry 
while others  \cite{Jaques2020Physics-as-Inverse-Graphics:, murthy2021gradsim} consider the inverse elasticity problem. 

While early methods were mostly physics-inspired, the community shifted towards learning, particularly neural networks, in the last decade. %
Recently, there has been growing interest in combining physics and learning approaches to achieve robust solutions, as physics is the intuition representing invariant properties of the physical world; and learning could extend this world rather than starting from scratch. 
In the sparse setting, Shimada \etal~\cite{PhysAwareTOG2021} achieve state-of-the-art results in human motion capture with their \textit{physionical} method, a neural approach that is aware of physical and environmental constraints. 
However, extending it to the dense case is not trivial; one of the reasons is the increased requirement for computational resources. 
\extended{ %
Li \etal~\cite{Li2021} generate training data with a physics simulator on-the-fly and use it to train a neural network for 3D human performance capture, including clothes deformations.}
Neural approaches~\cite{raissi2019physics, chen2018neural} could  additionally aid in solving deformation PDEs that, otherwise,  require numerical methods and are often computationally  prohibitive. 
\extended{
Santesteban \etal~\cite{santesteban2022snug} enable self-supervised training of dynamic 3D garment deformations worn by parametric human bodies.
They recast the physics-based deformation model as an optimization problem, as opposed to a traditional frame-by-frame solution using implicit integrators. 
}
Apart from these methods, there remains a wide range of problems where physics-based solutions remain underexplored, as they have only recently become feasible.

\section{Social Implications}\label{sec:social}

We discuss a wide range of potential upsides of monocular reconstruction in the introduction (Sec.~\ref{sec:introduction}), like VR/AR, content creation, robotics, medicine, and many others. 
There are, however, also some potential social downsides, which we discuss in detail here. 

\noindent\textbf{Environment.} 
The rise of neural methods increased the usage of GPUs, which can be harmful to the environment and climate due to the material needed, production, and energy usage when running. 
While the last issue can be addressed by the end user via clean energy, material sourcing and production are more challenging. 
 Still, reconstruction methods that are easy to use might help in environmental research, \eg 3D glacier reconstruction \cite{Pellitero2016, Samsonov2021}, thereby ultimately positively impacting the environment.

\noindent\textbf{Privacy and Consent.} 
Easy-to-employ reconstruction methods can potentially lead to unwanted misuse of personal likenesses. 
Especially when handling datasets containing identifiable data of humans, privacy and consent should be considered, both for training data and at test time when reconstructing other people. 
Furthermore, editability \cite{kim2018deep, Yuan_2022_CVPR, Menapace_2022_CVPR}, which is not a focus of this STAR, could lead to issues with visual content modified or generated with malevolent intent (\eg misinformation). 
The detection of edited content is an active research area \cite{roessler2019faceforensics++, Shiohara_2022_CVPR}. 
Such detectors often exploit expert knowledge about the design of state-of-the-art methods that generate such content in the first place, which makes continued research necessary. 
For a discussion specific to neural rendering, we refer to a recent survey~\cite{tewari2021advances}. 

\noindent\textbf{Inclusiveness.} 
Reconstruction methods can serve as a more inclusive basis for AR/VR if they cover a wider range of variation among people. %
We refer to \emph{Data Bias} and \emph{Model Variety} in Sec.~\ref{sec:discussion}.

\noindent\textbf{Authoritativeness.} 
In certain restricted settings like faces, reconstruction methods are reliable and can help with the virtual ageing of crime victims or face reconstruction from dry skulls \cite{egger20203d}. 
General reconstruction methods, however, should not be taken as authoritative, \eg in legal contexts. 
Since their problem setting %
is severely ill-posed, the results are only plausible: consistent but merely possible.
They do not infer reliable information about reality beyond what is in the input (\eg how a suspect handled a gun hidden behind their back while being recorded from the front). 

\noindent\textbf{Accessibility.} 
The research field is quite accessible: Papers are mirrored on public sites; code and dataset releases are common%
; some limited GPU resources are accessible for free in the cloud, with larger resources requiring `only' money and no longer one's own physical infrastructure; %
and RGB cameras are easily obtainable. 

\section{Conclusions}\label{sec:conclusion}

We traced how the deep learning revolution, including differentiable rendering and neural rendering, has spread the general non-rigid 3D reconstruction field beyond NRSfM and SfT. 
This is especially promising considering the saturation of improvements in NRSfM. 
Still, general methods remain in an early phase and 
far from being solved, with much of the design space under-explored at best. 
Category-specific methods for humans and faces are maturing, with close to photorealistic results, while hands and animals, with their unique challenges, have seen comparatively less work. %
Orthogonal to these, we discussed several promising developments that have recently become practically feasible for reconstruction, like physics simulation and event cameras. 
In addition, we described the components of the reconstruction pipeline in detail and commented on several open challenges and social implications, which we believe future research would benefit from considering. 
We hope this STAR will serve as an informative overview for established researchers and a helpful starting point for 
newcomers
entering this exciting and fast-changing area. %

\section{Acknowledgements.} 
N. Kairanda, V. Golyanik, and C. Theobalt are supported in part by the ERC Consolidator Grant \textit{4DReply} (770784). 
A. Kortylewski acknowledges support via his Emmy Noether Research Group funded by the German
Science Foundation (DFG) under Grant No. 468670075.
The authors thank Rhaleb Zayer for his helpful feedback on Sec.~\ref{sec:fundamentals}.

\bibliographystyle{eg-alpha-doi} 
\bibliography{main}

\end{document}